\documentclass[lettersize,journal]{IEEEtran}
\usepackage{amsmath,amsfonts}
\usepackage{array}
\usepackage[caption=false,font=normalsize,labelfont=sf,textfont=sf]{subfig}
\usepackage{textcomp}
\usepackage{stfloats}
\usepackage{url}
\usepackage{verbatim}
\usepackage{graphicx}
\usepackage{cite}
\usepackage{graphicx}
\usepackage{multirow}
\usepackage{times}
\usepackage{amsmath,amsthm}
\usepackage{amssymb}
\usepackage{amsfonts}
\usepackage{color}
\usepackage{mathrsfs}
\usepackage{booktabs} % For formal tables
\usepackage{makecell}
\usepackage{array}
\usepackage{caption}

\usepackage{color}
\usepackage{hyperref} 
\usepackage{url}
\usepackage{algpseudocode}
\usepackage{pifont}
\usepackage{bm}
\usepackage{comment}
\usepackage{wasysym}
\usepackage{float}
\usepackage{threeparttable}

\usepackage[utf8]{inputenc}

\usepackage[usenames,dvipsnames]{xcolor}
\usepackage{tikz} \usetikzlibrary{calc, arrows.meta, intersections, patterns, positioning, shapes.misc, fadings, through,decorations.pathreplacing,mindmap,trees}

\def\1{{\bf{1}}}
\def\0{{\bf{0}}}

\def\i{{\bf i}}
\def\j{{\bf j}}

\def\x{{\bf x}}

\def\H{{\bf H}}

\def\T{{\bf T}}

\def\Ecal{{\mathcal{E}}}

\def\Ncal{{\mathcal{N}}}

\def\Rcal{{\mathcal{R}}}

\def\Tcal{{\mathcal{T}}}

\def\Vcal{{\mathcal{V}}}

\def\Rbb{{\mathbb R}}

\def\Zbb{{\mathbb Z}}

\newtheorem{definition}{\textbf{Definition}}

\definecolor{ColorOne}{named}{MidnightBlue}
\definecolor{ColorTwo}{named}{Dandelion}
\definecolor{ColorThree}{named}{Plum}

\begin{document}

\title{Graph Learning and Its Advancements on \\ Large Language Models: A Holistic Survey}

        % <-this % stops a space

\author{Shaopeng~Wei,
        Jun~Wang,
        Yu~Zhao,~\IEEEmembership{Member,~IEEE},
        Xingyan~Chen,
        Qing~Li,~\IEEEmembership{Member,~IEEE},
        Fuzhen~Zhuang,~\IEEEmembership{Member,~IEEE},
        Ji~Liu,~\IEEEmembership{Member,~IEEE},
        Fuji~Ren, ~\IEEEmembership{Senior Member, ~IEEE},
        Gang~Kou

\thanks{Dingwei Chen, Xuehui Chen, Fangjing Chen, Yangxiang Zhou, Tiantian Zeng, Ruoqi Yang, Mingchen Ye, Leqi Chen, Yiran Wan and Siyi Yao are collecting and sorting out the literature for this work.}% <-this % stops a space

\thanks{S. Wei
and G. Kou are with School of Business Administration, Southwestern University of Finance and Economics,  China.\protect\\
E-mail: weishaopeng1997@gmail.com}

\thanks{F. Zhuang is with Institute of Artificial Intelligence, Beihang University, Beijing, China, and with Zhongguancun Laboratory, Beijing, China.}

\thanks{J. Liu is with Meta, USA.}

\thanks{F. Ren is with University of Electronic Science and Technology of China, China.}

\thanks{Y. Zhao, J. Wang, Q. Li and X. Chen are with Fintech Innovation Center, Financial Intelligence and Financial Engineering Key Laboratory of Sichuan Province, Institute of Digital Economy and Interdisciplinary Science Innovation, Southwestern University of Finance and Economics, China. }

\thanks{Y. Zhao (zhaoyu@swufe.edu.cn) and G. Kou (kougang@swufe.edu.cn)  are corresponding authors.}}


\markboth{Journal of \LaTeX\ Class Files,~Vol.~14, No.~8, August~2021}%
{Shell \MakeLowercase{\textit{et al.}}: A Sample Article Using IEEEtran.cls for IEEE Journals}

\IEEEpubid{0000--0000/00\$00.00~\copyright~2021 IEEE}

\maketitle

\begin{abstract}
Graph learning is a prevalent domain that endeavors to learn the intricate relationships among nodes and the topological structure of graphs. Over the years, graph learning has transcended from graph theory to graph data mining. With the advent of representation learning, it has attained remarkable performance in diverse scenarios. Owing to its extensive application prospects, graph learning attracts copious attention. While some researchers have accomplished impressive surveys on graph learning, they failed to connect related objectives, methods, and applications in a more coherent way. As a result, they did not encompass current ample scenarios and challenging problems due to the rapid expansion of graph learning. Particularly, large language models have recently had a disruptive effect on human life, but they also show relative weakness in structured scenarios. The question of how to make these models more powerful with graph learning remains open. Our survey focuses on the most recent advancements in integrating graph learning with pre-trained language models, specifically emphasizing their application within the domain of large language models.
Different from previous surveys on graph learning, we provide a holistic review that analyzes current works from the perspective of graph structure, and discusses the latest applications, trends, and challenges in graph learning. Specifically, we commence by proposing a taxonomy and then summarize the methods employed in graph learning. We then provide a detailed elucidation of mainstream applications. Finally, we propose future directions.
\end{abstract}


\IEEEPARstart{G}{raph} learning (GL) aims to model graphs, a type of non-Euclidean data that differs significantly from previous data structures in machine learning. Graphs are present in various real-world scenarios, 
% \IEEEPARstart{G}{raph} learning 
% has achieved tremendous progress in various scenarios, 
such as social network 
\cite{Mcpherson2001Birds,Zhang2015Cosnet},
% \cite{Mcpherson2001Birds,Leskovec2010Signed,Zhang2015Cosnet}, 
academic network
\cite{Chiang2019Cluster,dong2017metapath2vec,Hu2020Heterogeneous},
% \cite{tang2008arnetminer,dong2017metapath2vec,Hu2020Heterogeneous}, 
e-commerce networks \cite{JiliangTang2013SocialRA,WangHao2022HyperSoRecEH}, enterprise knowledge graphs \cite{cheng2020delinquent,yang2021financial,zhang2022heterogeneous}.
It is crucial to extract rich information from the complex connections between nodes and the topological structure of graphs. Graph learning benefits various tasks and applications involving graphs.

Additionally, graph data can be derived from traditional applications, such as computer vision \cite{Hu2018Relation,Sarlin2020Superglue}, language models \cite{yasunaga2021qa,Yao2018Graph},
% \cite{Phuc2022DB}
% physics \cite{Sanchez-Gonzalez2018Graph,Chen2022Physics}, 
and chemistry \cite{Schlichtkrull2018Modeling,Rong2020Self}. The potential assumption is that there are many latent connections among different entities that are not observed directly. Thus, graph learning is not only a way to deal with natural graph structures but also a way of thinking about various kinds of problems.

Owing to the auspicious future of graph learning, it has attracted a lot of interest worldwide.
%Due to the promising future of graph learning, it has garnered a lot of attention worldwide. 
However, despite previous theoretical works on graphs that help people understand various characters on graphs and provide basic analysis frameworks, these works usually concentrate on small and simulated graphs \cite{Mcpherson2001Birds,Fouss2007Random,kuang2012symmetric}. 
This constrains their applications in real scenarios, especially when intricate relationships and structures prevail on graphs.
% This limits their applications in real scenarios, especially when complex relationships and structures exist on graphs.

\begin{figure}[htb]
    \centering
    \includegraphics[width=0.40\textwidth]{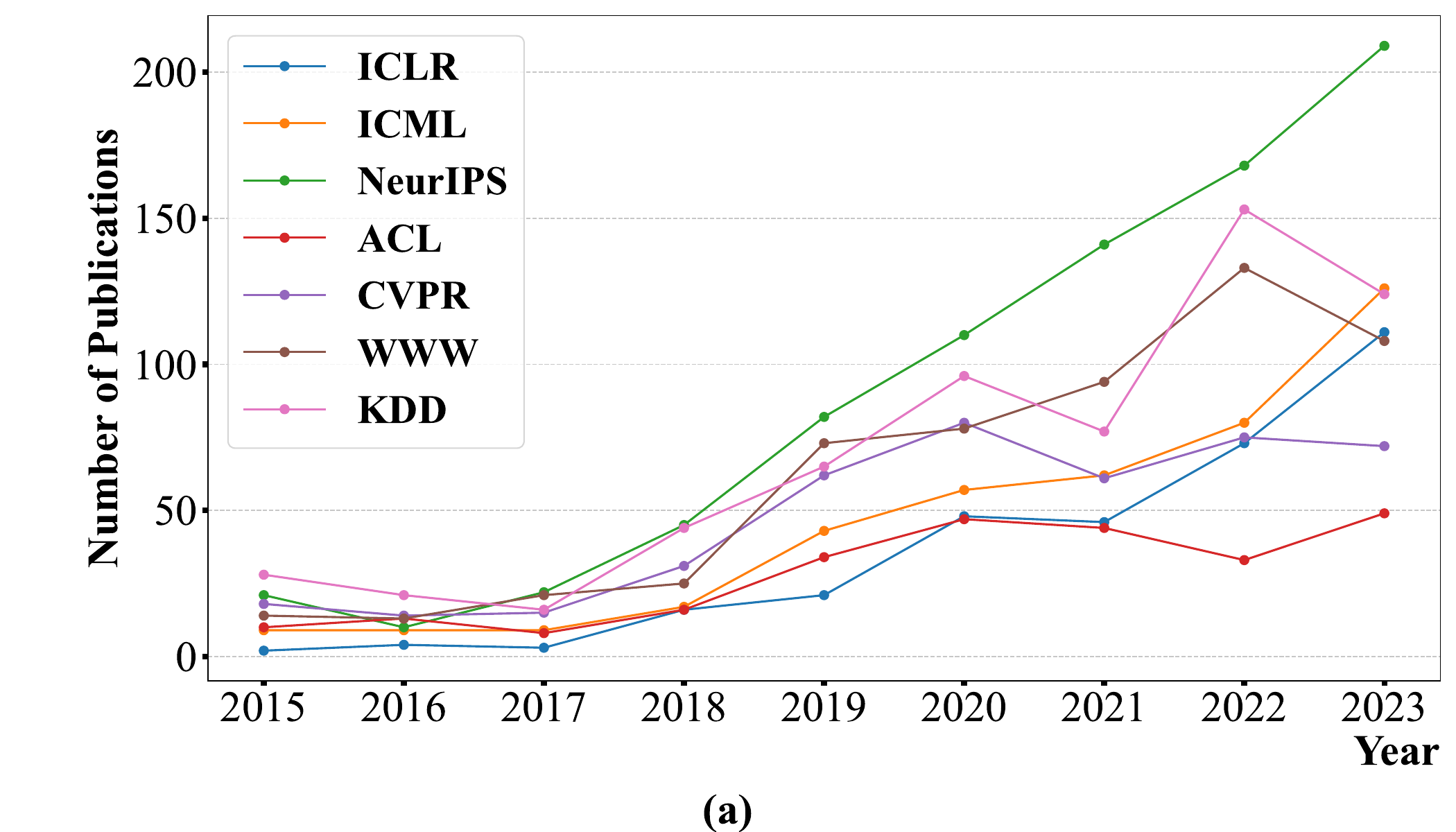}
    \includegraphics[width=0.40\textwidth]{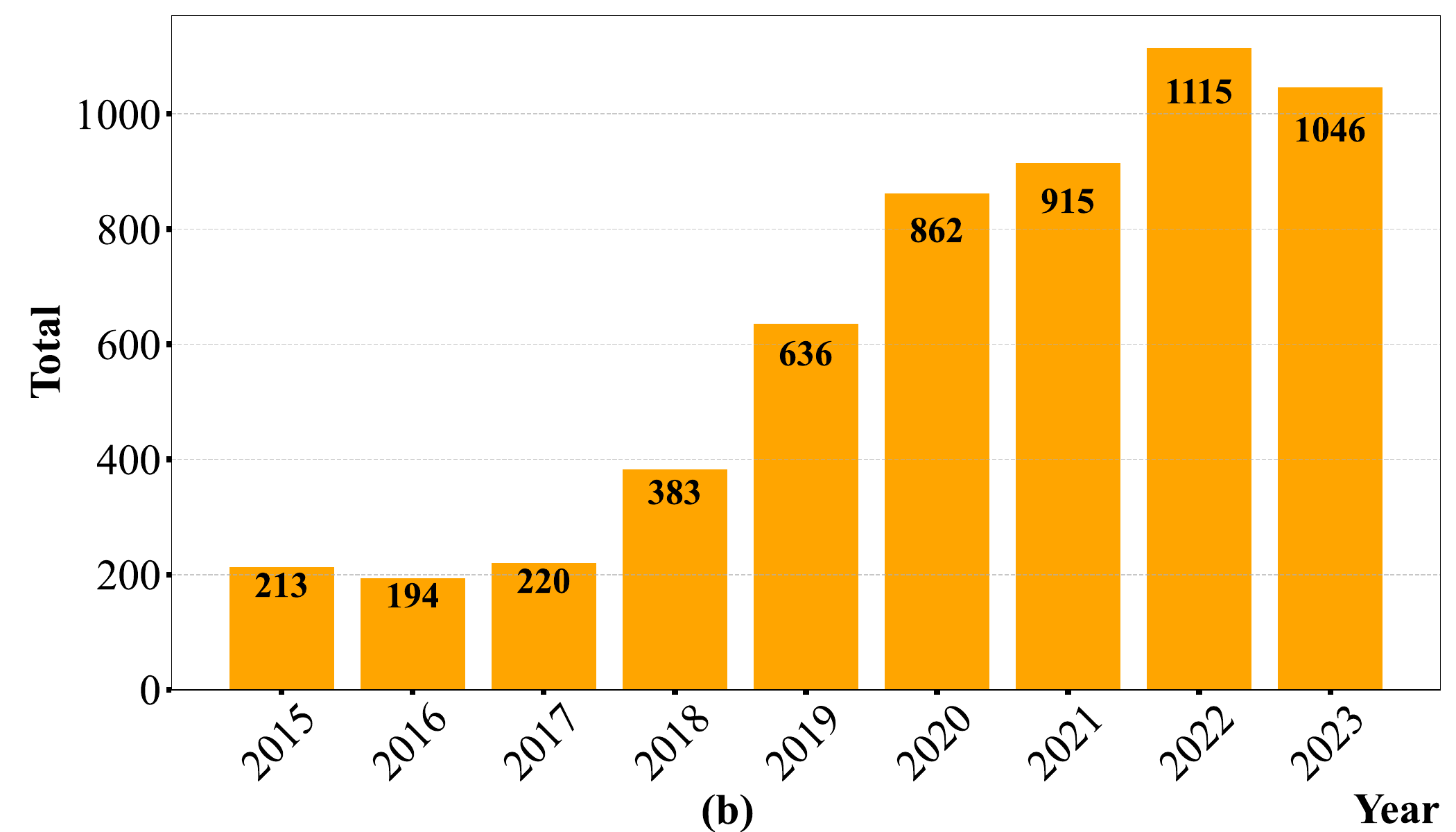}
     \caption{Number of GL publications in recent years on top venues. Figures (a) and (b) depict the number of GL publications at top conferences and the total number, respectively. }
    \label{fig:paper_number}
\end{figure}

\IEEEpubidadjcol

Representation learning provides a successful way to deal with various kinds of complex problems in other traditional scenarios, which also sheds light on the progress of graph learning. 
As shown in Figure \ref{fig:paper_number}, the number of graph learning works has increased rapidly in recent years, reaching a new peak of 1115 papers in 2022 on top venues. There are thousands of works that focus on different objectives, methods, and applications of GL,  making it imperative to survey these research and foster future progress on graph learning.
% making it crucial to survey these researches and promote future progress on graph learning.

\begin{table*}[t]
\caption{Comparison between our work and previous works}
\label{tab:comparison}
\newcommand{\tabincell}[3]{\begin{tabular}{@{}#1@{}}#2\end{tabular}}
\resizebox{\linewidth}{!}{
\begin{tabular}{l|ll}
\toprule
\textbf{Paper}                                                 & \textbf{Comparison}                                                                                                                                                                                                                         &  \\ \toprule \toprule
Goyal et al. \cite{goyal2018graph}   & \tabincell{l}{This work provides a new taxonomy
focusing on graph embedding methodologies. 
Precisely, it categorizes contemporary \\graph embedding methods into four primary classes: factorization, random walk, deep learning, and miscellaneous. 
Our \\survey offers a thorough and expansive examination that encompasses a wider array of techniques, encompassing both \\embedding and non-embedding methods. 
Besides, it addresses recent advancements in the amalgamation of LLMs and GL.
} & \\
% that concentrates on graph embedding techniques. Specifically, it divides current graph embedding \\ methods into factorization, random walk, deep learning and miscellaneous.  In our work, we provide a more comprehensive review that \\ considers more types of techniques, including embedding methods and non-embedding methods as well as latest progress on combining \\ Large language models and graph learning.}                                                                               &  \\ 
\midrule
% Zhou et al. \cite{Zhou2020Graph}     & \tabincell{l}{Zhou et al. concentrate on graph neural network models and provide a new taxonomy mainly from the view of propagation process of GNNs, including aggregation, updating, skip connection and so on. They also review the applications of GNNs. In this paper, we not only consider GNNs but also survey other traditional methods that help readers to have a comprehensive mind of past and current GL. }

Zhou et al. \cite{Zhou2020Graph}     & \tabincell{l}{
Zhou et al. focus their investigation on GNN models, presenting a taxonomy primarily centered on the propagation \\processes inherent in GNNs.
This taxonomy encompasses aspects such as aggregation, updating mechanisms, skip \\connections, and related elements. 
Additionally, they conduct a comprehensive review of the applications of GNNs. 
In \\this study, we extend our examination beyond GNNs, incorporating an exploration of traditional methods. 
This inclusive \\approach aims to provide readers with a holistic understanding of the historical and contemporary landscape of GL.
}

&  \\ \midrule
Wu et al. \cite{Wu2020Comprehensive} & \tabincell{l}{
% Wu et al. also propose a new taxonomy that divides current graph neural networks into Recurrent Graph Neural Networks, Convolution Graph Neural Networks, Graph Autoencoders and Spatial-temporal Graph Neural Networks. They introduce representative works of each types of GNNs with detailed description. In this work, we survey latest work on both GNNs and non-GNN methods. What's \\ more, we divide past works based on different types of used data structure, which is more intuitive for understanding various models.

Wu et al. categorize contemporary GNNs into Recurrent Graph Neural Networks, Convolution Graph Neural Networks, \\Graph Autoencoders, and Spatial-temporal Graph Neural Networks. 
The authors elaborate on representative instances \\within each GNN category, furnishing detailed descriptions.
In this study, we extend our inquiry to encompass recent \\developments in both GNNs and non-GNN methods. Furthermore, we organize prior works based on distinct data \\structures employed, enhancing the clarity and intuitiveness of understanding diverse modeling approaches.

} &  \\ \midrule

Liu et al. \cite{liu2022graph}       & \tabincell{l}{Liu et al. provide unified problem formulations and definitions related to self-supervised learning (SSL) on graphs. \\
Additionally, the authors identify technical limitations and outline prospective directions for the advancement of SSL \\on graphs. 
In the present study, we undertake a comprehensive survey encompassing self-supervised learning, \\unsupervised learning, and supervised learning on graphs, thereby contributing to a comprehensive understanding of GL.

}  &  \\ \midrule

Ren et al. \cite{ren2024survey} & \tabincell{l}{
Ren et al. categorize current research on large language models (LLMs) in graph-related tasks into four main areas:\\ GNN as a prefix,  LLM as a prefix, LLM-Graph integration, and LLM-only approaches. They also outline several \\ promising future directions. In this paper, we present a thorough survey, offering the latest insights into both GL and \\LLMs as individual fields, as well as their integration. Furthermore, we detail the emerging area of LLMs with graph \\retrieval augmentation, which we identify as a particularly promising avenue for future research.}

&  \\ \bottomrule                                                                                                                              
\end{tabular}}
\end{table*}

\begin{figure*}[t]
\tikzstyle{descript} = [text = black,align=center, minimum height=1.8cm, align=center, outer sep=0pt,font = \footnotesize]
\tikzstyle{activity} =[align=center,outer sep=1pt]
\resizebox{1\textwidth}{!}{
\begin{tikzpicture}[very thick, black]
\small
%% Coordinates
\coordinate (O) at (-1,0); % Origin
\coordinate (P1) at (4,0);
\coordinate (P2) at (8,0);
\coordinate (P3) at (12,0);
\coordinate (F) at (22,0); %End

%% Filled regions
%\fill[color=ColorOne!20] rectangle (O) -- (P1) -- ($(P1)+(0,1)$) -- ($(O)+(0,1)$); % Studies
%\path [pattern color=ColorOne, pattern=north east lines, line width = 1pt, very thick] rectangle ($(O)+(0.5,0)$) -- ($(O)+(2,0)$) -- ($(O)+(2,1)$) -- ($(O)+(0.5,1)$); % Something else
%\fill[color=ColorTwo!20] rectangle (P1) -- (P2) -- ($(P2)+(0,1)$) -- ($(P1)+(0,1)$); % Work
%\shade[left color=ColorThree, right color=white] rectangle (P2) -- (P3) -- ($(P3)+(0,1)$) -- ($(P2)+(0,1)$); % Current work

%% Text inside filled regions
%\draw ($(P1)+(-2.5,0.5)$) node[activity,ColorOne] {Studies};
%\draw ($(P2)+(-2,0.5)$) node[activity,ColorTwo] {Work};
%\draw ($(P3)+(-2,0.5)$)  node[activity, ColorThree] {Current activity};

%% Description
%\node[descript,fill=ColorTwo!15,text=ColorTwo](D2) at ($(P2)+(-2,-2.5)$) {%
%	\textbf{Where}\\
%	Project\\
%	description};

%\node[descript,fill=ColorThree!15,text=ColorThree](D3) at ($(P3)+(-2,-2.5)$) {%
%	\textbf{Where}\\
%	Project\\
%	description};
	
%% Events
\coordinate (E1) at (0,0); 
\coordinate (E2) at (0.1,0); 
\coordinate (E3) at (0.4,0);
\coordinate (E4) at (0.9,0);
\coordinate (E5) at (1.55,0);
\coordinate (E6) at (1.65,0);
\coordinate (E7) at (3.4,0);
\coordinate (E8) at (3.5,0);
\coordinate (E9) at (4.6,0);
\coordinate (E10) at (5.2,0);
\coordinate (E11) at (5.9,0);
\coordinate (E12) at (7.2,0);
\coordinate (E13) at (7.5,0);
\coordinate (E14) at (8.2,0);
\coordinate (E15) at (8.8,0);
\coordinate (E16) at (9.7,0);
\coordinate (E17) at (10.8,0);
\coordinate (E18) at (11.5,0);
\coordinate (E19) at (12.2,0);
\coordinate (E20) at (12.5,0);
\coordinate (E21) at (14.2,0);
\coordinate (E22) at (14.6,0);
\coordinate (E23) at (16.2,0);
\coordinate (E24) at (16.5,0);
\coordinate (E25) at (16.7,0);
\coordinate (E28) at (18.6,0); 
\coordinate (E29) at (20.6,0); 
\coordinate (E30) at (21.2,0); 
\coordinate (E31) at (17.5,0); 
\coordinate (E32) at (20.35,0);

\coordinate (E26) at (0.8,0);
\coordinate (E27) at (1.1,0);

\draw[<-,thick,color=SeaGreen] ($(E1)+(0,0.2)$) -- ($(E1)+(0,2.3)$) node [above=0pt,align=center,SeaGreen] {hMETIS-Kway\cite{Karypis2000Multilevel}};

\draw[<-,thick,color=SeaGreen] ($(E2)+(0,-0.1)$) -- ($(E2)+(0,-1.2)$) node [above=-15pt,align=center,SeaGreen] {Mcpherson et al. \cite{Mcpherson2001Birds}};

\draw[<-,thick,color=SeaGreen] ($(E3)+(0,0.2)$) -- ($(E3)+(0,0.8)$) node [above=0pt,align=center,SeaGreen] {Kempe et al. \cite{Kempe2003Maximizing}};

\draw[<-,thick,color=red] ($(E26)+(0,0.2)$) -- ($(E26)+(0,1.5)$) node [above=0pt,align=center,red] {NN4G\cite{micheli2005new}};

\draw[<-,thick,color=SeaGreen] ($(E4)+(0,-0.1)$) -- ($(E4)+(0,-0.45)$) node [above=-15pt,align=center,SeaGreen] {SCAN\cite{Xu2007Scan}};

\draw[<-,thick,color=red] ($(E27)+(0,-0.1)$) -- ($(E27)+(0,-1.6)$) node [above=-15pt,align=center,red] {GNN\cite{Scarselli2008The}};

\draw[<-,thick,color=SeaGreen] ($(E5)+(0,0.2)$) -- ($(E5)+(0,2.8)$) node [above=0pt,align=center,SeaGreen] {PathSim\cite{Sun2011Pathsim}};

\draw[<-,thick,color=SeaGreen] ($(E6)+(0,-0.1)$) -- ($(E6)+(0,-0.9)$) node [above=-15pt,align=center,SeaGreen] {Mcauley and Leskovec \cite{Mcauley2012Learning}};

\draw[<-,thick,color=blue] ($(E7)+(0,0.2)$) -- ($(E7)+(0,2.5)$) node [above=0pt,align=center,blue] {PTE\cite{Tang2015Pte}};

\draw[<-,thick,color=blue] ($(E8)+(0,-0.1)$) -- ($(E8)+(0,-1.3)$) node [above=-15pt,align=center,blue] {GraRep\cite{Cao2015Grarep}};

\draw[<-,thick,color=blue] ($(E9)+(0,0.2)$) -- ($(E9)+(0,1.8)$) node [above=0pt,align=center,blue] {Planetoid\cite{Yang2016Revisiting}\\Node2vec\cite{Grover2016Node2vec}};

\draw[<-,thick,color=red] ($(E10)+(0,-0.1)$) -- ($(E10)+(0,-1)$) node [above=-15pt,align=center,red] {GCN\cite{Kipf2016Semi-supervised}};

% \draw[<-,thick,color=blue] ($(E11)+(0,0.2)$) -- ($(E11)+(0,2.8)$) node [above=0pt,align=center,blue] {Node2vec\cite{Grover2016Node2vec}};

\draw[<-,thick,color=red] ($(E11)+(0,0.2)$) -- ($(E11)+(0,2.8)$) node [above=0pt,align=center,red] {VGAE\cite{Kipf2016Variational}};

\draw[<-,thick,color=blue] ($(E12)+(0,0.2)$) -- ($(E12)+(0,2)$) node [above=0pt,align=center,blue] {Metapath2vec\cite{dong2017metapath2vec}};

\draw[<-,thick,color=blue] ($(E13)+(0,-0.1)$) -- ($(E13)+(0,-1.3)$) node [above=-15pt,align=center,blue] {GraphSAGE\cite{Hamilton2017Inductive}};

\draw[<-,thick,color=red] ($(E14)+(0,0.2)$) -- ($(E14)+(0,1.1)$) node [above=0pt,align=center,red] {FastGCN\cite{Chen2018Fastgcn}\\GAT\cite{Velickovic2018Graph}};

\draw[<-,thick,color=blue] ($(E15)+(0,-0.1)$) -- ($(E15)+(0,-0.8)$) node [above=-15pt,align=center,blue] {DHNE\cite{Tu2018Structural}};

\draw[<-,thick,color=blue] ($(E16)+(0,0.2)$) -- ($(E16)+(0,2.3)$) node [above=0pt,align=center,blue] {GANDLERL\cite{Zhou2018Density}};

\draw[<-,thick,color=red] ($(E17)+(0,0.2)$) -- ($(E17)+(0,1.4)$) node [above=0pt,align=center,red] {HAN\cite{Wang2019Heterogeneous}};

\draw[<-,thick,color=red] ($(E18)+(0,-0.1)$) -- ($(E18)+(0,-1.8)$) node [above=-15pt,align=center,red] {SAGPool\cite{Lee2019Self-Attention}};

\draw[<-,thick,color=red] ($(E19)+(0,0.2)$) -- ($(E19)+(0,1.9)$) node [above=0pt,align=center,red] {DySAT\cite{sankar2020dysat}};

\draw[<-,thick,color=red] ($(E20)+(0,-0.1)$) -- ($(E20)+(0,-1)$) node [above=-15pt,align=center,red] {H$_{2}$GCN\cite{zhu2020beyond} \\ HGT \cite{Hu2020Heterogeneous}};

\draw[<-,thick,color=red] ($(E21)+(0,0.2)$) -- ($(E21)+(0,0.8)$) node [above=0pt,align=center,red] {ST-GNN\cite{yang2021financial}};

\draw[<-,thick,color=red] ($(E22)+(0,-0.1)$) -- ($(E22)+(0,-1.5)$) node [above=-25pt,align=center,red] {SGL\cite{Wu2021Self}\\RLGN\cite{meirom2021controlling}};

\draw[<-,thick,color=red] ($(E23)+(0,0.2)$) -- ($(E23)+(0,1.5)$) node [above=0pt,align=center,red] {DUET\cite{chen2022think}};

\draw[<-,thick,color=red] ($(E24)+(0,0.2)$) -- ($(E24)+(0,1)$) node [above=0pt,align=center,red] {Keriven et al. \cite{keriven2022not}};

\draw[<-,thick,color=red] ($(E25)+(0,-0.1)$) -- ($(E25)+(0,-1)$) node [above=-25pt,align=center,red] {Sancus\cite{peng2022sancus}\\HyperSCI\cite{ma2022learning}};

\draw[<-,thick,color=brown] ($(E28)+(0,0.2)$) -- ($(E28)+(0,1.9)$) node [above=0pt,align=center,brown] {Graph-ToolFormer \cite{zhang2023graph}};

\draw[<-,thick,color=brown] ($(E29)+(0,0.2)$) -- ($(E29)+(0,1.5)$) node [above=0pt,align=center,brown] {GraphGPT\cite{tang2024graphgpt}};

\draw[<-,thick,color=brown] ($(E30)+(0,-0.1)$) -- ($(E30)+(0,-1.5)$) node [above=-15pt,align=center,brown] {GraphRAG\cite{edge2024local}};

\draw[<-,thick,color=brown] ($(E31)+(0,-0.1)$) -- ($(E31)+(0,-0.7)$) node [above=-13pt,align=center,brown] {Andrus et al. \cite{andrus2022enhanced}};

\draw[<-,thick,color=brown] ($(E32)+(0,-0.2)$) -- ($(E32)+(0,-1)$) node [above=-15pt,align=center,brown] {GoT\cite{besta2024graph}};

%\draw[<-,thick,color=purple] ($(E1)+(0,0.2)$) -- ($(E1)+(0,1.5)$) node [above=0pt,align=center,purple] {Unexpected\\event\cite{barranco2019heterophily}};
%\draw[<-,thick,color=orange] ($(E3)+(0,-0.1)$) -- ($(E3)+(0,-1.5)$) node [above=-25pt,align=center,orange] {Unexpected\\event\cite{barranco2019heterophily}};
%\draw [decorate,decoration={brace,amplitude=6pt}]($(E2)+(-1,1.2)$) -- ($(E2)+(0.5,1.2)$) node [black,midway,above=6pt] {Something};
%% Arrows
%\path[->,color=ColorTwo] ($(P2)+(-1,-0.1)$) edge [out=-90, in=130]  ($(D2)+(0,1)$);
%\path[->,color=ColorThree]($(P3)+(-1,-0.1)$)  edge [out=-70, in=90]  ($(D3)+(0,1)$);
%% Arrow
\draw[->] (O) -- (F);
%% Ticks
\foreach \x in {2,...,20,21}   %坐标轴标记点
\draw(\x cm,3pt) -- (\x cm,-1pt);   % 坐标轴标记点上下长度
%% Labels
\foreach \i \j in {2/2015,4/2016,6/2017,8/2018,10/2019,12/2020,14/2021,16/2022,18/2023,20/2024}{
	\draw (\i,0) node[below=3pt] {\j} ;
}
\end{tikzpicture}
}
\caption{Chronological overview of graph learning methods. Methods in \textcolor{SeaGreen}{green}, \textcolor{blue}{blue}, \textcolor{red}{red} and \textcolor{brown}{brown} are graph mining methods, graph embedding methods, graph neural network methods and LLM-GLs, respectively. }
\label{fig:Chronological-overview}
\end{figure*}
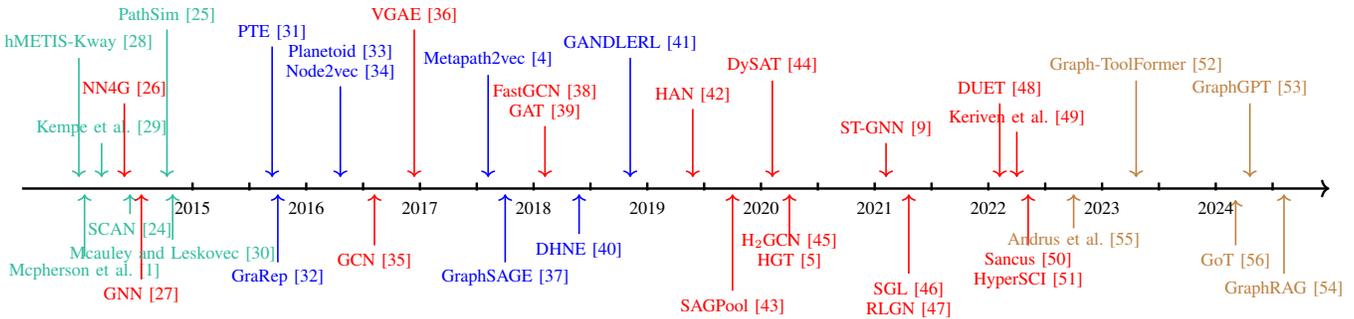

Figure \ref{fig:Chronological-overview} presents a chronological overview of influential GL methods, which can be classified into four main types: graph mining methods, graph representation methods, deep graph learning methods and LLM-GLs. In the early days of GL, most methods focused on characterizing graphs \cite{Mcpherson2001Birds} or utilizing graph structure information to perform downstream tasks on small graphs \cite{Xu2007Scan,Sun2011Pathsim}. Nowadays, representation learning on graphs is the dominant approach and can be broadly categorized into two types: graph embedding methods and graph neural network methods. Both methods aim to learn semantic representations of nodes, edges, or graphs. The former directly optimizes embeddings to preserve graph structure information, while the latter employs deep neural networks to model information passing processes on graphs, a technique that has been popular since early works \cite{micheli2005new,Scarselli2008The}.
% \cite{Micheli2009Neural}
% Due to the great achievements of LLMs, lots of researchers come into the area of LLMs, including traditional GL people. We will show the integration of LLMs and GL in Section \ref{section-llm_gl}.
Due to significant advancements in LLMs, many researchers, including those from the traditional GL community, have shifted their focus towards LLM-related research. In Section \ref{section-llm_gl}, we will explore the integration of LLMs and GL in detail.

%timetable

\subsection{Discussion of the Novelty}
While there have been some 
% notable and detailed 
remarkable and thorough surveys in the field \cite{Wu2020Comprehensive,Zhou2020Graph,goyal2018graph,liu2022graph}, there is still a need for a holistic survey that connects related objectives, methods, and tasks in an organic and logical manner. We compare our work with previous works in Table \ref{tab:comparison}.
Furthermore, there are hundreds of research papers on GL presented at top conferences every year, and their number continues to escalate rapidly. As a result of this rapid development, there is a lack of comprehensive surveys covering the latest applications, trends and challenges in the field, especially the combination of GL and large language models.

Therefore, we start with the graph structure, which makes it divergent from previous tabular data, by reviewing past and current works. 
Subsequently, we review the rapid development of GL, with the aim of promoting future research. 
Specifically, we summarize the technical trends of GL, including current pioneering works of large language models with graphs, mainstream applications, as well as benchmark datasets. Finally, based on the above, we discuss valuable future directions.

\subsection{Contribution}
\begin{figure*}[htb]
    \centering
    \includegraphics[width=0.7\textwidth]{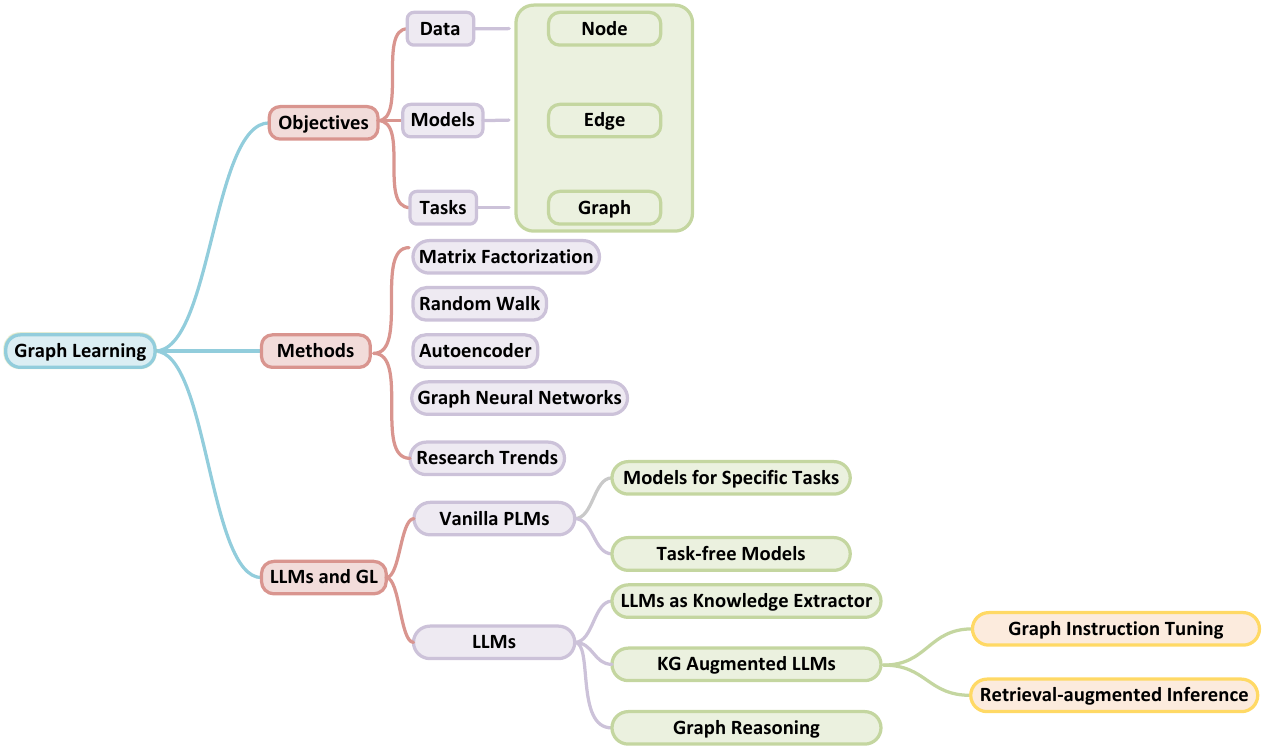}
	\caption{A taxonomy of graph learning by Objectives, Methods, Applications and Future Directions.}
    \label{figure-framework}
\end{figure*}

To fill the gap between rapid expansion of graph learning and the need of a comprehensive survey, we collect and sort a large number of work related to GL, including mentioned works in previous remarkable surveys \cite{Wu2020Comprehensive,Zhou2020Graph,goyal2018graph,liu2022graph,bacciu2020gentle,Cai2018comprehensive,xia2021graph,Zhang2020Deep,Shi2016survey,Wang2017Knowledge,Cui2018Survey} and recent two-years work on top conferences. We divide them into different types in terms of data, method, task, application. We choose representative work for each subsection considering content, year, citation. 
As depicted in Figure \ref{figure-framework}, this survey provides an intuitive taxonomy that considers the objectives of GL. Specifically, we sort previous works based on the elements of graphs, such as nodes, edges, and graph structures. Based on this taxonomy, we survey related methods and tasks on graphs and demonstrate the great performance of GL on various real-world applications. Finally, we discuss current trends and challenges in GL that are expected to further stimulate research.

The main contributions of this survey are summarized below.
\begin{itemize}
    \item 
    % We provide an organic taxonomy to survey previous researches with regard to two perspectives of goal and compoments. The former indicates data, models and tasks, and the latter refers to node, edge and graph structure.
    We propose a holistic classification to review previous research with respect to two dimensions of objective and constituents. The former denotes data, models and tasks, and the latter pertains to node, edge and graph structure.
    \item  We conduct a thorough review of pioneering research on the combination of pre-trained language models, particularly large language models, and graph learning with a new taxonomy.
    \item 
    % We summarize the current applications and widely used datasets of graph learning in the real world, which reflects the rapid development of this field. This highlights the importance of surveying the newest works as well.
    We summarize the current applications and prevalent datasets of graph learning in the real world, which manifests the rapid advancement of this field. This underscores the significance of reviewing the latest works as well.
    \item 
    % We discuss the current main trends and challenges of graph learning and aim to provide useful directions for future research.
    We explore the current salient trends and challenges of graph learning and aspire to offer valuable directions for future research.
\end{itemize}

% \subsection{Organization}
The remainder of this article is organized as follows. Section \ref{section-objective} proposes an intuitive taxonomy of previous works based on node, edge, and graph structure, from the viewpoint of data, model, and task. Section \ref{section-methods} demonstrates the main methods used for GL and current research trends.
Section \ref{section-llm_gl} reviews the latest advance on the combination of GL and LLMs.
Section \ref{section-applications} concludes the applications of GL in the real world. Finally, section \ref{section-future-directions} discusses the challenges of GL today.

\section{Objectives}
\label{section-objective}
Numerous methods have been proposed for mining complex information in graphs, focusing on various components. To categorize previous works, we divide them according to data, model, and task, and consider the basic components of graphs (i.e., nodes, edges, and graphs). This approach inherently links different methods that aim to achieve different objectives.
We provide a brief definition of the notations to be used in the following parts, as shown in Table \ref{tab:notations}. Other notations that can be inferred from the context are not listed here.

\begin{table}[htb]
\caption{Notations and Explanations}
\label{tab:notations}
\newcommand{\tabincell}[2]{\begin{tabular}{@{}#1@{}}#2\end{tabular}}
\begin{center}
\begin{tabular}{c l}
\toprule
\textbf{Notations}  & \textbf{Explanations}  \\ \midrule
$\Vcal$,$\Ecal$ & Node set and edge set\\
$v_i$, $e_j$ & A node $v_i \in \Vcal$ and edge $e_j \in \Ecal$\\
$G=(\Vcal,\Ecal)$ & A graph $G$ consists of $\Vcal$ and $\Ecal$\\
A & The adjacent matrix of $G$\\
D & The degree matrix of $G$\\
$I_N$ & identity matrix of N nodes \\
$\Tcal^v$,$\Tcal^e$ & The node type set and edge type set\\
$\Psi$,$\psi$ & Node type and edge type mapping function\\
\tabincell{c}{$X\in \Rbb^{\mid  \Vcal \mid \times d}$ }&  \tabincell{l}{Node feature matrix with $\mid \Vcal \mid$ nodes \\ and each node is with $d$ dimension} \\
$\Ncal(v_i)$ & The neighbours of node $v_i$ \\
$w_{i,j}$ & Edge weight between node $v_i$ and node $v_j$\\
$<h,r,t>$ & \tabincell{l}{Knowledge graph triplet, with head entity $h$,\\ relation $r$ and tail entity $t$}\\
% $hp$ & Hyperedge \\
$\T=\{t_1,t_2,\cdots,t_T\}$ & Timestamp set of a dynamic graph \\
$q^n=(u_n,v_n,r_n,t_n)$ &  \tabincell{l}{Event $q^n$ consists of linked nodes $v_i$, $v_j$ \\ and  relation $r_n$ in time $t_n$}\\
$h_i$ & Representation of node $v_i$\\
$\sigma$ & Activation function \\
$\text{LSTM}$ & Long Short Term Memory networks\\
 \bottomrule
\end{tabular}
\end{center}
\end{table}
% \begin{figure*}[htb]   
%     \centering
% 	\includegraphics[width=1\textwidth]{figures/FrameWork.pdf}
% 	\caption{A taxonomy of graph learning by Objective, methods and applications.}
% 	\label{FigureOne}
% \end{figure*}

% \begin{figure*}[htb]
%     \centering
%     \includegraphics[width=0.6\textwidth]{figures/FrameWork.pdf}
% 	\caption{A taxonomy of graph learning by Objective, methods and applications.}
%     \label{figure-framework}
% \end{figure*}

% \todo{add graph of Taxonomy framework}
\subsection{Data}
% \textcolor{red}{all the classifications are defined on the characters of the data used in papers}
In this subsection, we categorize current graphs based on their characteristics regarding node, edge, and graph structure. Note that all classifications are based on the data used in these works.
In the preliminary stages, researchers may only be able to work with native graph structures, even if they encounter a complex problem, such as treating dynamic graphs as static graphs. However, in some scenarios, it may not be necessary to use advanced methods. For instance, methods that target complex graph structures may result in an unaffordable computing cost.
\subsubsection{Node Perspective}

% \noindent \textbf{Homophily Graph vs Heterophily Graph}

% (1)hhjk
% \cite{Wan2019Your} nnnnnn
% (2)hjjkkk
% \cite{Velickovic2018Graph} enable (implicitly) specifying different weights to
% different nodes in a neighborhood。

\noindent \textbf{\textit{i. Homophily vs. Heterophily}}

% (1) Homophily Graph

Homophily, denoting the tendency for interconnected nodes to share the same class or exhibit similar characteristics, is a prevalent phenomenon in various types of networks, as noted in studies such as McPherson et al.'s exploration of homophily in social and academic networks \cite{Mcpherson2001Birds}. This homophilous nature proves advantageous for diverse tasks. For instance, Kossinets and Watts \cite{kossinets2009origins} delve into the origins of homophily within a large university community, establishing that the dynamic interplay of choice homophily can amplify even modest preferences for similar others, thereby generating discernible patterns of observed homophily. Tang et al. \cite{tang2013exploiting} contribute to the understanding of the homophily effect by employing homophily regularization in low-rank matrix factorization, applying it specifically to the trust prediction problem.

In real-world scenarios, there exist contexts where the principle of "opposites attract" prevails, leading to heterophily networks. Heterophily, characterized by connected nodes originating from different categories or possessing distinct characteristics, is observed in various domains, such as heterosexual dating networks and ecological food webs. Lazarsfeld and Merton \cite{barranco2019heterophily} succinctly define heterophily as the "mirror opposite of homophily." Nevertheless, real-world graph structures commonly manifest a combination of both homophily and heterophily. Notably, prevailing models often exhibit superior performance on graphs characterized by high homophily.

Addressing this challenge entails two primary approaches. One avenue involves the design of specialized models tailored to handle heterophily in graphs. For example, Chanpuriya and Musco \cite{chanpuriya2022simplified} propose a versatile model capable of adapting to both homophilous and heterophilous graph structures. Zhu et al. \cite{zhu2020beyond} introduce $H_2GCN$, specifically designed to effectively accommodate both heterophily and homophily. Another approach involves transforming heterophily graphs into homophily graphs. Suresh et al. \cite{suresh2021breaking} achieve this by converting the input graph into a computation graph with enhanced assortativity. Additionally, they emphasize the adaptive selection between structure and proximity when addressing diverse mixing patterns in graphs. \\

\noindent \textbf{\textit{ii. Uni-typed Graph vs. Bi-typed Graph vs. Multi-typed Graph}}
\begin{definition}
\label{definition-uni-typed-graph}
\textbf{Uni-typed graph \& Bi-typed graph \& Multi-typed graph}
A uni-typed graph can be defined as $G=(\Vcal, \Ecal, \Tcal^v, \Tcal^e)$. Here, $\Vcal=\{v_{i}\}_{i=1}^N$ denotes the set of nodes and $\Ecal=\{e_j\}_{j=1}^E$ denotes the set of edges. 
There are node type mapping function $\Psi(v): \Vcal \rightarrow \Tcal^v$ and 
edge type mapping function $\psi(e): \Ecal \rightarrow \Tcal^e$, where $\mid \Tcal^v \mid =1$ and $ \mid \Tcal^e \mid \in \Zbb^{+}$. For a bi-typed graph, the difference is that $\mid \Tcal^v \mid =2$. Fora multi-typed graph, the difference is that $\mid \Tcal^v \mid >2$.
\end{definition}

% \begin{definition}
% \label{definition-bi-typed-graph}
% \textbf{Bi-typed graph}
% A bi-typed graph can be defined as $\G=(\Vcal, \Ecal)$. Here, $\Vcal$ denotes the set of nodes and $\Ecal$ denotes the set of edges. $\Vcal=\{v_{1},v_{2},\cdots,v_{N}\}$ and $\Ecal=\{e_{1},e_{2},\cdots,e_E\}$.
% Node type mapping function $\Psi(v_i)\in \Tcal^v$
% Edge type mapping function $\psi(e_j)\in \Tcal^e$. $\mid \Tcal^v \mid =2$, $ \mid \Tcal^e \mid \in \Rbb$
% \end{definition}

% \begin{definition}
% \label{definition-multi-typed-graph}
% \textbf{Multi-typed graph}
% A multi-typed graph can be defined as $\G=(\Vcal, \Ecal)$. Here, $\Vcal$ denotes the set of nodes and $\Ecal$ denotes the set of edges. $\Vcal=\{v_{1},v_{2},\cdots,v_{N}\}$ and $\Ecal=\{e_{1},e_{2},\cdots,e_E\}$.
% Node type mapping function $\Psi(v_i)\in \Tcal^v$
% Edge type mapping function $\psi(e_j)\in \Tcal^e$. $\mid \Tcal^v \mid \geq 3$, $ \mid \Tcal^e \mid \in \Rbb$
% \end{definition}

According to the number of node types in graphs (see Definition \ref{definition-uni-typed-graph}), we can divide them into three types (i.e., uni-typed graph, bi-typed graph and multi-typed graph).
%(1) Uni-typed Graph

% As Definition \ref{definition-uni-typed-graph} shows, 

% There exists only one type of node in uni-typed graphs, which is one of the simplest types of graph. Uni-typed graphs helps to explore various problems on graph in a basic and reliable setting.
% For example,
% Page et al. \cite{Page1999Pagerank} describe PageRank, a method rating web pages objectively and mechanically, which effectively measures people's interest and attention to web pages.
% Inspired by this, a lot of incremental works improve the PageRank on different setting of graphs \cite{xing2004weighted,bahmani2010fast}.

Uni-typed graphs, characterized by the presence of a single node type, represent one of the simplest graph structures. This fundamental graph type serves as an advantageous framework for investigating diverse graph-related problems in a straightforward and dependable manner.
For instance, Page et al. \cite{Page1999Pagerank} introduced PageRank, a method for objectively and mechanically rating web pages. This algorithm effectively quantifies the level of interest and attention that users allocate to web pages. Building upon this seminal work, subsequent incremental advancements have been made to enhance the applicability of PageRank on various graph settings \cite{xing2004weighted, bahmani2010fast}.

Bi-typed graphs encompass graphs featuring two distinct node types, notably including bipartite graphs and bi-typed multi-relational heterogeneous graphs (BMHG), which hold significance in real-world applications. A bipartite graph is characterized by its ability to be decomposed into two disjoint vertex sets, where edges exclusively connect nodes from different types. Research by Li et al. \cite{li2019hierarchical} delves into the hierarchical representation learning of bipartite graphs.
In contrast to bipartite graphs, BMHG introduces intra-type relations, often overlooked and consequently compromising overall performance. To address this limitation, Zhao et al. \cite{zhao2022learning} propose DHAN, which employs both intra-type and inter-type attention-based encoders within a hierarchical mechanism to effectively model BMHG. Additionally, Zhao et al. \cite{zhao2022stock} apply the BMHG framework to model a Market Knowledge Graph for stock prediction. These endeavors contribute to a nuanced understanding and improved modeling of complex relationships within bi-typed graphs.

A multi-typed graph, characterized by the presence of three or more node types, typically exhibits a rich array of relationships among nodes. Modeling such graphs involves various strategies, one of which is the direct learning of relationships among nodes. Notably, Hu et al. \cite{Hu2020Heterogeneous} introduce HGT, leveraging a transformer-style attention mechanism \cite{vaswani2017attention} within a graph neural network to effectively model multiple heterogeneous relationships on web-scale graphs.
An alternative approach involves the utilization of meta-paths, which represent combinations of several types of relationships.
% Meta-paths are widely favored for their flexibility and interpretability, enabling the capture of complex semantic relations among objects. 
For instance, Dong et al. \cite{dong2017metapath2vec} contribute two scalable representation learning methods, named metapath2vec and metapath2vec++. Additionally, the automated discovery of meta-paths is recognized as a means to alleviate the expertise requirement \cite{Wan2020Reinforcement}.

\subsubsection{Edge Perspective}
% \noindent \textbf{\textit{A. Pair-wise Edge Graph}}

% \item \textbf{homogeneous graph}
% \item \textbf{multi-relation graph}
\noindent  \textit{\textbf{i. Unweighted Graph vs. Weighted Graph } }

Graphs can be distinguished into weighted and unweighted graphs based on the equality of weights assigned to their edges.
Graphs characterized by equal edge weights are termed unweighted graphs, a common occurrence in various applications \cite{Miller2013Parallel,Ellens2013Graph}. Modeling unweighted graphs is relatively straightforward, as demonstrated by Baswana et al. \cite{Baswana2006Approximate}, who construct approximate distance oracles in expected $O(n^2)$ time for unweighted graphs. Numerous methodologies, aimed at diverse tasks, rely on unweighted graphs, such as the improved parallel algorithm presented by Miller et al. \cite{Miller2013Parallel} for decomposing an undirected unweighted graph into components with small diameters.

Conversely, graphs with distinct edge weights, as observed in social networks where weights signify closeness \cite{Wang2022Minority} or in e-commerce networks where weights represent ratings \cite{Zhou2021Core}, provide additional information. Analyzing weighted graphs requires specialized design and algorithms, including the consideration of weights in adjacency matrix decomposition \cite{Toivonen2011Compression,Dhillon2007Weighted}. For instance, Umeyama \cite{Umeyama1988Eigendecomposition} discusses eigen decomposition of adjacency matrices for solving the weighted graph matching problem. Zhou et al. \cite{Zhou2021Core} introduce a novel definition of weighted coreness for vertices in a weighted graph and propose efficient algorithms for both weighted core decomposition and weighted core maintenance problems.

\noindent  \textit{\textbf{ii. Indirected Graph vs. Directed Graph }}

Indirected graphs are characterized by the presence of two symmetrical links for any types of edges, a configuration frequently encountered in applications such as community detection \cite{MarkNewman2004FindingAE} and natural language processing \cite{Peng2018Large}.
In contrast to indirected graphs, directed graphs depict asymmetric relationships between nodes, rendering them more versatile and applicable across various domains. For example, Shi et al. \cite{Shi2019Skeleton} represent skeleton data as a directed acyclic graph, leveraging the kinematic dependency between joints and bones in the natural human body. This directional representation proves indispensable for addressing complexities that cannot be adequately resolved using an indirected graph.

\begin{figure}[H]
    \centering
    \includegraphics[width=0.45\textwidth]{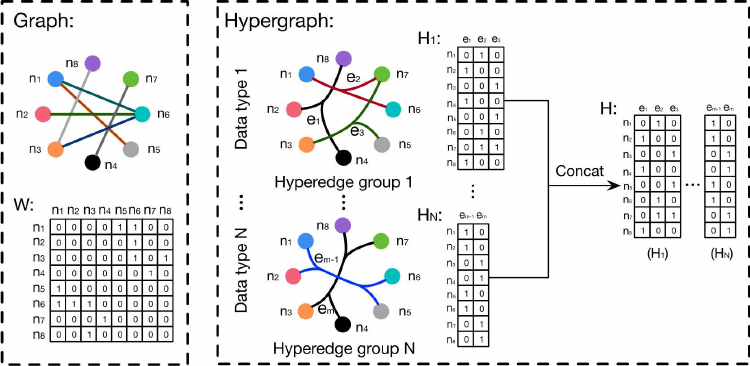}
     \caption{The comparison between graph and hypergraph \cite{Feng2019Hypergraph}}
    \label{fig:hypergraph}
\end{figure}

\begin{definition}
\label{definition-Homogeneous hypergraph}
\textbf{Homogeneous Hypergraph}. 
A homogeneous hypergraph can be defined as $ G=(\Vcal, \Ecal)$. Here, $v \in \Vcal$ denotes a node and 
$e \in \Ecal$ denotes a hyperedge. The relationship between nodes can be represented by an incidence matrix $\H \in \Rbb^{|\Vcal| \times |\Ecal|}$ with elements defined as:
 \begin{equation}
\H (v,e)=\left\{
             \begin{array}{lr}
             1, & \text{if}\  v \in e\\
             0, & \text{otherwise} \ .
             
             \end{array}
\right.
\end{equation}
\end{definition}

% \textbf{hypergraph}
\begin{definition}
\label{definiton-Heterogeneous hypergraph}
\textbf{Heterogeneous Hypergraph}. 
A heterogeneous hypergraph can be defined as $G=(\Vcal, \Ecal, \Tcal)$. 
Here, $v \in \Vcal$ denotes a node and $e \in \Ecal$ denotes an edge.
Besides, a heterogeneous hypergraph is associated with a hyperedge mapping function $\psi(e)$: $\Ecal \rightarrow \Tcal$, where $\Tcal$ denotes the set of hyperedge types, and $|\Tcal|>1$.
\end{definition}

\noindent \textit{\textbf{iii. Homogeneous Hypergraph vs. Heterogeneous Hypergraph} }

Hypergraphs are a graph structure wherein each edge can connect multiple nodes. The foundation of hypergraph theory can be traced back to \cite{Berge1984Hypergraphs}, primarily concentrating on hypergraph segmentation and the generalization of conventional graph theories to hypergraphs. Subsequently, research on hypergraphs has expanded to encompass a broader spectrum of tasks and methodologies. For instance, Sun et al. \cite{Sun2008Hypergraph} introduce a hypergraph spectral learning formulation tailored for multi-label classification, leveraging hypergraphs to capture correlation information among diverse labels.

In current research, there is a growing introduction of deep learning methods for hypergraph representation. Feng et al. \cite{Feng2019Hypergraph} propose a hyperedge convolution layer, expressed as:

\begin{equation}
 \mathbf{X}^{(l+1)}=\sigma\left(\mathbf{D}_v^{-1 / 2} \mathbf{H} \mathbf{W} \mathbf{D}_e^{-1} \mathbf{H}^{\top} \mathbf{D}_v^{-1 / 2} \mathbf{X}^{(l)} \Theta^{(l)}\right)
\end{equation}

where $\mathbf{H}$ represents the hypergraph adjacency matrix. $\mathbf{X}^{(l)}$ and $\mathbf{X}^{(l+1)}$ denote the learned representations of nodes in the $l$-th layer and $l+1$-th layer, respectively. Additionally, $\mathbf{D}_e$ and $\mathbf{D}_v$ serve as hyperedge degree and node degree matrices, functioning as normalization terms. The matrices $\Theta$ and $\mathbf{W}$ are trainable parameters.

Prior research has predominantly focused on homogeneous hypergraphs (refer to Definition \ref{definition-Homogeneous hypergraph}), often overlooking the heterogeneity inherent in hypergraphs, characterized by the presence of different types of hyperedges. Recent studies have recognized the significance of addressing this heterogeneity and propose models for heterogeneous hypergraphs (refer to Definition \ref{definiton-Heterogeneous hypergraph}).
For instance, Sun et al. \cite{Sun2021Heterogeneous} introduce a novel approach by initially projecting the heterogeneous hypergraph into a sequence of snapshots. Subsequently, they employ a wavelet basis with an efficient polynomial approximation to perform localized hypergraph convolution. The consideration of heterogeneity in hypergraphs proves essential for addressing various applications, as exemplified in the study by St-Onge et al. \cite{St2022Influential} focusing on the investigation of contagions.

\subsubsection{Graph Perspective}
% \begin{figure}[t]
%     \centering
%     \includegraphics[width=0.48\textwidth]{figures/xxx.pdf}
%     \caption{xxxx}
%     \label{fig:xxx}
% \end{figure}

\noindent \textit{\textbf{i. Homogeneous Graph vs. Heterogeneous Graph}}

\begin{definition}
\label{definition-homogeneous-graph}
\textbf{Homogeneous graph \& Heterogeneous graph}
A homogeneous graph can be defined as $G=(\Vcal, \Ecal, \Tcal^v, \Tcal^e)$. Here, $v \in \Vcal$ denotes a node and $e \in \Ecal$ denotes an edge. There are associated node 
type mapping function $\Psi(v): \Vcal \rightarrow \Tcal^v$ and
edge type mapping function $\psi(e): \Ecal \rightarrow \Tcal^e$, where $\mid \Tcal^v \mid =1$ and $ \mid \Tcal^e \mid = 1$.
For a heterogeneous graph, the difference is that $\mid \Tcal^v \mid + \mid \Tcal^e \mid > 2$.
\end{definition}

We formally define homogeneous graphs in Definition \ref{definition-homogeneous-graph}, characterized by the presence of a single node type and a single relationship type. Homogeneous graphs represent the most basic and extensively studied graph type. In contemporary research, a predominant focus has shifted towards employing deep learning techniques as opposed to traditional data mining methods. Notably, Graph Convolutional Networks (GCN) \cite{Kipf2016Semi-supervised} has emerged as one of the most successful models for handling homogeneous graphs, utilizing the adjacency matrix to perform convolution on a given graph. The message propagation process in GCN is expressed as:

\begin{equation}
    H^{(L+1)} = \sigma \big (\tilde{D}^{-\frac{1}{2}} \tilde{A} \tilde{D}^{-\frac{1}{2}} H^{(l)} W^{(l)}\big),
\end{equation}

where $\tilde{A} = A+ I_N$ denotes the adjacency matrix with self-connections, $\tilde{D}_{ii}=\sum_j \tilde{A}_{ij}$, and $W^{l}$ is a trainable weight matrix. Here, $H^{(l)}$ and $H^{(l+1)}$ represent the representations of nodes in the $l$-th and $l+1$-th layers, respectively. Inspired by GCN \cite{Kipf2016Semi-supervised}, Hamilton et al. \cite{Hamilton2017Inductive} present GraphSAGE, offering a scalable approach for semi-supervised learning on graph-structured data. Additionally, Velivckovic et al. \cite{Velickovic2018Graph} propose Graph Attention Networks (GAT), introducing different weights to neighbors' information during the message passing process. The weight learning and aggregation calculations are expressed as:

\begin{equation}
    \alpha_{i j}=\frac{\exp\left(\mathrm{LeakyReLU}\left(\vec{\bf a}^{T}[{\bf W}{\vec{h}}_i\vert\vert{\bf W}{\vec{h}}_{j}]\right)\right)}{\sum_{k\in N_{i}}\exp\left(\mathrm{LeakyReLU}\left(\vec{\bf a}^{T}[{\bf W}{\vec{h}}_i\vert \vert {\bf W}{\vec{h}}_{k}]\right)\right)},
\end{equation}

\begin{equation}
    \vec{h}_i^{\prime}=\sigma\left(\sum_{j \in \mathcal{N}_i} \alpha_{i j} \mathbf{W} \vec{h}_j\right),
\end{equation}

where $\vec{a}$ and $\bf W$ are the weight vector and trainable weight matrix, respectively. $\alpha_{ij}$ is the learned weight for the target node $i$ and neighbor node $j$.

However, real-world graphs are naturally modeled as heterogeneous graphs, as illustrated in Definition \ref{definition-homogeneous-graph}, wherein a graph contains more than one node type or relationship type. The presence of multiple node types, edges, and rich attribute information presents significant challenges in modeling heterogeneous graphs. Current methods addressing heterogeneous graphs can be broadly categorized based on their utilization of heterogeneous edges:

(1) \textit{Link-based Heterogeneous Graph Embedding:}
Schlichtkrull et al. \cite{Schlichtkrull2018Modeling} propose RGCN, which accounts for multiple relationships by assigning relation-specific transformation matrices in the aggregation process. The propagation model is expressed as:

\begin{equation}
    h_i^{(l+1)}=\sigma\left(\sum_{r \in \mathcal{R}} \sum_{j \in \mathcal{N}_i^r} \frac{1}{c_{i, r}} W_r^{(l)} h_j^{(l)}+W_0^{(l)} h_i^{(l)}\right),
\end{equation}

where $\mathcal{N}_i^r$ denotes the set of neighbors regarding node $i$ under relationship $r\in \Rcal$. $c_{i, r}$ is a normalization term that can be learned or set in advance.

Furthermore, Zhang et al. \cite{Zhang2019Heterogeneous} propose a comprehensive approach to capture heterogeneous information by incorporating special designs in sampling heterogeneous neighbors, encoding heterogeneous contents, and aggregating heterogeneous neighbors.

(2) \textit{Path-based Heterogeneous Network Representation:}
A meta-path is a path containing a sequence of relations defined on different types of objects. Wang et al. \cite{Wang2019Heterogeneous} advocate for utilizing a hierarchical attention mechanism to model path-based heterogeneous information. Specifically, they employ an attention mechanism to learn node-level representations for nodes under different relationships and aggregate the node-level representations with learned relation weights to obtain final semantic-level representations.

\noindent \textit{\textbf{ii. Static Graph vs. Dynamic Graph}}

% \begin{definition}
% \label{definition-static-graph}
% \textbf{Static graph.}
% A static graph can be defined as $\G_=(\Vcal, \Ecal)$. Here, $\Vcal=\{v_{1},v_{2},...\}$ denotes the set of nodes and $\Ecal=\{e_{1},e_{2},...\}$ denotes the set of edges.  and . Each node  $\v_i \in \Vcal$ and each edge $\e_i \in \Ecal$ only occurs once in $\G$.

% \end{definition}

\begin{definition}
\label{definition-DTDG}
\textbf{Discrete-time dynamic graph}. 
A discrete-time dynamic graph (DTDG) can be defined as a sequence: $\{G^1,G^2,\cdots,G^{T}\}$ in which ${G^t=\{\mathcal{V}^t,\mathcal{E}^{t}\}}$, $G^t$ is a snapshot in time \text{t}, $\mathcal{V}^t$ is the node set in $G^t$ and the $\mathcal{E}^{t}$ is the edge set in $G^t$.
\end{definition}

\begin{definition}
\label{definition-CTDG}
\textbf{Continuous-time dynamic graph}. 
A continuous-time dynamic graph (CTDG) can be defined as a sequence of event  quadruples $\{ q^1,q^2,...,q^N\}$, where $q^n=(u_n,v_n,r_n,t_n)$ means that node $u_n$ links to node $v_n$ with relation $r_n$ in time $t_n$. Each time $t_n$ is only assigned with corresponding event quadruple $q_n$. All nodes and links in different timestamps form a CTDG.
\end{definition}

\begin{definition}
\label{definition-dynamic-hypergraph}
\textbf{Dynamic hypergraph}. 
A Dynamic hypergraph can be defined as $G=(\Vcal, \Ecal, \Tcal)$. 
Here, $\Vcal$ denotes the set of nodes. 
$\Ecal = \{\Ecal^t\}_{t=1}^{T}$ denotes the sequence of hyperedge sets in different time and 
$\Ecal^t=\{e_1^t,e_2^t,\cdots,e_{E_t}^t\}$
denotes the set of hyperedge at time $t$. 
$\Tcal$ denotes the hyperedge type set, associated with a
hyperedge type mapping function  $\psi: \Ecal \rightarrow \Tcal$.
% Hyperedge time mapping function $\psi(e)$: $\Ecal \rightarrow \T=\{t_1,t_2,\cdots,t_T\}$.
\end{definition}

% \begin{definition}
% \label{Heterogeneous hypergraph}
% \textbf{Heterogeneous Hypergraph}. 
% A heterogeneous hypergraph can be defined as $\Gcal=(\Vcal, \Ecal, \Tcal)$. Here, $\Vcal$ denotes the set of nodes. 
% $\Ecal=\{hp_1,hp_2,...\}$
% denotes hyperedge set. $\Tcal_{hyper}=\{\Omega_1,\Omega_2,...,\Omega_M\}$ denotes hyperedge type set, and $|\Tcal_{hyper}|>1$ here. Hyperedge type map function $\psi$: $\psi(hp) \in \Tcal_{hyper}$.
% \end{definition}

Static graphs do not show any state changes over time.
Research related to static graphs can be divided into two categories based on the convolution method used:
\textit{(1) Spectral-based Graph Convolutional Networks:} 
Spectral convolution involves replacing the convolution kernel in the spectral domain with a diagonal matrix. For example, Graph Convolutional Networks (GCN) \cite{Kipf2016Semi-supervised} is essentially a localized first-order approximation of spectral graph convolution. 
\textit{(2) Spatial-based Graph Convolutional Networks:} 
The spatial-based approach characterizes graph convolution by considering the spatial relationships among nodes. Spatial-based Graph Convolutional Networks fundamentally propagate node information along the edges. In this context, Niepert et al. \cite{Niepert2016Learning} introduced a framework for acquiring convolutional neural networks tailored for arbitrary graphs. Their approach offers a general methodology for extracting locally connected regions within graphs.

Dynamic graphs pose increased complexity due to the inclusion of crucial temporal information, wherein links and nodes may manifest and vanish over time. The temporal granularity categorizes dynamic graphs into discrete dynamic graphs (DTDG, see Definition \ref{definition-DTDG}) and continuous dynamic graphs (CTDG, see Definition \ref{definition-CTDG}).

Discrete dynamic graphs are commonly characterized by multiple network snapshots captured at distinct time intervals. In this context, two distinct types of discrete dynamic graph neural networks (DGNN) for Discrete Time Dynamic Graphs (DTDG) can be identified.

\textit{(1) Stacked DGNNs} employ a Graph Neural Network (GNN) to model each snapshot, followed by the stacking of the outputs of previous GNNs into a time series model. For instance, Sankar et al. \cite{sankar2020dysat} introduced DySAT, a method for computing node representations through joint self-attention across two dimensions, namely structural neighborhood and temporal dynamics. The structural self-attention operation adopts a similar approach to Graph Attention Networks (GAT) \cite{Velickovic2018Graph}, facilitating the aggregation of neighbors' information. In the temporal self-attention operation, DySAT utilizes a transformer-style attention mechanism \cite{vaswani2017attention} to capture patterns in temporal changes. Specifically, it computes the output representation of node $v$ at time step $t$ as described by the following equation:

\begin{equation}
\begin{aligned}
Z_{v} &= \boldsymbol{\beta}_{v}\left(\boldsymbol{X}_{v} \boldsymbol{W}_{v}\right), \\
\beta_{v}^{i j} &= \frac{\exp \left(e_{v}^{i j}\right)}{\sum_{k=1}^{T} \exp \left(e_{v}^{i k}\right)}, \\
e_{v}^{i j} &= \frac{\left(\left(\boldsymbol{X}_{v} \boldsymbol{W}_{q}\right)\left(\boldsymbol{X}_{v} \boldsymbol{W}_{k}\right)^{T}\right)_{i j}}{\sqrt{F^{\boldsymbol{z}}}}+M_{i j},
\end{aligned}
\end{equation}

where $\boldsymbol{\beta}_{v} \in \mathbb{R}^{T \times T}$ represents the learned attention weight matrix, and $\boldsymbol{M} \in \mathbb{R}^{T \times T}$ is a mask matrix with each entry $M_{i j} \in\{-\infty, 0\}$ enforcing the auto-regressive property. In a recent study \cite{ye2022learning}, researchers explore methods to model the evolutionary and multi-scale interactions of time series data.

\textit{(2) Integrated DGNNs} integrate the GNNs and the time series model into a single layer, forming the encoder. Addressing the challenge of frequent changes in the node set, Pareja et al. \cite{pareja2020evolvegcn} introduce EvolveGCN, a method that adapts the Graph Convolutional Network (GCN) model along the temporal dimension without relying on node embeddings.

In reality, Discrete Time Dynamic Graphs (DTDG) can be viewed as a simplified manifestation of Continuous Time Dynamic Graphs (CTDG), as we can systematically convert a CTDG with fine granularity into graph snapshots with coarse granularity. Presently, three primary CTDG methodologies are available for continuous modeling.

\textit{(1) Time encoding methods} represent temporal information as a vector, akin to position embedding in language models \cite{gehring2017convolutional}. Previous approaches involved learning positional embeddings to capture the absolute position of nodes or edges. Currently, sinusoidal position encoding is commonly employed to represent the relative time distances \cite{Hu2020Heterogeneous}.

\textit{(2) Random process methods} can also be parameterized by neural networks. For instance, Trivedi et al. \cite{trivedi2019dyrep} propose a dynamic graph framework that frames representation learning as a latent mediation process, bridging two observed processes: the dynamics of the network (topological evolution) and dynamics on the network (activities between nodes).

\textit{(3) RNN-based methods} maintain node embeddings through Recurrent Neural Network (RNN)-based architectures. For example, DyGNN \cite{Ma2020Streaming} updates node information by capturing sequential information regarding edges, the time intervals between edges, and coherent information propagation. 
% DyGNN comprises two components: the update component and the propagation component. Both components include an interact unit, an update unit, and a merge unit, albeit with subtle distinctions. The update component refines the interaction information of involved nodes, aiming to capture the direct influence of an interaction. Subsequently, the propagation component disseminates interaction information to more related nodes, capturing indirect influence. The update unit employs an LSTM-style operation, considering the time-related weight decay.

Additionally, dynamic hypergraphs (refer to Definition \ref{definition-dynamic-hypergraph}) have garnered considerable attention in recent research. For instance, Zhang et al. \cite{Zhang2018Dynamic} introduced the first dynamic hypergraph structure learning method. This method aims to optimize both the label projection matrix (a common task in hypergraph learning) and the hypergraph structure concurrently. The objective function for dynamic hypergraph structure learning can be expressed as a dual optimization problem:

\begin{equation}
\begin{array}{c}
\begin{aligned}
\arg \min _{\mathbf{F}, 0 \preceq \mathbf{H} \preceq 1} \mathcal{Q}(\mathbf{F}, \mathbf{H}) & = \Psi(\mathbf{F}, \mathbf{H})+\beta \Omega(\mathbf{H})+\lambda \mathcal{R}_{\mathrm{emp}}(\mathbf{F})\\
& = \operatorname{tr}\Big (\left(\mathrm{I}-\mathbf{D}_{v}^{-\frac{1}{2}} \mathbf{H} \mathbf{W} \mathbf{D}_{e}^{-1} \mathbf{H}^{\mathrm{T}} \mathbf{D}_{v}^{-\frac{1}{2}}\right) \\ & \left(\mathbf{F} \mathbf{F}^{\mathrm{T}}+\beta \mathbf{X X}^{\mathrm{T}}\right) \Big ) + \lambda\|\mathbf{F}-\mathbf{Y}\|_{F}^{2}\ ,
\end{aligned}
\end{array}
\end{equation}

where $\beta$ and $\lambda$ are parameters that balance different components in the objective function. $\Psi(\mathbf{F}, \mathbf{H})$ serves as a commonly used hypergraph regularizer for the label projection matrix $\mathbf{F}$, emphasizing its smoothness on the hypergraph structure $\mathbf{H}$. $\Omega(\mathbf{H})$ imposes a constraint on $\mathbf{H}$ based on the input features $\mathbf{X}$. $\mathcal{R}_{\mathrm{emp}}(\mathbf{F})$ represents the empirical loss.

A limitation of existing methods is that the hypergraph structure remains fixed during training, lacking flexibility. Consequently, Jiang et al. \cite{Jiang2019Dynamic} propose DHGNN, comprising stacked layers of two modules: dynamic hypergraph construction and hypergraph convolution. Moreover, Zhu et al. \cite{zhu2020unsupervised} introduce a novel feature selection method involving the dynamic construction of a hypergraph Laplacian matrix within the framework of sparse feature selection.

\subsection{Models}

In this subsection, we categorize the output into two principal classes: non-embedding methods and embedding-based methods. The former pertains to task-specific output, exemplified by node similarity \cite{Jeh2002Simrankff}, node correspondences \cite{Koutra2013Big}, and local and global consistency \cite{Zhang2015Cosnet}. The latter encompasses representations of nodes, edges, graphs, and their hybrids, which can be employed for various downstream tasks.

% In this subsection, we generally divide the output into two typical categories: non-embedding methods and embedding-based methods. The former refers to task-specific output, such as node similarity \cite{Jeh2002Simrankff}, node correspondences\cite{Koutra2013Big}, and local and global consistency \cite{Zhang2015Cosnet}, 
% % link prediction \cite{Zhang2013predicting}, \cite{Kong2013Inferring} and node clustering \cite{Han2009mining}. 
% while the latter includes representations of nodes, edges, graphs, and the hybrid of these, which can be used for several downstream tasks.

\subsubsection{Non-Embedding Methods}
% \todo{Add non-embedding-based methods}

Before the advent of graph representation learning, numerous data mining methodologies were proposed to model diverse networks. for instance, Jeh and Widom \cite{Jeh2002Simrankff} employed a straightforward approach to learn the similarity of structural contexts. Koutra et al. \cite{Koutra2013Big} successfully identified node correspondences between two given graphs, particularly useful for aligning bipartite graphs. 
Sun et al. \cite{Sun2011Pathsim} introduced PathSim, a similarity search method designed for heterogeneous networks. In this context, the meta-path-based similarity measure, called PathSim, is constrained to symmetric meta-paths, reflecting the intuitive notion that similar peer objects should share comparable visibility, and the relation between them should be symmetric. For a given symmetric meta-path $P$, PathSim between two objects of the same type, denoted as $x$ and $y$, is computed using the following equation:

\small
\begin{equation}
    s(x, y)=\frac{2 \times\left|\left\{p_{x \rightsquigarrow y}: p_{x \rightsquigarrow y} \in \mathcal{P}\right\}\right|}{\left|\left\{p_{x \rightsquigarrow x}: p_{x \rightsquigarrow x} \in \mathcal{P}\right\}\right|+\left|\left\{p_{y \rightsquigarrow y}: p_{y \rightsquigarrow y} \in \mathcal{P}\right\}\right|} \ ,
\end{equation}
\small

where $p_{x \rightsquigarrow y}$ denotes a path instance between $x$ and $y$, and $p_{x \rightsquigarrow x}$.

\subsubsection{Embedding-based Methods}
In light of the popularity of representation learning, scholars are placing growing emphasis on acquiring semantic embeddings in graphs and leveraging them for diverse downstream tasks. In the subsequent subsections, we will delineate four distinct output types stemming from graph representation learning: node embedding, edge embedding, graph embedding, and hybrid embedding.

% embedding-based methods from four aspects according to the types of vector representations.
\noindent \textit{\textbf{i. Node Embedding}}

Node embedding stands out as a foundational output in graph representation learning, aiming to encapsulate both semantic information of nodes and the underlying graph structure. Its applicability extends across various structural scenarios, including social networks \cite{Zhang2017User} and academic networks \cite{gui2021pine}. 
% For instance, in social networks \cite{Zhang2017User}, the acquired node embeddings prove instrumental in predicting attributes like gender, education type, and identity of individuals. Similarly, in academic networks where nodes represent entities such as authors, papers, terms, and conferences, node embeddings find utility in classification tasks, such as forecasting the research area of a given author \cite{Hu2020Heterogeneous}.
Beyond the purview of structural scenarios, node embeddings find relevance in non-structural domains, exemplified by document networks \cite{Tang2015Pte}, chemistry \cite{Rong2020Self}, and computer vision \cite{Xu2017Scene}. In the context of a document network, nodes could represent words, sentences, paragraphs, or entire documents, each carrying distinct semantic information. Tang et al. \cite{Tang2015Pte} leverage both labeled and word information within a heterogeneous text network, constructing bipartite networks for words, documents, and labels. The embeddings are then learned by minimizing the sum of their respective objective functions. In the realm of computer vision, Xu et al. \cite{Xu2017Scene} employ a graph inference module on a detected set of objects within an image. Through iterative message passing on the constructed graph, incorporating edge and node features, the module learns node embeddings and edge embeddings. The resulting scene graph encompasses objects, object categories, and relationships between objects.
\noindent \textit{\textbf{ii. Edge Embedding}}

Edges represent another fundamental constituent of graphs, and acquiring embeddings for edges holds pivotal significance for edge-centric tasks while concurrently bolstering the efficacy of node-centric tasks.

Contemporary methodologies for learning edge embeddings often revolve around the relationships between connected nodes \cite{kim2019edge,jiang2020co,yasunaga2021qa}. For instance, QA-GNN \cite{yasunaga2021qa} initiates the process by learning relation embeddings based on the type embeddings associated with linked nodes. Subsequently, it computes messages between two nodes, integrating both node and relation embeddings.

A prevalent approach for learning edge embeddings from node embeddings involves constructing the line graph to reflect edge connections and updating edge embeddings based on the line graph's adjacency matrix. For example, CensNet \cite{jiang2020co} defines the edge adjacency matrix as follows:

\begin{equation}
    \tilde{A}_e=D_e^{-\frac{1}{2}}\left(A_e+I_{N_e}\right) D_e^{-\frac{1}{2}},
\end{equation}

Here, $D_e$ signifies the degree matrix of $A_e+I_{N_e}$. Subsequently, the propagation of edge features is executed as follows:

\begin{equation}
   H_e^{(l+1)}=\sigma\left(T^T \Phi\left(H_v^{(l)} P_v\right) T \odot \tilde{A}_e H_e^{(l)} W_e\right),
\end{equation}

where the matrix $T \in \Rcal^{N_v \times  N_e}$ denotes a binary transformation matrix, with $T_{i,m}$ representing the connection between edge $m$ and node $i$. Additionally, $P_v$ stands for the learned weight for nodes.

\noindent \textit{\textbf{iii. Graph Embedding}}

Graph embedding, distinct from substructure embedding, encapsulates the structural attributes of an entire graph within a latent vector space. Graph pooling stands out as a widely employed technique for acquiring graph embeddings, with two primary approaches: graph coarsening and node selection. In graph coarsening, nodes undergo clustering, followed by the synthesis of a super node to achieve pooling. Ying et al. \cite{Ying2018Hierarchical} introduce a graph pooling module proficient in generating hierarchical representations of graphs. On the other hand, node selection entails the identification of crucial nodes to substitute the original graph. Lee et al. \cite{Lee2019Self-Attention}, leveraging self-attention, employ the node selection pooling method to learn hierarchical representations with relatively few parameters. They initially compute self-attention scores using graph convolution, given by the expression: $\mathrm{idx}=\operatorname{top}-\operatorname{rank}(Z,\lceil k N\rceil)$, where $k\in(0,1]$ signifies the pooling ratio. Here, $\mathrm{idx}$ represents an indexing operation, and $Z_{\text {mask }}$ denotes the feature attention mask. The top $\lceil k N\rceil$ nodes are selected based on the self-attention score represented by $Z$, and $Z_{\text {mask }}$ corresponds to $Z_{\mathrm{idx}}$. The input graph undergoes the following operation: $X^{\prime}=X_{\mathrm{idx},:}, X_{out}=X^{\prime} \odot Z_{\text {mask }}$, where $X_{\mathrm{idx},:}$ denotes the row-wise indexed feature matrix.

Beyond graph pooling, additional research delves into direct graph representation learning methods. For instance, Pan et al. \cite{Pan2018Adversarially} present an adversarial graph embedding framework adept at encoding a graph into a compact representation.

\subsection{Tasks}
In this subsection, we summarize previous works based on their target tasks. Different tasks have different requirements for these methods, which results in significant differences between models.

\subsubsection{Node-based Tasks}
\noindent \textit{\textbf{i. Node Classification and Clustering}}

Node classification and clustering stand as foundational tasks in GL, typically executed based on node embeddings. In the realm of graph representation learning, these tasks are commonly treated as downstream experiments to assess the efficacy of models \cite{Kipf2016Semi-supervised,Park2020Unsupervised}. Conventional approaches predominantly leverage graph structure to accomplish such tasks \cite{Cao2015Grarep,Ribeiro2017Struc2vec}. However, augmenting the representation learning process with additional information can yield improved results. For example, Huang et al. \cite{Huang2017Label} incorporate label information to enhance the representation learning of attributed network embeddings. Sun et al. \cite{Sun2021Heterogeneous} take a different approach by transforming heterogeneous hypergraphs into a sequence of snapshots and applying localized hypergraph convolution to learn node embeddings. \\

\noindent \textit{\textbf{ii. Node Ranking}}

Node ranking constitutes a pivotal facet of graph analysis, holding significant implications for various domains such as viral marketing. Over the years, numerous effective methods have been introduced for node ranking, such as the renowned PageRank algorithm \cite{Page1999Pagerank}, initially devised for web page citation ranking and subsequently instrumental in shaping the Google search engine. Addressing the nuanced fairness concerns in GNNs, Dong et al. \cite{dong2021individual} present a node ranking framework designed to gauge individual fairness by assessing the similarity between node instances. Moreover, node ranking finds widespread application as a supportive tool for diverse tasks. Notably, Gilbert and Levchenko \cite{Gilbert2004Compressing} advocate for vertex ranking as a pivotal strategy in graph compression. \\

\subsubsection{Edge-based Tasks}

\noindent \textit{\textbf{Link Prediction.}} 
Link prediction seeks to unveil latent connections between nodes, a task of paramount importance with widespread applicability in real-world domains, including medicine and biology \cite{Ying2018Hierarchical}. Over the years, numerous enhanced methods for link prediction have been proposed. Clauset et al. \cite{Clauset2008Hierarchical} introduce a methodology leveraging network data to infer hierarchies within networks, replicating the topological properties and predicting missing links based on the properties of network hierarchies. Zhu et al. \cite{zhu2021neural} integrate the Bellman-Ford algorithm into the GNN framework to enhance link prediction tasks. Among various scenarios, social networks stand out as a well-known domain for link prediction applications. Xu et al. \cite{Xu2018Exploring} define two distinct types of links, learning two representations for each node, thereby aiding link prediction through the extraction of diverse semantic information pertaining to links within social networks.

\subsubsection{Graph-based Tasks}

\noindent \textit{\textbf{i. Community Detection}}

Community detection is a research area that focuses on identifying locally densely connected subgraphs by clustering nodes based on one or more features. This field has been extensively studied since Newman proposed the concept of modularity \cite{MarkNewman2004FindingAE}. One approach to modeling dynamic community structure is the classical stochastic block model-based optimization approach proposed by Anagnostopoulos et al. \cite{ArisAnagnostopoulos2016CommunityDO}. Another approach, proposed by Chen et al. \cite{Chen2017Supervised}, involves modifying GNNs by adding non-traceback operators, which allow them to exploit edge adjacency information for community detection. GNNs are also useful for detecting overlapping communities \cite{shchur2019overlapping}. This is particularly important because communities in real-world graphs are often not disjointed. 

\noindent \textit{\textbf{ii. Graph Classification}}

Graph classification is a technique that aggregates all node data into a vector representing the entire graph structure for classification, which is similar to node classification. Employing pooling is a common approach to preserving the structural information of the graph. For example, Hamilton et al. \cite{Hamilton2017Inductive} apply an element-wise max-pooling operation to aggregate information across neighbors. Ying et al. \cite{Ying2018Hierarchical} introduce a differentiable pooling method for GNNs to extract complex, hierarchical structures of real-world graphs. For multitask graph classification tasks, Pan et al. \cite{Pan2017Task} propose an algorithm that classifies each subgraph feature into one of three categories shared by specific tasks. 
% To prevent the loss of graph structure information during feature extraction by graph embedding methods, Peng et al. \cite{peng2020motif} presented a motif-based Generative Adversarial Network (GAN) for graph classification.

% \begin{itemize}
%     \item Information Diffusion
%     \item Graph/Network alignment
%     \item Anchor links prediction
%     \item Graph compression
%     \item \edit{Graph Summarization?}
%     \item Graph reconstruction 
%     \item network visualization
    
%     \item Network distance?
%     \item Network evolution
%     \item Graph Similarity Description
    
%     \item Graph pooling
    
% \end{itemize}   

\section{Methods}
\label{section-methods}
In this section, we categorize methods into two primary types: traditional models and graph neural networks. Traditional models can be further classified into three types: matrix factorization, random walk-based, and autoencoder-based models. Initially, we survey traditional models, some of which remain active or are integrated with GNNs in contemporary research, providing valuable insights for current models. GNNs typically demonstrate heightened expressivity and superior performance in comparison to traditional models. The section concludes by addressing two facets of GNNs.

\subsection{Matrix Factorization}

Factorization-based algorithms \cite{Ahmed2013Distributed, Cao2015Grarep, Ou2016Asymmetric} derive node embeddings through the decomposition of matrices measuring the probabilities of node connections. These matrices can manifest in various forms, including the node adjacency matrix, Laplace matrix, node transfer probability matrix, Katz similarity matrix, and others. For instance, Cao et al. introduce GraRep \cite{Cao2015Grarep} for learning node representations in weighted graphs. Their method integrates global structural information, encapsulating long-distance relationships and distinct connections between nodes. Drawing inspiration from the skip-gram model \cite{Mikolov2013Distributed}, they construct a probability transition matrix, ultimately quantifying relationships between vertices. Moreover, the effective and efficient implementation of factorization on these matrices is imperative. Ahmed et al. \cite{Ahmed2013Distributed} propose GF, the inaugural method to obtain a graph embedding in $O(|E|)$ time. GF achieves this by factorizing the adjacency matrix while minimizing the following loss function:

\begin{equation}
 \phi(Y, \lambda)=\frac{1}{2} \sum_{(i, j) \in E}\left(W_{i j}-<Y_i, Y_j>\right)^2+\frac{\lambda}{2} \sum_i\left\|Y_i\right\|^2 \ ,
 \end{equation}

where $ W_{i j}$ represents the edge weight between nodes $(i,j)$, replaceable as the inner product of the two nodes. $ Y_i, Y_j$ denote the representations of nodes $i$ and $j$, respectively, and $\lambda$ is the regularization coefficient. Ou et al. \cite{Ou2016Asymmetric} propose HOPE, perceived as an improvement on GF. The key idea of HOPE is to preserve asymmetric transitivity, capturing structural information in graphs. The approach introduces a class of high-order proximity measurements and then employs a time-efficient generalized SVD for their solution.

\subsection{Random Walks}
The random walk-based approach entails sampling a graph with an extensive array of paths, revealing the contextual information of connected vertices. These paths are generated by initiating walks from randomly selected initial nodes. Once the paths are established, a sequence of probabilistic models can be applied to these randomly sampled paths for learning node representations. Methods employing the random walk approach can be categorized into two groups: structure-based random walks and auxiliary information-based random walks.

% The random walk-based approach involves sampling a graph with a large number of paths. These paths indicate the context of the connected vertices and are obtained by starting the walk from randomly selected initial nodes.
% % These paths indicate the context of the connected vertices. The randomness of walking gives the ability to navigate the graph and capture global and local structural information by traversing adjacent vertices.
% Once the paths are built, a series of probabilistic models can be executed on these randomly sampled paths to learn the node representation. We can divide the methods based on the random walk approach into two categories: structure-based random walks and auxiliary information-based random walks.

\subsubsection{Structure-Based Random Walks}

In practical applications, numerous networks solely possess structural information while lacking vertex attribute data. The identification of graph structure details, such as pivotal nodes and latent connections, becomes challenging in the absence of vertex information. A proficient approach to address this challenge is through structure-based random walks.

Taking inspiration from Word2vec \cite{Mikolov2013Efficient}, Perozzi et al. \cite{Perozzi2014Deepwalk} introduce DeepWalk as a technique for capturing the relational links between nodes in a graph data structure. DeepWalk comprises two modules: random sampling on graphs to generate node sequences and training a skip-gram model for obtaining node embeddings.
To encompass the diversity of connection patterns in the network, Grover et al. \cite{Grover2016Node2vec} propose node2vec, which generates neighbor sequences by sampling nodes with varying probabilities. Specifically, it allows the sample strategy to be biased towards depth-first search or breadth-first search. Modern random walk algorithms typically incorporate a restart strategy. For instance, Zhang et al. \cite{Zhang2019Heterogeneous} employ a random walk with a restart strategy to sample heterogeneous neighbors.

\subsubsection{Auxiliary Information-based Random Walks}

Auxiliary information-based random walks are introduced to leverage the attribute information associated with nodes.
For instance, Yang et al. \cite{Yang2016Revisiting} present a semi-supervised graph learning framework tailored for nodes with attribute information. They initially sample pairs of instances and contexts using a random walk and subsequently formulate a loss function based on these pairs. The context is designed to jointly model both graph structure information and label information. Given that the predicted label is contingent on the input attributes of nodes, the sampled distribution encompasses both graph structure information and node attribute information.

\subsection{Autoencoder-based methods}

Graph Autoencoders (GAEs) employ an encoder to project the original graph into low-dimensional representations, which are then processed by a decoder for graph reconstruction. By minimizing the reconstruction loss, the learned representations encapsulate topological information, rendering them valuable for downstream tasks. For instance, Cao et al. \cite{Cao2016Deep} leverage stacked denoising autoencoders to encode and decode positive point-wise mutual information (PPMI) matrices using multilayer perceptrons. In contrast to previous approaches, Hamilton et al. \cite{Hamilton2017Inductive} utilize two graph convolutional layers to encode node features. Velickovic et al. \cite{Velickovic2019Deep} enhance local network embeddings to capture global structural information by maximizing local mutual information.
Rather than solely reconstructing the graph adjacency matrix, which may lead to overfitting, Kipf et al. \cite{Kipf2016Variational} propose VGAE, a variational version of VAE (Variational Autoencoder), to learn the data distribution. VGAE posits that the empirical distribution $q(Z|X,A)$ should closely align with the prior distribution $p(Z)$.

When dealing with multiple graphs, GAE can learn the distribution of graph generations by encoding the graph as a hidden representation and decoding the graph structure from a given hidden representation. Simonovsky and Komodakis \cite{Simonovsky2018Graphvae} introduce GraphVAE, which models the existence of nodes and edges as independent random variables. By assuming the posterior distribution $q_\phi{(z|G)}$ defined by an encoder and the generative distribution $p_\theta{(G|z)}$ defined by a decoder, GraphVAE optimizes the variational lower bound.

\begin{equation}
    L(\phi, \theta ; G)=E_{q_{\phi}(z \mid G)}\left[-\log p_{\theta}(G \mid \mathbf{z})\right]+K L\left[q_{\phi}(\mathbf{z} \mid G) \| p(\mathbf{z})\right],
\end{equation}

where $p(z)$ follows a Gaussian prior, and $\phi$ and $\theta$ are learnable parameters. Utilizing a Convolutional Graph Neural Network (ConvGNN) as the encoder and a simple multi-layer perceptron as the decoder, GraphVAE generates a graph complete with its adjacency matrix, node attributes, and edge attributes.

\subsection{Graph Neural Networks}
% \subsubsection{General Frameworks}
% message passing neural network
Graph Neural Networks (GNNs) leverage deep neural networks to capture message-passing dynamics on graphs, demonstrating remarkable efficacy across diverse tasks and contexts. Existing methodologies can be categorized into two primary domains: learning and inference. The former encompasses investigations into the incorporation of supervisory signals within GNN frameworks. The latter encompasses prior studies exploring the capacity of GNNs to infer unseen graphs.
% \subsubsection{Supervised vs. Semi-supervised methods vs. Unsupervised vs. Self-supervised methods}
\subsubsection{Learning Perspective}

\noindent \textit{\textbf{i. Supervised Methods.} }

Most Graph Neural Networks (GNNs) are trained through supervised signals, which encompass node labels \cite{Hu2020Heterogeneous, zhao2022fisrebp}, edge labels \cite{Zhou2018Density, zhao2022connecting}, and graph labels \cite{Pan2017Task, peng2020motif}, representing various properties of graphs. Analogous to conventional supervised learning, supervised GL employs cross-entropy and mean squared error (MSE) for classification and regression tasks. For instance, Peng et al. \cite{peng2020motif} introduce a motif-based attentional graph convolutional neural network for the classification of chemical compounds. Zhao et al. \cite{zhao2022fisrebp} construct an enterprise knowledge graph with nodes representing companies and persons, optimizing their proposed model with cross-entropy loss: $\mathcal{L}=-\sum_{i \in \mathcal{Y}_L} y_i \log \left(\tilde{y}_i\right)$. The labels indicate the state of the companies, distinguishing between bankruptcy and ongoing operations.

\noindent \textit{\textbf{ii. Semi-supervised Methods.} }

Semi-supervised learning concurrently employs both labeled and unlabeled data, proving particularly beneficial in scenarios where only a limited number of samples are labeled, and the majority remain unlabeled.
Hu et al. \cite{Hu2019Heterogeneous} introduce HGAT, a method designed for semi-supervised short text classification. HGAT efficiently utilizes sparse labeled data and abundant unlabeled data through information propagation across a graph. The approach offers a versatile framework for modeling short texts, enabling the incorporation of diverse additional information types and the capture of their relations to address semantic sparsity.
Graph-based Semi-Supervised Learning (SSL) seeks to extend labels from a small set of labeled data to a larger pool of unlabeled data using graph structures. Wan et al. \cite{Wan2021Contrastive} propose a SSL algorithm based on GCN to enhance supervision signals by leveraging both data similarities and graph structure.
Jiang et al. \cite{jiang2019semi} introduce GLCN, a method that combines a semi-supervised learning approach with a GCN model to perform tag classification.

\noindent \textit{\textbf{iii. Self-supervised Methods}}

Self-supervised learning methods primarily employ auxiliary tasks to extract supervised information from large-scale unsupervised data and subsequently train the network using this constructed supervised information. There are four main types of self-supervised methods suitable for graphs. The first type utilizes node feature similarity clustering to obtain pseudo tags and achieve self-supervised training. For instance, M3S \cite{Sun2020Multi} leverages the DeepCluster technique \cite{tian2017deepcluster}, a popular form of self-supervised learning. It introduces a corresponding alignment mechanism in the embedding space to refine the Multi-Stage Training Framework. The alignment mechanism aims to transform the categories in clustering into the classes in classification. Specifically, the aligned class is calculated as follows:

\begin{equation}
c^{(l)}=\underset{m}{\arg \min }\left\|v_l-\mu_m\right\|^2,
\end{equation}

where $\mu_m$ refers to the center of class $m$ in labeled data, and $v_l$ is the center of cluster $l$ in unlabeled data. The second type constructs a supervision signal based on the correlation within the input data or context connections to achieve self-supervised training. Wu et al. \cite{Wu2021Self} propose SGL, enhancing node representation learning by exploring internal relationships between nodes and building supervisory signals based on internal correlations of input data. The third involves masking the node features of the center node and reconstructing the node features of the center node using the features of its neighbors. The reconstruction error serves as the self-supervised loss function for model optimization and training \cite{Hu2020Gpt}. The fourth type combines multi-view learning, constructs the contrast loss function for self-supervised training on the graph. For instance, Wang et al. \cite{wang2021self} propose HeCo, which leverages network schema and meta-path structure as two different views for cross-view contrast learning. HeCo employs a view mask mechanism to extract positive and negative embeddings from both views, enabling cooperative and mutually supervised learning between the two views.

\noindent \textit{\textbf{iv. Unsupervised Methods}}

Unsupervised learning entails deriving statistical rules or the intrinsic structure of data from unlabeled data, employing techniques such as clustering, dimension reduction, and probability estimation. Unsupervised learning on graphs shares similarities with self-supervised learning on graphs, where models are optimized using cross-entropy loss, contrast loss, or their variants. For instance, Kipf et al. \cite{Kipf2018Neural} propose NRI, optimizing the reconstruction error and Kullback-Leibler divergence during the training process. Additionally, some models use combinatorial loss functions to optimize unsupervised graph learning. Tang et al. \cite{Tang2012Unsupervised} introduce LUFS, leveraging the concept of pseudo-class labels to guide unsupervised learning. The optimization of the model is defined by:

\begin{equation}
    \begin{aligned}
    \min _{\mathbf{W}} \operatorname{Tr}\left(\mathbf{W}^{\top} \mathbf{X} \mathbf{L} \mathbf{X}^{\top} \mathbf{W}\right)+\beta\|\mathbf{W}\|_{2,1} \\
+\alpha\operatorname{Tr}\left(\mathbf{W}^{\top} \mathbf{X}\left(\mathbf{I}_{n}-\mathbf{F F}^{\top}\right) \mathbf{X}^{\top} \mathbf{W}\right) \ ,
    \end{aligned}
\end{equation}

where $\mathbf{F}$ is the weighted social dimension indicator matrix. $\mathbf{X}=\left(\mathbf{x}_{1}, \mathbf{x}_{2}, \ldots, \mathbf{x}_{n}\right) \in \mathbb{R}^{m \times n}$ is the conventional representation of the node set $\mathbf{u}=\left\{u_{1}, u_{2}, \ldots, u_{n}\right\}$. The mapping matrix $\mathbf{W} \in \mathbb{R}^{m \times c}$ assigns each data point a pseudo-class label, where $c$ is the number of pseudo-class labels. The pseudo-class label indicator matrix is $\mathbf{Y}=\mathbf{W}^{\top} \operatorname{diag}(\mathbf{s}) \mathbf{X} \in \mathbb{R}^{c \times n}$.

\subsubsection{Inference Perspective}

% \textbf{Transductive learning} is the inference from specific data observed to specific data. Specifically, the model is trained with the training set data and the test set data and then tested with the test set data. Such as \cite{Mcpherson2001Birds},\cite{girvan2002community},\cite{Wu2021Self},\cite{Zhu2021Graph},etc.

\noindent \textit{\textbf{i. Transductive Methods}}

\begin{definition}
\textbf{Transductive GNNs} Given a graph $G=(\mathcal{V},\mathcal{E})$, label information $\mathcal{Y}$
and Transductive Graph Neural Networks (\text{TGNNs}), 
TGNNs' training and testing processes are all performed on $G = G_{train} = G_{test}$ with $\mathcal{Y}_{train}, \mathcal{Y}_{test} \subseteq \mathcal{Y}$ and $\mathcal{Y}_{train} \cap \mathcal{Y}_{test}=\emptyset$. Noted that the label information can be node-level, edge-level or graph-level.
\label{definition:transductive gnns}
\end{definition}

% Transductive learning (see Definition \ref{definition:transductive gnns}) refers to the application of current knowledge to given data. In simple terms, TGNNs is trained with training set data and test set data and then tested with test set data \cite{Mcpherson2001Birds},\cite{Wu2021Self}.\\
% % \cite{girvan2002community},\cite{Zhu2021Graph}
Transductive learning, as defined in Definition \ref{definition:transductive gnns}, entails leveraging existing knowledge to make predictions on a given dataset. In the context of TGNNs, this involves training the model using both training and test set data, followed by evaluating the model's performance on the test set. In essence, transductive learning is the utilization of known information to make predictions on new data \cite{Mcpherson2001Birds,Wu2021Self}. \\

\noindent \textit{\textbf{ii. Inductive Methods}}
\begin{definition}
\textbf{Inductive GNNs} Given a graph $G=(\mathcal{V},\mathcal{E})$, label information $\mathcal{Y}$ and Inductive Graph Neural Networks (\text{IGNNs}), IGNNs' training and  testing are performed  separately on training graph and testing graph with corresponding label information, where $G = G_{train} \cup G_{test}$ and $G_{train} \cap G_{test}=\emptyset$. $\mathcal{Y}_{train}, \mathcal{Y}_{test} \subseteq \mathcal{Y}$ and $\mathcal{Y}_{train} \cap \mathcal{Y}_{test}=\emptyset$. Noted that the label information can be node-level, edge-level or graph-level.
\label{definition:inductive gnns}
\end{definition}

Inductive learning, as defined in Definition \ref{definition:inductive gnns}, involves the learning process without utilizing test set or validation set samples during training, or in cases where these sets are not visible during training. The distinct advantage of IGNNs lies in their ability to leverage information from known nodes to generate embeddings for unknown nodes. A notable illustration of inductive learning on graphs is GraphSAGE \cite{Hamilton2017Inductive}, a versatile inductive framework that computes node representations by sampling neighborhoods of each node and employing a specific aggregator for effective information fusion. This approach efficiently generates node embeddings for previously unseen data, incorporating node feature information such as text attributes \cite{Chen2017Supervised}.
In contrast to GraphSAGE \cite{Hamilton2017Inductive}, FastGCN \cite{Chen2018Fastgcn} by Chen et al. interprets graph convolutions as integral transforms of embedding functions and independently samples nodes in each layer. Another inductive model, GraIL \cite{teru2020inductive} by Teru et al., reasons over local subgraph structures, enabling generalization to unseen entities and graphs after training.
Traditional inductive learning approaches, like node2vec \cite{Grover2016Node2vec}, can also be employed for this purpose.

\begin{table*}[htb]
\caption{Representative Works in Terms of Citations}
\label{tab:representation-works}
\newcommand{\tabincell}[2]{\begin{tabular}{@{}#1@{}}#2\end{tabular}}
\resizebox{\linewidth}{!}{
\begin{tabular}{l|l|c|c}
\toprule
Title & Venue & Year & Citation \\
\midrule
     Emergence of scaling in random networks \cite{barabasi1999emergence} & Science &  1999 & 46660       \\
     Semi-supervised classification with graph convolutional networks \cite{Kipf2016Semi-supervised} &	ICLR &	2016 &	37677  \\
     Birds of a feather: Homophily in social networks \cite{Mcpherson2001Birds}
   &  Annual Review of Sociology
 &  2001     &  24650        \\
 Community structure in social and biological networks \cite{girvan2002community}	& PNAS &	2002 &	19883   \\
 The pagerank citation ranking: Bringing order to the web \cite{Page1999Pagerank} &	WWW &	1999 &	19224       \\
Finding and evaluating community structure in networks \cite{newman2004finding} &	Physical Review E &	2004 &	17991       \\
Graph attention networks \cite{Velickovic2018Graph} &	ICLR &	2017 &	16283     \\
Community detection in graphs \cite{fortunato2010community} &	Physics Reports &	2010 &	12735      \\
Maximizing the spread of influence through a social network \cite{Kempe2003Maximizing} & KDD & 2003 & 10594      \\
Convolutional neural networks on graphs with fast localized spectral filtering \cite{defferrard2016convolutional} &	NeurIPS &	2016 &	9611     \\
Translating Embeddings for Modeling Multi-relational data \cite{Bordes2013Translating}  &	NeurIPS &	2013 &	9223     \\
The graph neural network model \cite{Scarselli2008The}	&	IEEE Transactions on Neural Networks	&	2008	&	9003	      \\
Neural message passing for quantum chemistry \cite{Gilmer2017Neural} &	ICML &	2017 &	8693      \\
How powerful are graph neural networks? \cite{Xu2019How} &	ICLR &	2018 &	8492      \\
 The link prediction problem for social networks \cite{Liben2003Link} &	\tabincell{l}{CIKM} &	2003 &	7315      \\
Line: Largescale information network embedding \cite{Tang2015Line} &	WWW &	2015 &	6417      \\
Spectral networks and locally connected networks on graphs \cite{Bruna2013Spectral} 	& 	ICLR	& 	2013	&	6301	     \\
Finding community structure in networks using the eigenvectors of matrices 
 \cite{newman2006finding} &	Physical review E &	2006 &	5901	       \\
\tabincell{l}{The emerging field of signal processing on graphs:\\ Extending high-dimensionaldata analysis to networks and other irregular domains} \cite{Shuman2013Emerging}&	IEEE Signal Processing Magazine &	2013 &	4645        \\
Knowledge graph embedding by translating on hyperplanes \cite{wang2014knowledge}  &	AAAI &	2014 &	4381      \\
Convolutional networks on graphs for learning molecular fingerprints \cite{Duvenaud2015Convolutional} &	NeurIPS &	2015 &	4308       \\
Friends and neighbors on the web \cite{Adamic2003Friends}&	Social Networks &	2003 &	3946 	       \\
Graph embedding and extensions: A general framework for dimensionality reduction \cite{Yan2006Graph}	&	TPAMI	&	2006	&	3638  \\
Graph evolution: Densification and shrinking diameters \cite{Leskovec2007Graph} &	TKDD &	2007 &	3352       \\
Graphs over time: densification laws, shrinking diameters and possible explanations\cite{Leskovec2005Graphs} &	KDD &	2005 &	3323    \\
The dynamics of viral marketing \cite{Leskovec2007Dynamics}&	ACM Transactions on the Web &	2007 &	3256     \\
Social structure from multiple networks. i. blockmodels of roles and positions \cite{White1976Social} &	American Journal of Sociology &	1976 &	3100       \\
A measure of betweenness centrality based on random walks \cite{Newman2005Measure} &	Social Networks	& 2005 &	3062	     \\
Detecting community structure in networks \cite{Newman2004Detecting} &	The European Physical Journal B &	2004 &	2829      \\
Simrank: a measure of structural-context similarity \cite{Jeh2002Simrankff} &	KDD &	2002 &	2802    \\

\bottomrule
\end{tabular}}
\footnotesize
\begin{tablenotes} 
\item[1] Source: Google Scholar, update time: 2023-06-07.
\end{tablenotes}
\end{table*}

\subsection{Research Trends}

We present notable methodologies, examining their citations in Table \ref{tab:representation-works}. 
A thorough analysis of the table reveals a historical evolution in the focus of GL research. Early endeavors delved into the intrinsic characteristics of graphs, highlighting elements like homophily \cite{Mcpherson2001Birds} and dynamics \cite{sankar2020dysat}. During this phase, various graph-centric tasks, including community detection \cite{girvan2002community}, link prediction \cite{Liben2003Link}, and influence maximization \cite{Kempe2003Maximizing}, were proposed. In contemporary research, there is a discernible shift towards addressing the effective modeling and solution of these fundamental problems \cite{Kipf2016Semi-supervised,Velickovic2018Graph}.
Moreover, an upsurge in works focusing on Large Language Models (LLMs) has transpired in recent months. This trend underscores a growing interest in augmenting LLMs with GL methodologies to bolster their reasoning capabilities.

\section{LLMs and GL}
\label{section-llm_gl}

Large Language Models (LLMs) have garnered substantial acclaim within both academic circles and industry due to their remarkable advancements. However, despite their prowess, LLMs exhibit imperfections, prompting the exploration of GL as a means to leverage structural data and address certain weaknesses.

\begin{figure*}[htb]
    \centering
    \includegraphics[width=0.98\textwidth]{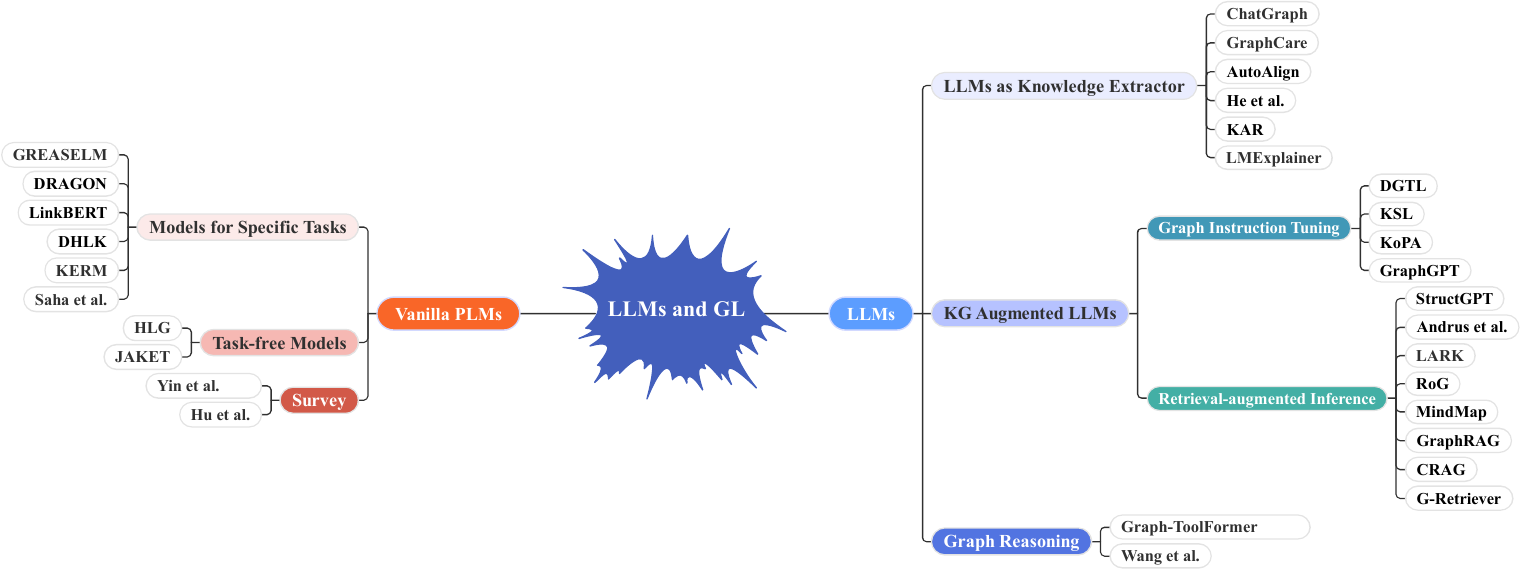}
     \caption{Current research on the combination of LLMs and GL.}
    \label{fig:LLM_GL}
\end{figure*}

Originating from Pre-trained Language Models (PLMs), LLMs inherit extensive knowledge from encyclopedic and commonsense corpora, empowering them to excel in various natural language processing (NLP) tasks such as question answering and sentiment analysis. Despite these capabilities, most PLMs encounter limitations in domain-specific tasks \cite{yin2022survey}. An effective strategy to mitigate this limitation involves augmenting PLMs with structured knowledge \cite{zhang2022greaselm}. Furthermore, incorporating structured knowledge has demonstrated notable efficacy in enhancing the interpretability of PLMs' outputs \cite{shi2023chatgraph}.
Despite the heightened resource requirements, including extensive training data and computing resources, LLMs have achieved unprecedented superiority across a spectrum of NLP tasks. In this section, we first illustrate how graph structural knowledge contributes to enhancing conventional PLMs. Subsequently, we delve into the integration of GL with Large Language Models (LLMs). The comprehensive structure of this section is depicted in Figure \ref{fig:LLM_GL}.

% Large Language Models have achieved great progress and appealed wide attention among both academic community and industry. However, LLMs are not perfect and GL is a way to help LLMs make use of structural data and overcome some weakness.
% As LLMs develop from Pre-trained Language Models (PLMs), we first show how to combine PLMs with GL and then show the current combination of LLMs and GL.

% Pre-trained Language Models are trained on encyclopedic and commonsense corpora, endowing them with great capabilities for various natural language processing (NLP) tasks, such as question answering and sentiment analysis. However, most PLMs lack the ability to perform specific domain tasks \cite{yin2022survey}. One effective way to solve this problem is to combine PLMs with structured knowledge \cite{zhang2022greaselm}.
% Moreover, structured knowledge has shown superiority in enhancing the explainability of PLMs' outputs \cite{shi2023chatgraph}. 
% LLMs requires larger amount of resources, including training data and computing resources, while also achieve unprecedented superiority over various NLP tasks. 
% In this section, we first demonstrate how graph structure knowledge helps improve vanilla PLMs. Then we show the combination of graph learning and Large Language Models (LLMs). The whole structure of this section is shown in Figure \ref{fig:LLM_GL}.

\subsection{Vanilla Pre-trained Language Models}
Vanilla Pre-trained Language Models are constructed upon the transformer architecture and trained on a diverse corpus, rendering them versatile for a range of downstream tasks through fine-tuning. In the subsequent sections, we delineate the most recent and representative methodologies across distinct tasks, wherein the incorporation of structural knowledge proves instrumental in enhancing the performance of these vanilla PLMs. This includes methodologies specifically tailored for particular tasks as well as task-agnostic approaches.

% \noindent \textit{\textbf{A. ModelS for Specific Tasks}}
\subsubsection{Models for Specific Tasks}

Vanilla PLMs are extensively employed in the Question Answering (QA) task, as evidenced by studies such as \cite{yasunaga2022linkbert, yasunaga2022deep, wang2023dynamic}. In this context, leveraging structural knowledge becomes pivotal for enhancing the efficacy of PLMs. For instance, many existing methodologies exhibit limitations in capturing latent relationships between concepts. Addressing this gap, Zhang et al. \cite{zhang2022greaselm} introduce GREASELM, a methodology that integrates representations from both vanilla PLMs and GNNs. LinkBERT \cite{yasunaga2022linkbert} adopts a similar approach, enhancing the capacity of PLMs by learning links between documents. DRAGON \cite{yasunaga2022deep} is also designed to learn a comprehensive fusion of text and knowledge graphs (KGs) at scale.

Moreover, certain models are specifically tailored for distinct scenarios. In the realm of passage re-ranking, where the goal is to improve the retrieval of relevant information, methods relying solely on vanilla PLMs struggle with vocabulary mismatches and lack domain-specific knowledge. To address this, Dong et al. propose KERM \cite{dong2022incorporating}, a methodology that introduces explicit knowledge to vanilla PLMs. KERM employs a reliable knowledge meta-graph derived from noisy and incomplete knowledge graphs. This meta-graph, along with vanilla Graph-based Language Models (GLMs), serves as two encoders contributing to the final outputs.
Additionally, Saha et al. \cite{saha2022explanation} identify challenges with vanilla PLMs in generating structured outputs that align with graph constraints and maintain semantic correctness, especially when confronted with limited supervision. To overcome this, they introduce a graph perturbation method and combine it with Max-Margin and InfoNCE losses, leading to substantial performance improvements.

% Vanilla PLMs are widely used for the Question Answer (QA) task \cite{yasunaga2022linkbert,yasunaga2022deep,wang2023dynamic}, where structural knowledge can be used to improve PLMs.
% For example,
% Most of previous methods lack the ability to represent latent relationships between concepts. To solve this problem, Zhang et al. \cite{zhang2022greaselm} propose GREASELM, which fuses the representations from vanilla PLMs and GNNs.
% LinkBERT \cite{yasunaga2022linkbert} adopts similar idea to improve PLMs' capacity by learning links between documents.
% DRAGON \cite{yasunaga2022deep} is also trained to learn a deep fusion of text and KG at scale.

% There are also models concentrate on specific scenarios.
% Passage re-ranking helps vanilla PLMs retrieve better messages. Previous vanilla PLMs-based methods cannot deal with the vocabulary mismatch problem and lack domain knowledge. Thus, Dong et al. introduce explicit knowledge to vanilla PLMs with KERM \cite{dong2022incorporating}. They first obtain a reliable knowledge meta-graph from noisy and incomplete knowledge graphs and then utilize the meta-graph and vanilla GLMs as two encoders to get final outputs.
% Saha et al. \cite{saha2022explanation} find that with limited supervision, vanilla PLMs cannot generate structured outputs that meet graph constraints and are semantically correct. They propose a simple yet effective graph perturbation method and utilize it with Max-Margin and InfoNCE losses, which lead to significant improvements.

% \noindent \textit{\textbf{B. Task-free Models}}
\subsubsection{Task-free Models}

Task-free methodologies, independent of task-specific fine-tuning, constitute another avenue in advancing PLMs. A noteworthy example is the work by Yu et al. \cite{yu2022jaket}, who introduce JAKET, a joint pre-training framework that seamlessly integrates both vanilla PLMs and GNNs. This design exhibits commendable performance across various knowledge-aware NLP tasks.
Furthermore, HLG \cite{li2022enhancing} represents a task-free enhancement module dedicated to fortifying Chinese PLMs by incorporating linguistic knowledge. HLG adopts a unique approach by constructing a heterogeneous graph and utilizing a GNN to capture the intricate structure of Chinese linguistics.

For a more comprehensive exploration of the synergy between Knowledge Graphs (KGs) and PLMs, interested readers are directed to additional surveys in the literature \cite{yin2022survey, hu2023survey}.

% There are also task-free methods that do not rely on task-specific fine-tuning. For instance, Yu et al. \cite{yu2022jaket} propose JAKET, a joint pre-training framework, to leverage both vanilla PLMs and GNNs. The design achieves good performance on several knowledge-aware NLP tasks.
% HLG \cite{li2022enhancing} is a task-free enhancement module, which aims to enhance Chinese PLMs by integrating linguistics knowledge. Specifically, HLG constructs a heterogeneous graph and uses a GNN to model the structure of Chinese linguistic.

% We refer readers to more survey on the combination of KGs and PLMs \cite{yin2022survey,hu2023survey}.

\subsection{Large Language Models}

\begin{figure*}[htb]
    \centering
    \includegraphics[width=0.85\textwidth]{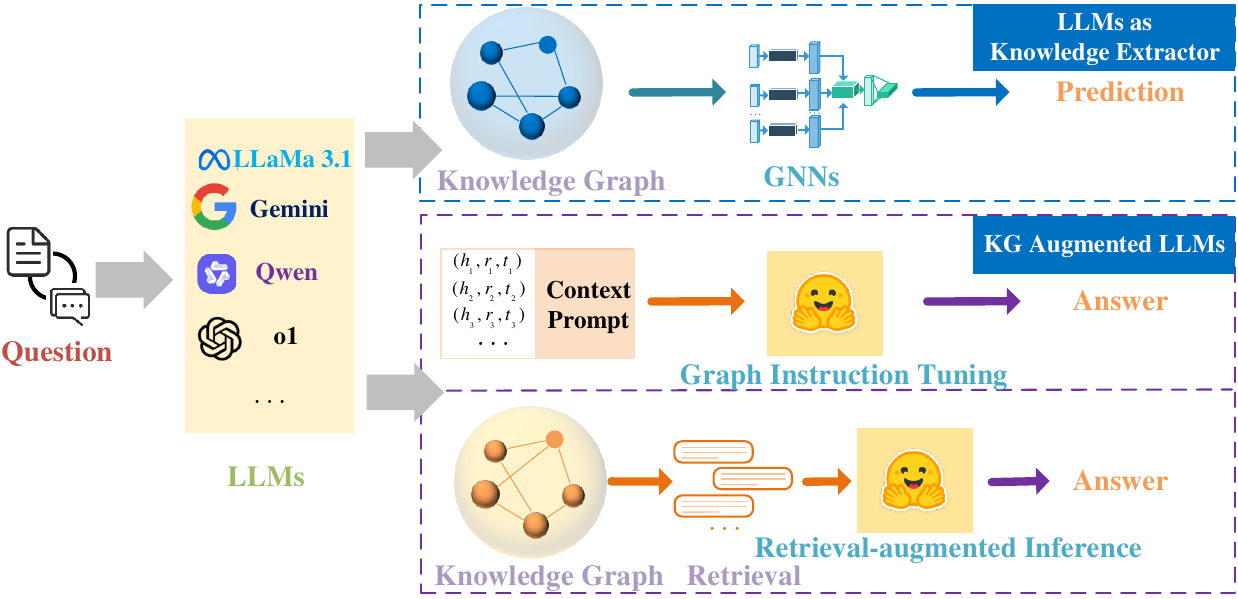}
     \caption{Current ways to improve LLMs with graphs.}
    \label{fig:LLM_Graph}
\end{figure*}

In contrast to vanilla Pre-trained Language Models, Large Language Models (LLMs) are trained on vast datasets, boasting billions of parameters, exemplified by models like OPT (175 billion parameters) \cite{zhang2022opt}, PaLM 2 \cite{anil2023palm}, and GPT-4 \cite{openai2023gpt4}. The substantial parameter count equips LLMs with the capacity to address intricate problems, yielding remarkable performance across a spectrum of tasks, including multi-modal domains.
Numerous methodologies have recently emerged to fully exploit the capabilities of LLMs. We categorize these endeavors under the theme of "LLMs as Knowledge Extractors," where LLMs serve as knowledge extractors, contributing to or enhancing downstream tasks. Despite their superior performance, LLMs encounter challenges akin to PLMs, such as the lack of factual knowledge and an understanding of latent relationships between entities. Consequently, structured knowledge remains pivotal in mitigating these shortcomings.
It is noteworthy that integrating structured knowledge with LLMs differs from analogous efforts with PLMs, primarily due to the huge cost of fine-tuning LLMs. 
As illustrated in Figure \ref{fig:LLM_Graph}, there are two principal methods for combining structured knowledge with LLMs: LLMs as knowledge extractor and KG augmented LLMs. 
The former focuses on enhancing the performance of graph GNNs or other downstream tasks, while the latter aims to augment the capabilities of LLMs directly.
Furthermore, researchers are exploring various dimensions regarding how knowledge graphs can augment LLMs beyond these mainstream approaches.

% Unlike vanilla PLMs, Large Language Models (LLMs) are trained on massive datasets and have billions of parameters, such as OPT (175 billion parameters) \cite{zhang2022opt}, PaLM 2 \cite{anil2023palm}, and GPT-4 \cite{openai2023gpt4}. The huge number of parameters makes LLMs capable of dealing with complex problems and achieving amazing performance on various tasks, including multi-modal tasks.
% Recently lots of methods are proposed to make full use of such ability of LLMs. We conclude those research with "LLMs as Knowledge Extractor", where LLMs are used as knowledge extractors to assist or help improving further downstream tasks.
% Furthermore, LLMs are also facing similar problems as PLMs, despite of better performance. As a result, structured knowledge still play an important role on helping LLMs overcome shortcoming like lacking factual knowledge and understanding latent relationships between entities.
% The main difference of combining PLMs or LLMs with other models is that it is not easy to fine-tune LLMs to train a new end-to-end model.
% We show the two main ways to combine structured knowledge with LLMs in Figure \ref{fig:LLM_Graph}.
% Besides, some researchers also explore other interesting aspects that how knowledge graphs enhance LLMs. 

% However, using flattened texts as inputs limits LLMs' performance in some scenarios, especially for knowledge-intensive tasks. To overcome this drawback, there are two main ways to improve LLMs, as shown in Figure \ref{fig:LLM_Graph}. Additionally, we introduce works on LLMs' abilities in graph reasoning.
% However, 

% \textbf{A. Enhancing LLMs With Graphs}

% \noindent \textit{\textbf{A. LLMs as Knowledge Extractor}}
\subsubsection{LLMs as Knowledge Extractor}

One prominent approach to leveraging LLMs is to use them as knowledge extractors for GNNs. This method reduces the need for manual intervention while maintaining high accuracy and facilitates the generation of human-understandable explanations for various predictions. Illustratively, Shi et al. \cite{shi2023chatgraph} employ ChatGPT as a knowledge extractor to distill refined and structured information from raw data. Subsequently, this knowledge is fed into a GCN for text classification, concurrently enhancing interpretability.
In entity alignment tasks between KGs, Zhang et al. \cite{zhang2023autoalign} introduce AutoAlign, utilizing an LLM to extract concepts with similar meanings but different descriptions from two KGs. These extracted concepts form the basis for constructing a predicate-proximity graph. 
% Similarly, Bi et al. \cite{bi2023codekgc} propose CodeKGC, which transforms natural language inputs into code-format inputs and uses them as schema-aware prompts to improve LLMs' performance on knowledge graph construction.
He et al. \cite{he2023explanations} leverage an LLM for obtaining textual explanations of target nodes in text-attributed graphs, employing zero-shot classification. Subsequently, an LLM-to-LM interpreter is designed to convert these explanations into features for input into GNNs.
Chen et al. \cite{chen2023lmexplainer} also explore the explanation of LMs with the proposed model LMExplainer.
LMExplainer uses LMs to generate language embeddings and to construct a question-related subgraph. These are then put into GNN and GAT for final prediction. Attention scores from the GAT can be used for weighing the importance of GNN embeddings and for the answer with language embeddings.
% uses KG and GAT to extract key decision signals of language models. It then uses the input context, model-predicted output, trigger sentence, and extracted key components as prompts for LLMs to generate explanations.
LLMs serving as knowledge extractors have a wide range of applications. For example,
KAR \cite{xi2023towards} employs LLMs to extract external knowledge for both users and items in recommendation systems. 
In specialized domains requiring specific expertise, LLMs also function as effective knowledge extractors. For instance, GraphCare \cite{jiang2023graphcare} addresses healthcare prediction, a domain reliant on medical knowledge. In this context, GraphCare extracts knowledge from both LLMs and external biomedical KGs to generate patient-specific KGs. These personalized KGs are then employed for healthcare prediction using GNNs.

% The first way to use LLMs is to treat them as knowledge extractors for GNNs, which avoids large manual effort and keep high accuracy meanwhile.
%  For instance, 
% Shi et al. \cite{shi2023chatgraph} utilize ChatGPT to extract refined and structured knowledge from raw data, and then feed this knowledge to GCN for text classification. 
% % The proposed method also improves the interpretability of the results.
% Zhang et al. \cite{zhang2023autoalign} propose AutoAlign for entity alignment between KGs. AutoAlign utilizes an LLM to extract concepts that belongs to same meaning but with different description between two KGs, which are used to construct predicate-proximity-graph.
% He et al. \cite{he2023explanations} use an LLM to get textual explanation of target node in a text-attributed graph by implementing zero-shot classification. Afterwards, they design an LLM-to-LM interpreter to transform the former explanations into features as inputs of GNNs. 
% KAR \cite{xi2023towards} also uses LLMs to extract external knowledge for both users and items in recommendation system.
% In scenarios that requires specific skills, LLM can also perform as a good knowledge extractor. For example, GraphCare \cite{jiang2023graphcare} focuses on healthcare prediction, a specific scenario that requires medical knowledge. To address this challenge, GraphCare extracts knowledge from both LLMs and external biomedical KGs to generate patient-specific KGs, which are then used for healthcare prediction with GNNs.

% \noindent \textit{\textbf{B. KG Augmented LLMs}}
\subsubsection{KG Augmented LLMs}
\label{subsubsec-kg-augmented-llms}
Although the LLMs have achieved great success, they still face the challenge of hallucination, outdated knowledge, private domain knowledge and complex inference in specific fields. Fortunately, knowledge graph can help to allievate such problems by providing external knowledge for LLMs and also helps to improve the inference ability of LLMs on long and complex context. There are two main ways to leverage the KG to augment the ability of LLMs: 
% A conventional strategy involves enhancing LLMs with KGs to address issues such as outdated knowledge, the absence of external knowledge, and the generation of hallucinations. KGs play a pivotal role in mitigating these challenges. 

\noindent \textit{\textbf{i. Graph Instruction Tuning}}

The first approach involves fine-tuning LLMs with structured information. It is widely recognized that incorporating structural knowledge can enhance the performance of LLMs. Numerous studies have focused on this method, varying in how they represent structural information and integrate it with LLMs to improve their capacity. 
For instance,
DGTL \cite{qin2023disentangled} combines the capabilities of LLMs and GL by generating text embeddings of given context and utilizing them along with learned embeddings from GNNs to obtain final answers. 
Unlike DGTL, more studies opt for fine-tuning LLMs.
For example,
KSL \cite{feng2023knowledge} retrieve relevant knowledge subgraph and encode it into text prompt for further instruct tuning of LLMs.
There are also studies that train both the graph learning component and fine-tune the LLMs.
For instance,
KoPA \cite{zhang2023making} first pre-trains structural embeddings for the KGs (i.e., entities and relations) and then employs instruction tuning to fine-tune the LLM.
GraphGPT \cite{tang2024graphgpt} adopts a similar approach. It first aligns graph tokens (i.e., nodes and their neighbors) and language tokens (i.e., instructions and node text). Then it constructs instruction consisting of graph tokens and human question for self-supervised instruction tuning and task-specific instruction tuning.

\noindent \textit{\textbf{ii. Retrieval-augmented Inference}}

Given the high cost of fine-tuning LLMs and the rapid advancement in LLM technology, many studies are now concentrating on fully utilizing KGs to enhance LLMs without requiring fine-tuning. The core of this approach involves accurately retrieving relevant content from KGs and using appropriate prompts to guide the generation of responses.
We categorize current LLMs that retrieve knowledge from KGs into two primary types: flattened retrieval and graph-based retrieval. Most previous research has focused on flattened retrieval, which directly extracts relevant content based on relationships defined in KGs.
For instance,
StructGPT \cite{jiang2023structgpt} integrates external structured data, such as KGs, tables, and databases, through an iterative reading-then-reasoning procedure.
Andrus et al. \cite{andrus2022enhanced} propose extracting an external dynamic KG from the original story text, employing it as a prompt to augment LLMs.
To address the limited context window of LLMs, Choudhary and Reddy \cite{choudhary2023complex} introduce LARK, decomposing input queries using KGs and converting them into LLM prompts, resulting in a refined set of answers. 
Diverging from methods treating KGs solely as factual knowledge bases, RoG \cite{luo2023reasoning} synergizes LLMs with KGs for faithful and interpretable reasoning on graphs. RoG generates faithful relation paths grounded by KGs and retrieves valid reasoning paths from KGs as inputs for LLMs. Similarly, MindMap \cite{wen2023mindmap} underscores the use of reasoning pathways from KGs to enhance the interpretability of LLM outputs.

\begin{figure}[t]
    \centering
    \includegraphics[width=0.49\textwidth]{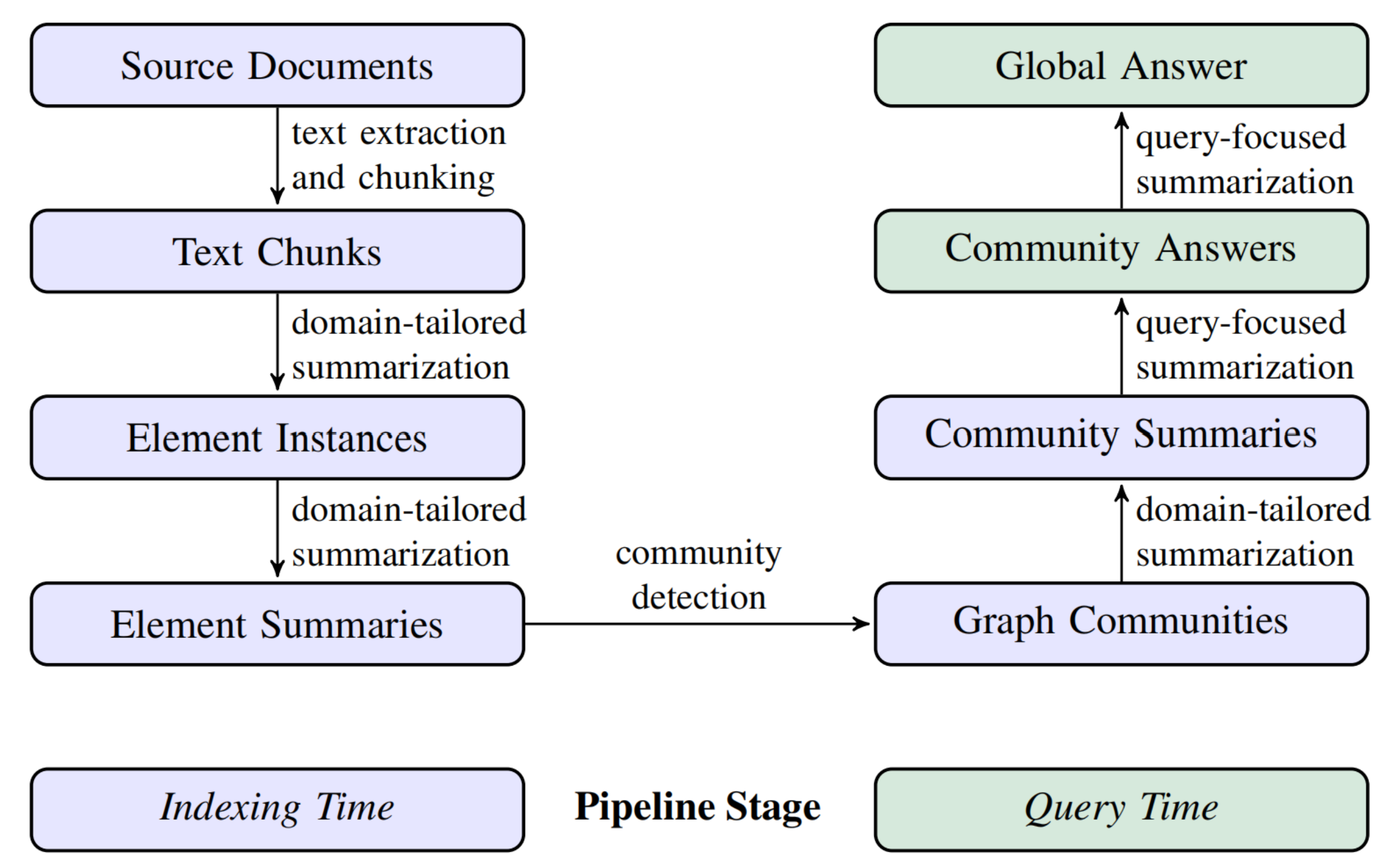}
	\caption{GraphRAG pipline \cite{edge2024local}
 }
    \label{fig:graphrag}
\end{figure}

Flattened retrieval methods often struggle with the query-focused summarization (QFS) task, especially when presented with broad, global questions about an entire text corpus \cite{edge2024local}. For example, when asked "What's the theme of the story?" based on \textit{The Lord of the Rings}, it is challenging to find an answer by searching any single paragraph. However, by segmenting the document into text chunks and constructing graphs with summaries, LLMs can more effectively handle QFS tasks. Pioneering works such as GraphRAG \cite{edge2024local}, CRAG \cite{hu2024grag}, Xu et al. \cite{xu2024retrieval} and G-Retriever \cite{he2024g} have been introduced to advance this area.
As illustrated in Figure \ref{fig:graphrag}, GraphRAG segments documents into text chunks and constructs graphs composed of entities, relationships, and claims. The entity types are predefined, and each entity is associated with a summary. After constructing the graphs, GraphRAG applies community detection algorithms, such as Leiden \cite{traag2019louvain}, to partition the graphs into hierarchical communities. Finally, GraphRAG utilizes LLMs to generate community-level summaries for answering global queries. 
Unlike GraphRAG, which does not involve any training, G-Retriever trains a GNN model to align the outputs of the GNN with text tokens. Specifically, G-Retriever first transforms nodes and edges into representations using a pretrained LM. It then uses the same LM to generate a query representation, which is employed to retrieve relevant nodes and edges, constructing a subgraph. G-Retriever applies a GAT to obtain the graph token and projects this token into the same vector space as the first layer of a frozen LLM. This alignment allows G-Retriever to effectively train the GNN to synchronize graph and text representations, leveraging the LLM’s capabilities to perform a variety of graph-related downstream tasks.

\subsubsection{Graph Reasoning}

LLMs have demonstrated proficiency in tasks involving structured inputs, such as code generation. However, their performance falls short in GL tasks. To tackle this challenge, Jiawei Zhang introduces Graph-ToolFormer, a framework that instructs LLMs to leverage external graph reasoning API tools \cite{zhang2023graph}.
In a related vein, Wang et al. present NLGraph, a comprehensive benchmark designed to evaluate the efficacy of LLMs in solving graph-based problems \cite{wang2023can}. Their findings indicate that LLMs exhibit preliminary graph reasoning capabilities but are susceptible to graphs with spurious correlations. Zhu et al. \cite{zhu2023llms} scrutinize the abilities of LLMs in KG construction and reasoning, determining that while LLMs can address complex problems, they fall behind State-of-the-Art (SOTA) models.
Guo et al. \cite{guo2023gpt4graph} contribute to the evaluation of LLMs' proficiency in comprehending graph data, further contributing to the understanding of LLMs' capabilities in the context of graph-based tasks.

% We direct readers to additional surveys exploring the integration of KGs and LLMs \cite{liu2023towards,chen2023exploring,pan2023unifying,yang2023chatgpt}.

% LLMs have performed well on some tasks with structured inputs, such as code generation. However, they exhibit flaws when it comes to graph learning tasks. To address this problem, Jiawei Zhang proposes Graph-ToolFormer, a framework that teaches LLMs to utilize external graph reasoning API tools \cite{zhang2023graph}.
% Wang et al. propose NLGraph, a comprehensive benchmark for LLMs' abilities in solving graph-based problems \cite{wang2023can}. They find that LLMs can demonstrate preliminary graph reasoning, but they are vulnerable to graphs with spurious correlations. Zhu et al. \cite{zhu2023llms} test LLMs' abilities on KG construction and reasoning and conclude that LLMs can address complex problems but lag behind SOTA models.
% Guo et al. \cite{guo2023gpt4graph} also evaluate the proficiency of LLMs in comprehending graph data.

% We refer readers to more surveys on the combination of knowledge garphs and LLMs \cite{liu2023towards,chen2023exploring,pan2023unifying,yang2023chatgpt}.

\section{Applications}
\label{section-applications}
This section provides an overview of the prevalent applications of GL within contemporary contexts. These applications span traditional machine learning domains, encompassing recommendation systems, natural language processing, computer vision, and financial technology (FinTech). 
% Furthermore, emergent applications are observed in scientific domains, specifically within the realms of chemistry, biology, physics, and mathematics. 
To conclude, an encapsulation of widely employed datasets within the field of GL is presented.

\begin{figure*}[t]
    \centering
    \includegraphics[width=1\textwidth]{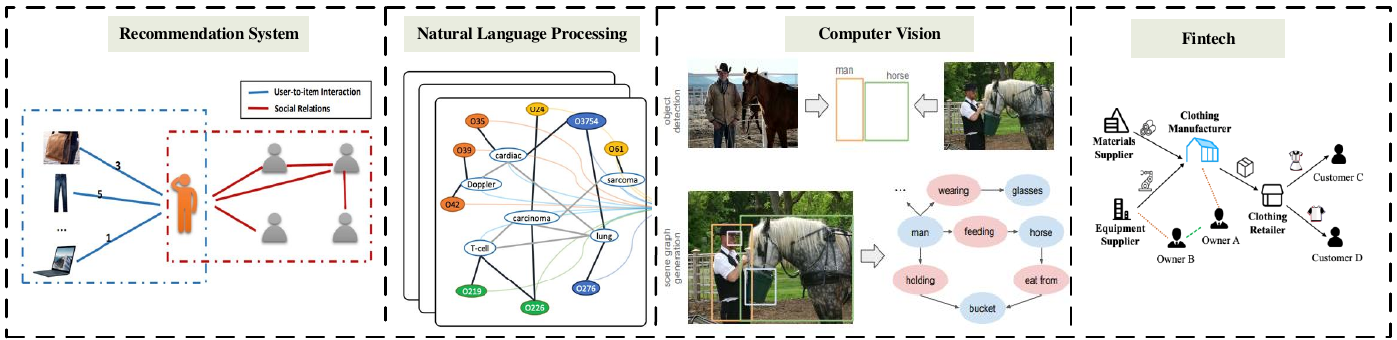}
	\caption{Examples of applications. These applications include Recommendation System \cite{Fan2019Graph}, Natural Language Processing \cite{Yao2018Graph}, Computer Vision \cite{Xu2017Scene}, Fintech \cite{yang2021financial}.
 % Chemistry and Biology \cite{Rong2020Self}, Physics and Mathematics \cite{Kipf2018Neural}.
 }
    \label{fig:application}
\end{figure*}

% This section introduces the current major applications of graph learning in the real world. These include traditional machine learning scenarios such as recommendation systems, natural language processing, computer vision, and FinTech. There are also burgeoning applications in scientific scenarios such as chemistry, biology, physics, and mathematics. Finally, we summarize popular datasets used in graph learning.

% \subsection{Scenarios}
\subsection{Recommendation System}
Recommendation systems function as information filters, suggesting items aligned with users' preferences or needs. Predominant research in this domain has traditionally relied on collaborative filtering and subsequently incorporated deep neural networks. However, in recent years, the pervasive adoption of graph neural network models, leveraging their capacity to effectively represent linked data structures, has become prominent within recommendation systems. Notably, these models excel in capturing intricate relationships between users and products \cite{Fan2019Graph, Wu2021Self}.

% Recommendation systems are information filters that recommend items that users need or are interested in. Previous related research has mainly been based on collaborative filtering and then deep neural networks. In recent years, due to the advantage of graph neural network modeling linked data, they have been widely used in recommendation systems to model the relation between users and products \cite{Fan2019Graph}, \cite{Wu2021Self}. 
% % In addition, different variants of GNNs have been used in recommendations.

\subsubsection{Heterogeneous Graph}
Heterogeneous graphs are frequently employed in recommendation systems to model users and items characterized by distinct properties. For instance, Dong et al. \cite{Dong2012Link} undertake link prediction and recommendation tasks across heterogeneous social networks. Shi et al. \cite{Shi2015Semantic} introduce the notion of semantic paths, wherein weights are assigned to heterogeneous information graphs and meta-paths to enhance the model's ability to seamlessly integrate diverse information. Subsequently, Shi et al. \cite{Shi2019Heterogeneous} employ an innovative network embedding approach on heterogeneous graphs to explore high-level structural information pertaining to users and items.

% The heterogeneous graph is often used to model users and items with two different properties in recommendation systems. 
% For instance, Dong et al. \cite{Dong2012Link} perform link prediction and recommendation across heterogeneous social networks. Shi et al. \cite{Shi2015Semantic} propose the concept of semantic paths by setting weights on heterogeneous information graphs and meta-paths to assist the model in better integrating heterogeneous information. Then, Shi et al. \cite{Shi2019Heterogeneous} utilize a novel network embedding method on heterogeneous graphs to explore high-level structural information of users and items.

\subsubsection{Hypergraph}
The hypergraph serves as a model to elucidate high-order relationships among users and items within the context of recommendation systems. Yu et al. \cite{yu2021self} introduce a multi-channel hypergraph convolutional network designed to leverage high-order user relations, such as social connections among users and shared purchasing behaviors. similarly, Yang et al. \cite{yang2022multi} employ hypergraph neural networks to capture global multi-behavior dependencies, relying on the dynamic and heterogeneous relationships existing among users and items. Additionally, Xia et al. \cite{xia2022self} propose a hypergraph transformer framework utilizing a self-supervised approach to augment user representation through the incorporation of user-item interactions.

% The hypergraph describes high-order relationships between multiple users and items for recommendation. Yu et al. \cite{yu2021self} propose a multi-channel hypergraph convolutional network to exploit high-order user relations such as social relations between users and their same purchase behavior. Yang et al. \cite{yang2022multi} use hypergraph neural networks to model global multi-behavior dependencies based on the dynamic heterogeneous relations of users and items. Xia et al. \cite{xia2022self} propose a hypergraph transformer framework based on a self-supervised approach to enhance user representation with user-item interactions. 

\subsubsection{Dynamic Graph}
The dynamic graph paradigm introduces a temporal dimension into recommendation systems, accounting for the evolving nature of user interactions over time. Addressing the dynamic nature of user behavior in online communities, Song et al. \cite{Song2019session} present a recommendation model predicated on a dynamic graph attention network. This network adeptly captures the dynamic shifts in user interests and considers the impact of user interactions in discerning their evolving preferences.
Zhang et al. \cite{zhang2022dynamic} adopt a dynamic graph framework for user sequence recommendation, wherein distinct users' historical sequences are collectively modeled, as opposed to focusing solely on individual user sequences.
\subsection{Natural Language Processing}
Natural Language Processing (NLP) constitutes a pivotal domain in the field of artificial intelligence, centering its focus on language as the subject of investigation. Its primary objective is to achieve the comprehension and application of human language by machines. Following are some typical research directions.

% Natural Language Processing (NLP) is a crucial area of artificial intelligence. It takes language as the research object and aims to realize the understanding and application of human language by machine. Following are some typical research directions.

\subsubsection{Text Classification}
Text classification stands as a foundational challenge in the landscape of NLP research. Addressing the realm of weakly supervised text classification, Peng et al. \cite{Peng2018Large} adeptly devise a model that delineates the graph structure of keywords. Employing GNNs, they further quantify the correlation between keywords, enhancing the nuanced understanding of textual content.
In contrast to prevalent applications relying on CNN, Yao et al. \cite{Yao2018Graph} introduce the TextGCN model. This model is designed to assimilate representations of both words and documents, offering a distinctive approach to text classification. Correspondingly, Lin et al. \cite{lin2021bertgcn} embrace a graph-based CNN methodology, strategically employed to capture intricate semantic information over extended distances, thereby contributing to the refinement of text classification outcomes.

% Text classification is a classic problem in NLP research. For the weakly supervised text classification, Peng et al. \cite{Peng2018Large} model keywords graph structure and calculate the correlation between keywords with the help of GNN. 
% While unlike many applications of CNN, Yao et al. \cite{Yao2018Graph} propose the TextGCN model to learn representations of words and documents. Similarly, Lin et al. \cite{lin2021bertgcn} adopt a graph-based CNN to capture long-distance semantic information for text classification. 
% Zhang et al. \cite{zhang2021weakly} combine BERT and GCN modules for transductive learning and obtain promising results on text classification datasets.

\subsubsection{Information Extraction}
Relation extraction constitutes a fundamental pursuit within the domain of information extraction, directed at discerning target relations embedded within textual entities \cite{zhao2022multi}. Endeavors to enhance the efficacy of relation extraction have led to diverse methodological innovations. Rocktaschel et al. \cite{Rocktaschel2015Injecting} introduce a matrix factorization approach to refine the embedding representation of entity pairs and associated relations. This method aims to augment the generalization capacity of relation extraction models.
Song et al. \cite{Song2018N-ary} present a novel graph-state LSTM model. This model not only captures document graphs but also preserves their inherent structure, mitigating information loss arising from graph fragmentation. Simultaneously, Zhang et al. \cite{Zhang2018Graph2} leverage GCN to extract entity dependencies in text based on the underlying dependency tree. To uphold the precision of information extraction, a novel pruning strategy is introduced. These approaches collectively contribute to the advancement of relation extraction methodologies.

\subsubsection{Question Answering}
Question answering is a crucial task in natural language generation. Zhang et al. \cite{zhang2022greaselm} integrate PLMs and GNN for encoding to capture language contextual relations. Furthermore, the exploration of visual question answering has emerged as a prominent research focus in recent years. Teney et al. \cite{Teney2017Graph} develop a graph structure based deep neural network to model the scene objects and the question words, instead of using CNN or LSTM. In the domain of 'fact-based' question answering, Narasimhan et al. \cite{Narasimhan2018Out} undertake the modeling of entity graphs utilizing Graph Convolutional Networks (GCN). This methodology facilitates sophisticated reasoning about answers within fact-centric contexts.
% {Document Summarization}

% {Keyphrase Extraction}

% {Question Answering}

% KGQA

% {Text Classification}

% \begin{itemize}
%     \item Text classification
%     \item Sequence labeling
%     \item Neural machine translation
%     \item Relation extraction
%     \item Event extraction
%     \item Other applications
% \end{itemize}

\subsection{Computer Vision}
%\textbf{3D point cloud recognization}
% \textcolor{red}{logic, relation with graph learning}
Graph-structured data finds widespread application in diverse computer vision tasks, spanning images, videos, and icons. Researchers have extensively investigated the utility of graph structures across various facets, notably in scene graph generation and object recognition.

The Scene Graph, characterized as a directed graph, encapsulates the semantic essence of a scene, with nodes denoting objects and edges signifying the relationships between these objects. Xu et al. \cite{Xu2017Scene} contribute an end-to-end model adept at generating a structured scene representation from input images utilizing scene graphs.
% Yang et al. \cite{Yang2018Graph} present Graph R-CNN, an adept and efficient model for detecting objects and their interrelations in images. Their innovative approach introduces an attentional Graph Convolutional Network (aGCN) that effectively captures contextual information, enhancing the discernment of relationships between objects.
Moreover, 
% Jain et al. \cite{Jain2016Structural} advocate for harnessing the synergy between high-level spatio-temporal graphs and the success of RNNs in sequence learning. Wang et al. \cite{Wang2019Dynamic} introduce EdgeConv, which operates on dynamically computed graphs at each layer, proving advantageous for CNN-based high-level tasks on point clouds, encompassing tasks such as classification and segmentation.
Sarlin et al. \cite{Sarlin2020Superglue} propose SuperGlue, a neural network designed for matching two sets of local features by concurrently identifying correspondences and rejecting non-matchable points. This approach facilitates joint reasoning about the underlying 3D scene and the assignment of features.

\subsection{FinTech}
Fintech, an amalgamation of artificial intelligence and other computational technologies, delineates the strategic utilization of these tools within the conventional financial industry to enhance operational efficacy and effectiveness. The financial market, comprising enterprises, individuals, and regulatory entities, inherently manifests as an extensive and intricate network, meriting comprehensive exploration. Consequently, GL has emerged as a prevalent methodology within the Fintech domain, with particular emphasis on two principal facets: corporate entities and individual stakeholders.

% Fintech refers to the application of AI and other computer technologies to the traditional financial industry to improve the effect and efficiency of its operation. Financial market, including enterprises, persons and regulators, naturally forms a giant complex network which is worth to mine. Thus, graph learning is widely applied in Fintech, mainly divided into two aspects: company and individual.

\subsubsection{Company Aspect}

\noindent \textit{\textbf{i. Supply Chain Network}} 
The supply chain represents a pivotal component of business operations. Addressing the imperative of mitigating visibility challenges within the supply chain, Kosasih and Brintrup \cite{kosasih2021machine} employ a GNN methodology to identify potential buyers within the network of automotive transactions. Yang et al. \cite{yang2021financial} propose a GNN model designed to unearth the intricate supply chain relationships among small and medium-sized enterprises(SMEs), with a specific focus on predicting financial risks. Cheng and Li \cite{cheng2021modeling} leverage supply chain dynamics and additional relational information to construct inter-company relationships. Subsequently, they employ GAT to model the spillover effects on a company's momentum, enhancing the precision of stock movement predictions. \\

% Supply chain is an important part of business operations. To solve the visibility problem on the supply chain, Kosasih and Brintrup \cite{kosasih2021machine} detect potential buyers on the car network through a GNN method. Yang et al. \cite{yang2021financial} propose a novel GNN model to mine the supply chain relationships of small and medium-sized enterprises for financial risk prediction. Cheng and Li \cite{cheng2021modeling} utilize supply chains and other relations to build company relations and then use GAT to model the company's momentum spillover effect for stock movement prediction.\\

\noindent \textit{\textbf{ii. Guarantee Network}}
The guarantee network predominantly arises from mutual guarantees among SMEs, contributing to the enhancement of credit ratings. In the context of monitoring potential default risks within the guarantee network, Cheng et al. \cite{cheng2020contagious} propose to calculate the risk score of contagion chains based on the structured loan data, structured in the form of a graph. Additionally, Cheng et al. \cite{cheng2020delinquent} employ GAT to acquire node representations within the network. This enables the prediction of default probabilities for a company, taking into consideration both its temporal behavior and its structural position within the guarantee network. \\

% Guarantee network is mainly the product of small- and medium-sized enterprises (SMEs) guaranteeing each other to improve credit rating. To monitor the potential default risk on the guarantee network, Cheng et al. \cite{cheng2020contagious} calculate the risk score of the contagion chain based on the graph-structured loan data. Cheng et al. \cite{cheng2020delinquent} also leverage GAT to learn network node representations and predict the default probability of a company according to its temporal behavior and structural position in the guarantee network. \\

\noindent \textit{\textbf{iii. Investment-shareholding Relation Network}}

The treatment of the investment-shareholding relation network as heterogeneous graphs has become a prevailing approach for modeling spillover effects relevant to diverse financial tasks, including stock predictions \cite{li2021modeling}, asset pricing \cite{huang2022asset}, and risk propagation \cite{Cheng2022Subsequence}. 
% Grounded in the investment relations within the capital market, Cheng et al. \cite{cheng2019risk} employ GCN to model the intricate relationships between the target company and its affiliated entities within a graph framework, thereby contributing to the prediction of stock movements.
Li et al. \cite{li2021modeling} propose LSTM-RGCN, a model specifically designed to simulate linkages between stocks and enhance predictive efficacy. Z
heng et al. \cite{zheng2021heterogeneous} propose a heterogeneous-attention-network-based model, leveraging relational information within available financial networks to address the bankruptcy prediction problem concerning SMEs.
Moreover, hypergraph structures have demonstrated utility in the prediction of enterprise risks \cite{zhao2022fisrebp}. Zhang et al. \cite{zhang2022heterogeneous} contribute to this paradigm by modeling bank data as a heterogeneous information network. They systematically mine relations within the financial activities of small and micro-enterprise users, particularly in commercial banking service scenarios, to facilitate comprehensive default analysis \cite{Cheng2022Subsequence}.

% It is natural to treat investment-shareholding relation network as heterogeneous graphs to model spillover effects for stock predictions \cite{tan2022finhgnn, cheng2021modeling}, asset pricing \cite{huang2022asset} and risk propagation \cite{Cheng2022Subsequence}.
% Based on the investment relation in the capital market, Cheng et al. 
% \cite{cheng2019risk} utilize GCN to model the relationship between the target company and its related companies on graph for stock movement prediction.
% % Tan et  al. \cite{tan2022finhgnn} propose FinHGNN to model spillover effects for stock predictions using heterogeneous graph neural networks.
% % Huang et al. \cite{huang2022asset} also utilize heterogeneous graph neural networks to solve the problem of asset pricing.
% Li et al. \cite{li2021modeling} also propose LSTM-RGCN, a model to simulate the linkages between stocks to enhance the effect of prediction. 
% Zheng et al. \cite{zheng2021heterogeneous} put forward a heterogeneous-attention-network-based model to explore the bankruptcy prediction problem of SMEs by leveraging the relational information in accessible financial networks.
% Besides, hypergraph can also promote prediction of enterprise risk \cite{zhao2022fisrebp}.
% Zhang et al. \cite{zhang2022heterogeneous} model bank data as a heterogeneous information network and thoroughly mine the relations existing in the financial activities of small and micro-enterprise users under the commercial banking service scenario to assist in default analysis
% \cite{Cheng2022Subsequence}.

\subsubsection{Individual Aspect}

\noindent \textit{\textbf{Credit Risk Assessment.}} 
% Credit risk assessment aims to detect and identify users' potential fraud or default. Cheng et al. \cite{cheng2020graph} propose a spatial-temporal graph attention network method for credit card fraud detection depending on time and location transaction information, which jointly learns attention weights with 3D convolutional network.
% Liu et al. \cite{liu2021pick} focus on the problem of fraud detection with the imbalanced class on the graph, adopting a supervised GNN method combined with a specific sampling strategy. They leverage a label-balanced sampler to establish sub-graphs and aggregate information from choosed neighborhood to form the representation of the target nodes.
% %Based on the bank's user credit card data, Ji et al. \cite{ji2022detecting} leverage a graph model to detect the cash-out fraud behaviors in the credit card business of commercial banks. 
% Chen and Tsourakakis \cite{chen2022antibenford} design an AntiBenford subgraph framework and corresponding algorithm based on Benford's law, detecting abnormal subgraphs on the cryptocurrency transaction network and identifing abnormal transaction data.
Credit risk assessment endeavors to discern and characterize potential instances of fraud or default among users. Cheng et al. \cite{cheng2020graph} propose a spatial-temporal graph attention network method for credit card fraud detection. This method relies on transaction information involving both time and location, wherein attention weights are jointly learned in conjunction with a 3D convolutional network.
In addressing the challenge of fraud detection within imbalanced class structures on graphs, Liu et al. \cite{liu2021pick} adopt a supervised GNN method. Their approach is complemented by a specific sampling strategy, incorporating a label-balanced sampler to establish sub-graphs. Information from chosen neighborhoods is aggregated to formulate the representation of target nodes.
Furthermore, Chen and Tsourakakis \cite{chen2022antibenford} design an AntiBenford subgraph framework and corresponding algorithm grounded in Benford's law. This framework facilitates the detection of abnormal subgraphs within cryptocurrency transaction networks, thereby enabling the identification of anomalous transaction data.

\subsection{Datasets}
We provide a summary of the most popular datasets in Table \ref{tab:dataset}. Within recommendation systems, the representation of users, items, and their interactions manifests as heterogeneous nodes and edges. This formulation engenders diverse graph types, encompassing bi-typed graphs, hypergraphs, and dynamic graphs. Notably, social networks often exhibit homogeneity, while networks in fintech and academic domains typically manifest heterogeneity in both nodes and edges.
In the domains of NLP and CV, graph construction occurs indirectly, rendering the number of nodes and edges contingent upon experimental settings. Conversely, knowledge graphs, a specialized category within NLP, are characterized by knowledge triples. Finally, in the realms of chemistry and biology, both nodes and edges within graphs are endowed with attributes, a distinctive feature that sets them apart from other datasets.

% We provide a summary of the most popular datasets in Table \ref{tab:dataset}.
% % for different scenarios and aim to encourage further research.
% In recommendation systems, users, items, and their interactions are represented as heterogeneous nodes and edges, resulting in various graph types such as bi-typed graphs, hypergraphs, and dynamic graphs. Social networks often exhibit homogeneity, while networks in fintech and academic settings tend to be heterogeneous on both nodes and edges.
% For NLP and CV scenarios, graphs are constructed indirectly, so the number of nodes and edges depends on the experiment settings. In contrast, knowledge graphs, a specific type in NLP, consist of knowledge triples. Finally, in chemistry and biology, both nodes and edges in those graphs usually have attributes, which distinguish them from other datasets.

\begin{table*}
\begin{threeparttable}
    \setlength{\abovecaptionskip}{0cm}
    \renewcommand\arraystretch{1.2}
    \setlength\tabcolsep{5pt}
    \caption{A table of data summary}\label{tab:tablenotes}
    \centering
    \label{tab:dataset}
    % \begin{tabular}{lcccp{7.5cm}}
    \begin{tabular}{p{2cm}p{2.3cm}p{1.4cm}p{1.4cm}p{9.2cm}}
    \toprule[1.5pt]  
        \textbf{Scenarios} & \textbf{Dataset} & \textbf{Nodes} & \textbf{Edges} &  \textbf{Description} \\ 
        \midrule
        \multirowcell{4}{Recommendation \\ System} 
        & Amazon \cite{zhang2022dynamic} & $2.098 \times 10^7$ & $8.283 \times 10^7$ & This is a large crawl of product reviews from Amazon. \\
        % & Movielens \cite{Cao2021Bipartite} & 80,555 & 10,000,054 & This dataset describes people’s expressed preferences for movies.\\ 
        & IMDB \cite{Tang2015Pte} & 1,199,919 & 3,782,463 & A binary sentiment analysis dataset. \\ 
        & Flickr \cite{Zhang2015Cosnet} & 499,610 & 8,545,307 & A popular photo sharing network on Flicker. \\ & LastFM \cite{Zhang2015Cosnet} & 136,420 & 1,685,524 & Social network of LastFM users from Asia.\\ 
        \midrule
        \multirowcell{5}{NLP} & WordNet18  \cite{Schlichtkrull2018Modeling} & 146,005 & 656,999 & 
        % Also called WN18, 
        This dataset includes 18 relations scraped from WordNet.\\
        & DBpedia \cite{zhang2021weakly} & 3,966,924 & 13,820,853 & Structured information from Wikipedia.\\
        & FB15k \cite{Schlichtkrull2018Modeling} & 14,951 & 1,345 &
        % Also called Freebase15K, 
        KB relation triples and textual mentions of Freebase entity pairs.\\
        & NELL \cite{Wan2020Reinforcement} & $ 1.7 \times 10^6 $ & $ 2.4 \times 10^6 $ & This dataset offers the common sense knowledge rules.\\
        & YAGO \cite{Wan2020Reinforcement} & $ 1 \times 10^6 $ & $ 1.2 \times 10^8 $ & A KG that augments WordNet.\\
        \midrule
        \multirowcell{5}{CV\tnote{1}} 
        & Visual Genome \cite{Xu2017Scene} &38 &22 & This dataset connects language and vision using crowdsourced dense image.\\
        & COCO \cite{Hu2018Relation}&- & -& An object detection, segmentation, key-point detection, and captioning dataset.\\
        % & USPS \cite{Chen2015Discrete} & - & - & A database for handwritten text recognition research.\\
        & CAD \cite{Jain2016Structural} & 40 & - & A dataset comprised of video sequences of humans performing activities.\\
        % & SVHN \cite{cubuk2018autoaugment} & -&- & A digit classification benchmark dataset.\\
        % & AFW \cite{feng2021detect} & 5,171& & A face detection dataset.\\
        & INRIA Person\cite{hu2021naturalistic} &- & -& A dataset of images of persons used for pedestrian detection.\\
        & Cityscapes \cite{chen2018searching} &- &- & A database focusing on semantic understanding of urban street scenes.\\
        % & KITTI \cite{mahjourian2018unsupervised} &- &- & A dataset used in mobile robotics and autonomous driving.\\
        \midrule
        \multirowcell{3}{Fintech} 
        & SMEsD \cite{zhao2022fisrebp} & 3,976 & 21,818 & A multiplex enterprise KG consisting of SMEs and related persons in China.\\
        & DGraph \cite{huang2022dgraph} & 3,700,550 & 4,300,999 & This graph represents a realistic user-to-user social network in financial industry. \\
        & eBay-small \cite{rao2021xfraud} & $2.89 \times 10^5$ & $6.13 \times 10^5 $ & A real transaction network on eBay for fraud detection.\\
        %  in China from 2014 to 2021
        % & YelpChi \cite{liu2021pick} & 45,954 & 3,846,979 & 
        % Also called Yelp-Fraud, 
        % A multi-relational graph dataset built upon the Yelp spam review.\\
        % & Amazon \cite{liu2021pick} & 11,944 & 4,398,392 & 
        % Also called Amazon Fraud, 
        % A multi-relational graph dataset built upon the Amazon review dataset.\\
        \midrule
        \multirowcell{7}{Chemistry \\ and \\ Biology }
        & MUTAG	\cite{Schlichtkrull2018Modeling} & $ 1.793 \times 10^4 $ & $ 1.979 \times 10^4 $	& A collection of nitroaromatic compounds.\\
        & NCI1 \cite{Pan2017Task} & $ 2.987 \times 10^4 $ & $ 3.23 \times 10^4 $ & A dataset related to anti-cancer screening.\\
        & PTC \cite{Pan2017Task} & $ 1.456 \times 10^4 $ & $ 1.5 \times 10^4 $  & A collection of chemical compounds reporting the carcinogenicity for rats.\\
        & Tox21 \cite{Rong2020Self} & $ 2.235 \times 10^4 $ & $ 2.332 \times 10^4 $ & A dataset represents chemical compounds.\\
        & ENZYMES \cite{Ying2018Hierarchical} & $ 3.263 \times 10^4 $ & $ 6.214 \times 10^4 $ &  A dataset of protein tertiary structures.\\
        & PROTEINS \cite{Ying2018Hierarchical} & $ 3.906 \times 10^4 $ & $ 7.282 \times 10^4 $ & A dataset of proteins classified as enzymes or non-enzymes.\\
        % & QM9 \cite{Cao2018MolGAN} & $ 1.803 \times 10^4 $ & $ 1.863 \times 10^4 $ & Contains 133,885 organic compounds up to 9 heavy atoms.\\
        \midrule
        \multirowcell{6}{Academic}
        & Pubmed \cite{Kipf2016Semi-supervised} & 8,341,043 & 737,869,083 & This dataset consists of publications pertaining to diabetes.\\
        & DBLP \cite{Tang2015Pte} & 317,080 & 1,049,866 & DBLP collaboration network.\\
        % & Citeseer \cite{Yang2015Network} & 384,413 & 1,751,463 & A dataset consists of 384,413 scientific publications.\\
        % & Cora \cite{Jiang2019Dynamic} & 23,166 & 91,500 & A dataset consists of 23,166 scientific publications classified into seven classes.\\
        & MAG240M \cite{Chiang2019Cluster} & 244,160,499 & 1,728,364,232 & An academic graph extracted from Open Graph Benchmark.\\
        % & MAG240M \cite{Chiang2019Cluster-gcn} & 244,160,499 & 1,728,364,232 & An academic graph extracted from Open Graph Benchmark.\\
        & OAG \cite{Hu2020Heterogeneous} & 1,858,395 & 3,965,744 & A large Knowledge Graph unifying Microsoft Academic Graph and Aminer.\\
        & AMiner \cite{dong2017metapath2vec} & 1,369,055 & 8,650,089 & A large academic network.\\
        \midrule
        \multirowcell{4}{Social \\ Network}
        % & Epinions \cite{JunfaLin2022GraphNN} & 876,252 & 13,668,320 & Who-trusts-whom network of Epinions.com.\\
	    & Google+ \cite{Zhang2017User} & 28,943,739 & 462,994,069 & A network contains a snapshot of the Google+ social structure.\\
	    & YouTube \cite{Chen2017Supervised} & 3,223,589 & 9,375,374 & A network of YouTube users and their friendship connections.\\
	    & Twitter \cite{Xu2018Exploring} & 465,017 & 834,797 & A dataset containing information about who follows whom on Twitter.\\
	    & Livejournal \cite{Zhang2015Cosnet} & 4,847,571 & 68,993,773 & LiveJournal online social network.\\
	    % & Foursquare \cite{Zhan2015Influence} & 2,153,471 & 27,098,490 & A dataset consists of spatial user ratings.\\

\bottomrule[1.5pt]
\end{tabular}
\footnotesize
\begin{tablenotes} 
\item[1] The numbers of nodes and links in the CV part are counted per image on average.
\end{tablenotes}
\end{threeparttable}
\end{table*}

\section{Future Directions}
\label{section-future-directions}
Despite the achievements of GL, many challenges remain unsolved. In this section, we summarize major trends and challenges, which also serve as future directions of GL. In particular, we identify six directions that are worth considering: graph foundation model, integration of GL and LLM,
graph scalability, interpretability, Over-smoothing and social effects.

\subsection{Graph Foundation Model}
Foundation models \cite{bommasani2021opportunities}, trained on vast amounts of data, demonstrate the ability to perform a wide range of downstream tasks. In NLP, models like LLMs, and in computer vision (CV), models such as SAM \cite{kirillov2023segment} and SpectralGPT \cite{hong2024spectralgpt}, have outperformed traditional models that focus on specific tasks. This phenomenon sheds light on the advancements in graph learning, leading to the emergence of promising models like OpenGraph \cite{xia2024opengraph} and InstructGLM \cite{ye2023natural}. 
Interestingly, OpenGraph and InstructGLM take two fundamentally different approaches to learning graph foundation models. OpenGraph, following a more traditional route, first introduces a unified graph tokenizer and a scalable graph transformer to learn node representations, then utilizes LLMs to augment the data. In contrast, InstructGLM proposes fine-tuning LLMs to enable them to comprehend graph structures described in natural language, thereby allowing the model to perform a variety of downstream graph machine learning tasks.

However, unlike CV and NLP, graphs are non-Euclidean and exhibit locality, making them fundamentally different. This raises several open research questions for graph foundation models, including the scaling laws for graph foundation models, whether pure graph learning models are more effective, or if combining graph learning with LLMs provides superior performance.

\subsection{Integration of GL and LLM}
Numerous studies have considered both GL and LLMs, which can be categorized into two main approaches: GL-centered or LLM-centered. GL-centered approaches often fail to fully exploit the capabilities of LLMs, typically using them merely as improved tokenizers for node or edge attributes. On the other hand, LLM-centered approaches face limitations by primarily leveraging graph structure without incorporating advanced graph learning techniques. Fortunately, research on Retrieval-augmented Inference (refer to Section \ref{subsubsec-kg-augmented-llms}) offers a promising pathway for simultaneously harnessing both GL and LLMs. Additionally, integrating GL, LLMs, and agents into a unified framework is a critical direction for future research.

\subsection{Graph Scalability}
Graph neural networks have demonstrated remarkable efficacy in extracting insights from graph-structured data, garnering widespread utilization across diverse domains. Nevertheless, graphs encountered in certain domains pose unique challenges, characterized by both extensive size and heterogeneity, often comprising millions or even billions of vertices and edges of distinct types. To confront this challenge, diverse training models have been advanced, specifically tailored for large-scale graphs. In this manuscript, we delineate prevalent strategies for implementing few-shot learning, with a dedicated emphasis on two pivotal aspects.

% Graph neural networks have achieved great success in learning from graph-structured data. They are widely used in a variety of domains, but graphs in partial domains are usually large and heterogeneous, containing millions or billions of vertices and edges of different types. To tackle this challenge, some training models on large-scale graphs have been proposed. We introduce common approaches to implementing few-shot learning from the following two aspects.
% Graph neural networks have proven highly effective in learning from graph-structured data and are now widely used across various domains. However, graphs in some domains can be both large and heterogeneous, often containing millions or even billions of vertices and edges of different types. To address this challenge, various training models have been proposed for large-scale graphs. In this article, we introduce common approaches for implementing few-shot learning, focusing on two key aspects.

\subsubsection{Mini-batch Training}
% Mini-batch training is commonly used to solve the training problem on large-scale graphs, and only a part of the graph nodes are selected for training each time. In mini-batch training, people have designed various neighborhood sampling methods, among which the most basic method is node sampling \cite{Hamilton2017Inductive} %\cite{,feng2022grand+}
% and the common ones are layer sampling \cite{Chen2018Fastgcn} 
% % \cite{min2022graph,ran2022cryptogcn}
% % \cite{Huang2018Adaptive,}
% and subgraph sampling \cite{Hu2020Heterogeneous}. 
% \cite{lin2022resource,zeng2020graphsaint,zheng2022distributed}
Mini-batch training is a widely used technique for training large-scale graph data, where only a part of the graph nodes are selected for training each time. Several neighborhood sampling methods have been proposed for mini-batch training, including the most basic method of node sampling \cite{Hamilton2017Inductive}, as well as layer sampling \cite{Chen2018Fastgcn} and subgraph sampling \cite{Hu2020Heterogeneous}, which are commonly used.

\subsubsection{Embedding Compression}
Node embeddings serve a crucial role in diverse downstream tasks, yet the computational challenges arising from the linear increase in parameters with the number of nodes necessitate effective solutions. In response to this issue, various embedding compression methods have been introduced. Notably, Yeh et al. \cite{yeh2022embedding} propose a compression technique that represents each node compactly using a bit vector, as opposed to a floating-point vector. Remarkably, the parameters of this compression method can be concurrently trained with graph neural networks, as demonstrated by Liu et al. \cite{liu2021exact}.
Concurrently, Cui et al. \cite{cui2022allie} adopt a reinforcement learning agent incorporating an imbalance-aware reward function. This agent strategically samples from both majority and minority classes, subsequently applying a graph coarsening strategy to reduce the search space of the agent.

\subsection{Interpretability}
% https://www.graphframex.com/

% Causal Attention for Interpretable and Generalizable Graph Classification

% CLEAR: Generative Counterfactual Explanations on Graphs

% Debiasing Graph Neural Networks via Learning Disentangled Causal Substructure

% Explaining Graph Neural Networks with Structure-Aware Cooperative Games

% Task-Agnostic Graph Explanations

% AgraSSt: Approximate Graph Stein Statistics for Interpretable Assessment of Implicit Graph Generators

% The problem of machine learning interpretability is divided from the perspective of interpretable objects. The first is the post-hoc interpretable model for the existing models

The challenge of interpretability in machine learning is bifurcated based on the perspective of interpretable objects. The first facet involves post-hoc interpretable models designed for existing models \cite{ying2019gnnexplainer}. The second facet entails pre-event interpretability models exhibiting both exceptional performance and interpretability in downstream tasks \cite{agarwal2021neural}. The interpretability of graph structure data, which incorporates not only node features but also network topology, distinguishes it from conventional machine learning interpretability methods.
Post-hoc interpretability presently encompasses two key levels: instance level and model level. Amara et al. \cite{amara2022graphframex} introduce a framework to evaluate the explanations provided by current methods.

\subsubsection{Instance-level interpretability} 
Instance-level interpretability concentrates on elucidating the interpretability of classification outcomes for a given node. Ying et al. \cite{ying2019gnnexplainer} propose a methodology for identifying influential nodes, node characteristics, and critical pathways significantly impacting the classification of a specific node.
Dai et al. \cite{dai2021towards} present an approach that concurrently provides predictions and explanations. Specifically, they utilize K-nearest labeled nodes for each unlabeled node to furnish node classification explanations.
Distinguishing itself from existing explainers for GNNs, DIR-GNNs \cite{wu2022discovering} approach GNN interpretability on graph-structured data from a causal perspective.

\subsubsection{Model-level interpretability} 
Model-level interpretability directs its focus towards discerning global and generic patterns, treating them as inherent logic governing graph neural network classification. Illustratively, Luo et al. \cite{luo2020parameterized} employ RL to iteratively produce subgraphs as global interpretability outcomes. Similarly, Yuan et al. \cite{yuan2020xgnn} utilize RL to train a graph generator, ensuring that the generated graph patterns optimize specific predictions made by GNNs. Notably, PGExplainer \cite{luo2020parameterized} advocates for the parameterization of the explanation generation process, thereby enabling the model to collectively elucidate multiple instances. This innovative approach enhances the interpretability of GNNs by capturing overarching patterns that contribute to their classification decisions.

% Model-level interpretability focuses on global, generic patterns and treats them as intrinsic logic for graph neural network classification.
% % Yuan et al. \cite{yuan2022explainability} and 
% For instance, Luo et al. \cite{luo2020parameterized} use RL to iteratively generate subgraphs as the global interpretability result.
% % \cite{Zhang2019Heterogeneous} gives an interpretable subgraph structure for a given graph neural network model classification results by means of a Monte Carlo tree search.
% Yuan et al. \cite{yuan2020xgnn} also use RL to train a graph generator so that the generated graph patterns maximize a certain prediction of the GNNs.
% PGExplainer \cite{luo2020parameterized}
% propose to parameterize the generation process of explanations. Thus the model can explain multiple instances collectively.

% There are also some researchers implement the transparency interpretability of graph neural network models, for example, Dai et al. \cite{dai2021towards} 
% % and Liu et al. \cite{liu2019towards} 
% propose a new framework which can find K-nearest labeled nodes for each unlabeled node to give explainable node classification.
% Other scholars have applied the interpretability of graph neural networks to different domains, Schnake et al. \cite{schnake2021higher} use the GNN-LRP model to test under different scenarios of the graph neural networks, including text emotion classification, quantum chemistry, and image classification.

\subsection{Over-smoothing}
% \subsubsection{Over-smoothing} 
The issue of over-smoothing poses a significant challenge in GNNs. This problem arises due to the common assumption in GNNs that nodes with similar features tend to be connected, leading to the convergence of node representations after multiple convolutional layers, commonly known as over-smoothing. To address this challenge, two primary methods have been employed.

\noindent \textit{\textbf{i. Model-specific Methods}}

Model-specific methods focus on designing specialized anti-over-smoothing modules tailored to specific GNN models. For example, Klicpera et al. \cite{Klicpera2018Predict} integrate the PageRank algorithm with GCN to enhance traditional propagation methods, effectively reducing the number of parameters. This modification enlarges and adjusts the neighborhood, alleviating over-smoothing. Chen et al. \cite{chen2020simple2} introduce GCNII, incorporating initial residual and identity mapping mechanisms to maintain model effectiveness with increased depth.

\noindent \textit{\textbf{ii. General Methods}}

General methods offer greater flexibility, portability, and scalability, as they are not model-specific. Common strategies involve regularization to mitigate over-smoothing effects. Do et al. \cite{do2021graph} propose a novel regularization method called DropNode, which randomly removes portions of the graph to decrease overall graph connectivity. Zeng et al. \cite{zeng2021decoupling} employ a subgraph extractor to exclude noisy nodes, forming a subgraph composed of a small number of crucial neighbor nodes. This subgraph is then transformed into informative representations to prevent over-smoothing across the entire graph.

\subsection{Social Effects}
\subsubsection{Fairness}
GNNs have exhibited substantial promise in modeling data structured as graphs. However, akin to other machine learning models, GNNs are susceptible to making predictions influenced by protected sensitive attributes such as skin color and gender. Biases can manifest in two forms within graph-structured data: one mirrors biases prevalent across various data types like tables and texts, while the other is specific to the characteristics of graph-structured data. These biases pose a threat to the fairness of GNNs, introducing potential shortcomings in their predictions.

In addressing fairness concerns in GNNs, various algorithms can be categorized into adversarial debiasing \cite{Dong2022Edits}, fairness constraints \cite{Song2022Guide}, and other methodologies \cite{Zhang2022Fairness}. Dong et al. \cite{Dong2022Edits} introduce novel definitions and metrics for quantifying bias in attributed networks. This leads to an optimization objective aimed at mitigating bias, resulting in GNNs trained on less biased data. Song et al. \cite{Song2022Guide} propose GUIDE, a GNN framework operating on individual similarity matrices to learn personalized attention, thereby achieving individual fairness while mitigating group-level disparities. Zhang et al. \cite{Zhang2022Fairness} explore a more general scenario involving quantities of unlabeled data, introducing a novel learning paradigm known as fair semi-supervised learning.

% GNNs have demonstrated significant potential in modeling graph-structured data. However, like other machine learning models, GNNs may make predictions based on protected sensitive attributes, e.g., skin color and gender. Two kinds of biases may exist in graph-structured data: one is the bias that exists widely in various types of data such as tables and texts; The other is the bias characteristic of graph-structured data. These biases may compromise the fairness of GNNs, leading to flaws in their predictions.

% For GNNs fairness algorithms, we classify them into adversarial debiasing \cite{Dong2022Edits}, fairness constraints
% \cite{Song2022Guide}, and other methods \cite{Zhang2022Fairness}. 
% Dong et al. \cite{Dong2022Edits} propose novel definitions and metrics to measure the bias in an attributed network, which leads to the optimization objective to mitigate bias, feeding GNNs with less biased data. Song et al. \cite{Song2022Guide} propose a novel GNN framework, GUIDE, which operates on the similarity matrix of individuals to learn personalized attention to achieve individual fairness without group level disparity. Zhang et al. \cite{Zhang2022Fairness} explore a more general case where quantities of unlabeled data are provided, leading to a new form of learning paradigm, namely fair semi-supervised learning.

\subsubsection{Privacy}
Preserving the privacy of sensitive information within model training sets, particularly in the context of model publication or service provision, is a critical concern. Scholars have dedicated efforts to explore defenses against various privacy attacks. For instance, Hsieh et al. \cite{Hsieh2021Netfense} introduce NetFense, a graph perturbation-based approach designed to counter privacy attacks targeting GNNs. NetFense simultaneously ensures the unnoticeability of graph data, maintains the prediction confidence for targeted label classification, and diminishes the prediction confidence associated with private label classification.
Wu et al. \cite{Wu2022Federated} propose FedPerGNN, a federated GNN framework tailored for privacy-preserving recommendation. This framework leverages decentralized graph mining techniques to enhance privacy safeguards.

% The private information in the model training set may be leaked in the published model or the provided service. The defense against different privacy attacks is the research direction of many scholars. Hsieh et al. \cite{Hsieh2021Netfense} present a graph perturbation-based approach, NetFense, to against GNN-based privacy attacks. NetFense can simultaneously keep graph data unnoticeability, maintain the prediction confidence of targeted label classification, and reduce the prediction confidence of private label classification. 
% % And some researches \cite{Wu2022Federated,Yang2020Graph} present GNN frameworks for both effective and privacy-preserving personalization.
% Wu et al. \cite{Wu2022Federated} propose FedPerGNN, a federated GNN framework for privacy preserving recommendation, which utilize decentralized graph mining.

% fairness
% privacy
% equivariant graph neural networks

% Improving Fairness in Graph Neural Networks via Mitigating Sensitive Attribute Leakage

% equivariant graph neural networks
% Learning Physical Dynamics with Subequivariant Graph Neural Networks
% Learning Invariant Graph Representations Under Distribution Shifts
% Equivariant Graph Hierarchy-based Neural Networks

% stable

% \subsection{GNNs' application on various scenarios}
% Fintech
% Recommendation system
% Physics
% Computer vision
% \subsection{Continues dynamic graph}

% \subsection{Graph Compression?}

% \subsection{Graph Entropy?}

\section{Conclusion}
\label{section-conclusion}
Graphs are vital for depicting complex real-world relationships. Deep learning techniques have greatly improved GL, accommodating various graph types: social networks, textual, visual, and tabular data-derived graphs. Despite existing surveys, a comprehensive overview of this rapidly evolving field is lacking. Our survey fills this gap by meticulously analyzing recent works, categorizing them based on fundamental graph elements and methods. We also explore the integration of GL with LLMs, focusing on PLMs and their evolution in conjunction with GL. Current applications across diverse scenarios are consolidated, and future directions for research in graph learning are presented to stimulate further advancements.

\section*{Acknowledgments}
% This should be a simple paragraph before the References to thank those individuals and institutions who have supported your work on this article.

The research is supported by the Key Technologies Research and Development Program under Grant No. 2020YFC0832702, 
and National Natural Science Foundation of China under Grant Nos. 71910107002, 62376227, 61906159, 62302400, 62176014, and Sichuan Science and Technology Program under Grant No. 2023NSFSC0032, 2023NSFSC0114, and Guanghua Talent Project of Southwestern University of Finance and Economics.

\bibliographystyle{IEEEtran}
\bibliography{sample-base}

\begin{thebibliography}{100}
\providecommand{\url}[1]{#1}
\csname url@samestyle\endcsname
\providecommand{\newblock}{\relax}
\providecommand{\bibinfo}[2]{#2}
\providecommand{\BIBentrySTDinterwordspacing}{\spaceskip=0pt\relax}
\providecommand{\BIBentryALTinterwordstretchfactor}{4}
\providecommand{\BIBentryALTinterwordspacing}{\spaceskip=\fontdimen2\font plus
\BIBentryALTinterwordstretchfactor\fontdimen3\font minus \fontdimen4\font\relax}
\providecommand{\BIBforeignlanguage}[2]{{%
\expandafter\ifx\csname l@#1\endcsname\relax
\typeout{** WARNING: IEEEtran.bst: No hyphenation pattern has been}%
\typeout{** loaded for the language `#1'. Using the pattern for}%
\typeout{** the default language instead.}%
\else
\language=\csname l@#1\endcsname
\fi
#2}}
\providecommand{\BIBdecl}{\relax}
\BIBdecl

\bibitem{Mcpherson2001Birds}
M.~McPherson, L.~Smith-Lovin, and J.~M. Cook, ``Birds of a feather: Homophily in social networks,'' \emph{Annual Review of Sociology}, vol.~27, no.~1, pp. 415--444, 2001.

\bibitem{Zhang2015Cosnet}
Y.~Zhang, J.~Tang, Z.~Yang, J.~Pei, and P.~S. Yu, ``Cosnet: Connecting heterogeneous social networks with local and global consistency,'' in \emph{KDD}, 2015, pp. 1485--1494.

\bibitem{Chiang2019Cluster}
W.-L. Chiang, X.~Liu, S.~Si, Y.~Li, S.~Bengio, and C.-J. Hsieh, ``Cluster-gcn: An efficient algorithm for training deep and large graph convolutional networks,'' in \emph{KDD}, 2019, pp. 257--266.

\bibitem{dong2017metapath2vec}
Y.~Dong, N.~V. Chawla, and A.~Swami, ``metapath2vec: Scalable representation learning for heterogeneous networks,'' in \emph{KDD}, 2017, pp. 135--144.

\bibitem{Hu2020Heterogeneous}
Z.~Hu, Y.~Dong, K.~Wang, and Y.~Sun, ``Heterogeneous graph transformer,'' in \emph{WWW}, 2020.

\bibitem{JiliangTang2013SocialRA}
J.~Tang, X.~Hu, and H.~Liu, ``Social recommendation: a review,'' \emph{Social Network Analysis and Mining}, vol.~3, no.~4, pp. 1113--1133, Jan. 2013.

\bibitem{WangHao2022HyperSoRecEH}
W.~Hao, L.~Defu, T.~Hanghang, L.~Qi, H.~Zhenya, and C.~Enhong, ``Hypersorec: Exploiting hyperbolic user and item representations with multiple aspects for social-aware recommendation,'' \emph{ACM TOIS}, 2022.

\bibitem{cheng2020delinquent}
D.~Cheng, Z.~Niu, and L.~Zhang, ``Delinquent events prediction in temporal networked-guarantee loans,'' \emph{TNNLS}, 2020.

\bibitem{yang2021financial}
S.~Yang, Z.~Zhang, J.~Zhou, Y.~Wang, W.~Sun, X.~Zhong, Y.~Fang, Q.~Yu, and Y.~Qi, ``Financial risk analysis for smes with graph-based supply chain mining,'' in \emph{IJCAI}, 2021, pp. 4661--4667.

\bibitem{zhang2022heterogeneous}
Z.~Zhang, Y.~Ji, J.~Shen, X.~Zhang, and G.~Yang, ``Heterogeneous information network based default analysis on banking micro and small enterprise users,'' \emph{arXiv}, 2022.

\bibitem{Hu2018Relation}
H.~Hu, J.~Gu, Z.~Zhang, J.~Dai, and Y.~Wei, ``Relation networks for object detection,'' \emph{CVPR 2018}, vol.~2, no.~3, pp. 3588--3597, 2018.

\bibitem{Sarlin2020Superglue}
P.-E. Sarlin, D.~DeTone, T.~Malisiewicz, and A.~Rabinovich, ``Superglue: Learning feature matching with graph neural networks,'' in \emph{CVPR}, 2020, pp. 4938--4947.

\bibitem{yasunaga2021qa}
M.~Yasunaga, H.~Ren, A.~Bosselut, P.~Liang, and J.~Leskovec, ``Qa-gnn: Reasoning with language models and knowledge graphs for question answering,'' \emph{arXiv}, 2021.

\bibitem{Yao2018Graph}
L.~Yao, C.~Mao, and Y.~Luo, ``Graph convolutional networks for text classification,'' in \emph{AAAI}, vol.~33, no.~01, 2019, pp. 7370--7377.

\bibitem{Schlichtkrull2018Modeling}
M.~Schlichtkrull, T.~N. Kipf, P.~Bloem, R.~van~den Berg, I.~Titov, and M.~Welling, ``Modeling relational data with graph convolutional networks,'' in \emph{ESWC}, 2018, pp. 593--607.

\bibitem{Rong2020Self}
Y.~Rong, Y.~Bian, T.~Xu, W.~Xie, Y.~Wei, W.~Huang, and J.~Huang, ``Self-supervised graph transformer on large-scale molecular data,'' in \emph{NeurIPS}, 2020.

\bibitem{Fouss2007Random}
F.~Fouss, A.~Pirotte, J.-M. Renders, and M.~Saerens, ``Random-walk computation of similarities between nodes of a graph with application to collaborative recommendation,'' \emph{IEEE Transactions on knowledge and data engineering}, vol.~19, no.~3, pp. 355--369, 2007.

\bibitem{kuang2012symmetric}
D.~Kuang, C.~Ding, and H.~Park, ``Symmetric nonnegative matrix factorization for graph clustering,'' in \emph{Proceedings of the 2012 SIAM international conference on data mining}.\hskip 1em plus 0.5em minus 0.4em\relax SIAM, 2012, pp. 106--117.

\bibitem{goyal2018graph}
P.~Goyal and E.~Ferrara, ``Graph embedding techniques, applications, and performance: A survey,'' \emph{KBS}, vol. 151, pp. 78--94, 2018.

\bibitem{Zhou2020Graph}
J.~Zhou, G.~Cui, S.~Hu, Z.~Zhang, C.~Yang, Z.~Liu, L.~Wang, C.~Li, and M.~Sun, ``Graph neural networks: A review of methods and applications,'' \emph{AI Open}, vol.~1, pp. 57--81, 2020.

\bibitem{Wu2020Comprehensive}
Z.~Wu, S.~Pan, F.~Chen, G.~Long, C.~Zhang, and S.~Y. Philip, ``A comprehensive survey on graph neural networks,'' \emph{TNNLS}, vol.~32, no.~1, pp. 4--24, 2020.

\bibitem{liu2022graph}
Y.~Liu, M.~Jin, S.~Pan, C.~Zhou, Y.~Zheng, F.~Xia, and P.~Yu, ``Graph self-supervised learning: A survey,'' \emph{TKDE}, 2022.

\bibitem{ren2024survey}
X.~Ren, J.~Tang, D.~Yin, N.~Chawla, and C.~Huang, ``A survey of large language models for graphs,'' in \emph{Proceedings of the 30th ACM SIGKDD Conference on Knowledge Discovery and Data Mining}, 2024, pp. 6616--6626.

\bibitem{Xu2007Scan}
X.~Xu, N.~Yuruk, Z.~Feng, and T.~A. Schweiger, ``Scan: a structural clustering algorithm for networks,'' in \emph{KDD}, 2007, pp. 824--833.

\bibitem{Sun2011Pathsim}
Y.~Sun, J.~Han, X.~Yan, P.~S. Yu, and T.~Wu, ``Pathsim: Meta path-based top-k similarity search in heterogeneous information networks,'' in \emph{VLDB Endowment}, vol.~4, no.~11, 2011, pp. 992--1003.

\bibitem{micheli2005new}
A.~Micheli and A.~S. Sestito, ``A new neural network model for contextual processing of graphs,'' in \emph{Neural Nets}.\hskip 1em plus 0.5em minus 0.4em\relax Springer, 2005, pp. 10--17.

\bibitem{Scarselli2008The}
F.~Scarselli, M.~Gori, A.~C. Tsoi, and G.~Monfardini, ``The graph neural network model,'' \emph{TNNLS}, vol.~20, no.~1, pp. 61--80, 2008.

\bibitem{Karypis2000Multilevel}
G.~Karypis and V.~Kumar, ``Multilevel k-way hypergraph partitioning,'' \emph{VLSI design}, vol.~11, no.~3, pp. 285--300, 2000.

\bibitem{Kempe2003Maximizing}
D.~Kempe, J.~Kleinberg, and {\'E}.~Tardos, ``Maximizing the spread of influence through a social network,'' in \emph{KDD}, 2003, pp. 137--146.

\bibitem{Mcauley2012Learning}
J.~J. McAuley and J.~Leskovec, ``Learning to discover social circles in ego networks,'' in \emph{NeurIPS}, "2012", pp. 548--56.

\bibitem{Tang2015Pte}
J.~Tang, M.~Qu, and Q.~Mei, ``Pte: Predictive text embedding through large-scale heterogeneous text networks,'' in \emph{KDD}, 2015.

\bibitem{Cao2015Grarep}
S.~Cao, W.~Lu, and Q.~Xu, ``Grarep: Learning graph representations with global structural information,'' in \emph{CIKM}, 2015, pp. 891--900.

\bibitem{Yang2016Revisiting}
Z.~Yang, W.~Cohen, and R.~Salakhudinov, ``Revisiting semi-supervised learning with graph embeddings,'' in \emph{ICML}, 2016, pp. 40--48.

\bibitem{Grover2016Node2vec}
A.~Grover and J.~Leskovec, ``Node2vec: Scalable feature learning for networks,'' in \emph{KDD}, 2016, pp. 855--864.

\bibitem{Kipf2016Semi-supervised}
T.~N. Kipf and M.~Welling, ``Semi-supervised classification with graph convolutional networks,'' in \emph{ICLR}, 2016.

\bibitem{Kipf2016Variational}
T.~N. Kipf and W.~Max, ``Variational graph auto-encoders,'' \emph{arXiv}, 2016.

\bibitem{Hamilton2017Inductive}
W.~L. Hamilton, R.~Ying, and J.~Leskovec, ``Inductive representation learning on large graphs,'' in \emph{NeurIPS}, 2017, pp. 1025--1035.

\bibitem{Chen2018Fastgcn}
J.~Chen, T.~Ma, and C.~Xiao, ``Fastgcn: fast learning with graph convolutional networks via importance sampling,'' in \emph{ICLR}, 2018.

\bibitem{Velickovic2018Graph}
P.~Velickovic, G.~Cucurull, A.~Casanova, A.~Romero, P.~Lio, and Y.~Bengio, ``Graph attention networks,'' in \emph{Proceddings of ICLR}, 2018.

\bibitem{Tu2018Structural}
K.~Tu, P.~Cui, X.~Wang, F.~Wang, and W.~Zhu, ``Structural deep embedding for hyper-networks,'' in \emph{AAAI}, 2018.

\bibitem{Zhou2018Density}
Y.~Zhou, S.~Wu, C.~Jiang, Z.~Zhang, D.~Dou, R.~Jin, and P.~Wang, ``Density-adaptive local edge representation learning with generative adversarial network multi-label edge classification,'' in \emph{ICDM}, 2018, pp. 1464--1469.

\bibitem{Wang2019Heterogeneous}
X.~Wang, H.~Ji, C.~Shi, B.~Wang, P.~Cui, P.~Yu, and Y.~Ye, ``Heterogeneous graph attention network,'' in \emph{WWW}, 2019, pp. 2022--2032.

\bibitem{Lee2019Self-Attention}
J.~Lee, I.~Lee, and J.~Kang, ``Self-attention graph pooling,'' in \emph{ICML}, 2019.

\bibitem{sankar2020dysat}
A.~Sankar, Y.~Wu, L.~Gou, W.~Zhang, and H.~Yang, ``Dysat: Deep neural representation learning on dynamic graphs via self-attention networks,'' in \emph{WSDM}, 2020, pp. 519--527.

\bibitem{zhu2020beyond}
J.~Zhu, Y.~Yan, L.~Zhao, M.~Heimann, L.~Akoglu, and D.~Koutra, ``Beyond homophily in graph neural networks: Current limitations and effective designs,'' in \emph{NeurIPS}, vol.~33, 2020, pp. 7793--7804.

\bibitem{Wu2021Self}
J.~Wu, X.~Wang, F.~Feng, X.~He, L.~Chen, J.~Lian, and X.~Xie, ``Self-supervised graph learning for recommendation,'' in \emph{SIGIR}, 2021.

\bibitem{meirom2021controlling}
E.~Meirom, H.~Maron, S.~Mannor, and G.~Chechik, ``Controlling graph dynamics with reinforcement learning and graph neural networks,'' in \emph{ICML}.\hskip 1em plus 0.5em minus 0.4em\relax PMLR, 2021, pp. 7565--7577.

\bibitem{chen2022think}
S.~Chen, P.-L. Guhur, M.~Tapaswi, C.~Schmid, and I.~Laptev, ``Think global, act local: Dual-scale graph transformer for vision-and-language navigation,'' in \emph{CVPR}, 2022, pp. 16\,537--16\,547.

\bibitem{keriven2022not}
N.~Keriven, ``Not too little, not too much: a theoretical analysis of graph (over) smoothing,'' in \emph{NeurIPS}, 2022.

\bibitem{peng2022sancus}
J.~Peng, Z.~Chen, Y.~Shao, Y.~Shen, L.~Chen, and J.~Cao, ``Sancus: sta le n ess-aware c omm u nication-avoiding full-graph decentralized training in large-scale graph neural networks,'' \emph{in VLDB Endowment}, vol.~15, no.~9, pp. 1937--1950, 2022.

\bibitem{ma2022learning}
J.~Ma, M.~Wan, L.~Yang, J.~Li, B.~Hecht, and J.~Teevan, ``Learning causal effects on hypergraphs,'' in \emph{KDD}, 2022, pp. 1202--1212.

\bibitem{zhang2023graph}
J.~Zhang, ``Graph-toolformer: To empower llms with graph reasoning ability via prompt augmented by chatgpt,'' \emph{arXiv}, 2023.

\bibitem{tang2024graphgpt}
J.~Tang, Y.~Yang, W.~Wei, L.~Shi, L.~Su, S.~Cheng, D.~Yin, and C.~Huang, ``Graphgpt: Graph instruction tuning for large language models,'' in \emph{Proceedings of the 47th International ACM SIGIR Conference on Research and Development in Information Retrieval}, 2024, pp. 491--500.

\bibitem{edge2024local}
D.~Edge, H.~Trinh, N.~Cheng, J.~Bradley, A.~Chao, A.~Mody, S.~Truitt, and J.~Larson, ``From local to global: A graph rag approach to query-focused summarization,'' \emph{arXiv preprint arXiv:2404.16130}, 2024.

\bibitem{andrus2022enhanced}
B.~R. Andrus, Y.~Nasiri, S.~Cui, B.~Cullen, and N.~Fulda, ``Enhanced story comprehension for large language models through dynamic document-based knowledge graphs,'' in \emph{AAAI}, vol.~36, no.~10, 2022, pp. 10\,436--10\,444.

\bibitem{besta2024graph}
M.~Besta, N.~Blach, A.~Kubicek, R.~Gerstenberger, M.~Podstawski, L.~Gianinazzi, J.~Gajda, T.~Lehmann, H.~Niewiadomski, P.~Nyczyk \emph{et~al.}, ``Graph of thoughts: Solving elaborate problems with large language models,'' in \emph{Proceedings of the AAAI Conference on Artificial Intelligence}, vol.~38, no.~16, 2024, pp. 17\,682--17\,690.

\bibitem{bacciu2020gentle}
D.~Bacciu, F.~Errica, A.~Micheli, and M.~Podda, ``A gentle introduction to deep learning for graphs,'' \emph{Neural Networks}, vol. 129, pp. 203--221, 2020.

\bibitem{Cai2018comprehensive}
H.~Cai, V.~W. Zheng, and K.~C.-C. Chang, ``A comprehensive survey of graph embedding: Problems, techniques, and applications,'' \emph{IEEE Transactions on Knowledge and Data Engineering}, vol.~30, no.~9, pp. 1616--1637, 2018.

\bibitem{xia2021graph}
F.~Xia, K.~Sun, S.~Yu, A.~Aziz, L.~Wan, S.~Pan, and H.~Liu, ``Graph learning: A survey,'' \emph{TAI}, vol.~2, no.~2, pp. 109--127, 2021.

\bibitem{Zhang2020Deep}
Z.~Zhang, P.~Cui, and W.~Zhu, ``Deep learning on graphs: A survey,'' \emph{IEEE}, pp. 1--1, 2020.

\bibitem{Shi2016survey}
C.~Shi, Y.~Li, J.~Zhang, Y.~Sun, and S.~Y. Philip, ``A survey of heterogeneous information network analysis,'' \emph{IEEE}, vol.~29, no.~1, pp. 17--37, 2016.

\bibitem{Wang2017Knowledge}
Q.~Wang, Z.~Mao, B.~Wang, and L.~Guo, ``Knowledge graph embedding: A survey of approaches and applications,'' \emph{IEEE Transactions on Knowledge and Data Engineering}, vol.~29, no.~12, pp. 2724--2743, 2017.

\bibitem{Cui2018Survey}
P.~Cui, X.~Wang, J.~Pei, and W.~Zhu, ``A survey on network embedding,'' in \emph{IEEE}, 2018.

\bibitem{kossinets2009origins}
G.~Kossinets and D.~J. Watts, ``Origins of homophily in an evolving social network,'' \emph{American sociology}, vol. 115, no.~2, pp. 405--450, 2009.

\bibitem{tang2013exploiting}
J.~Tang, H.~Gao, X.~Hu, and H.~Liu, ``Exploiting homophily effect for trust prediction,'' in \emph{WSDM}, 2013, pp. 53--62.

\bibitem{barranco2019heterophily}
O.~Barranco, C.~Lozares, and D.~Muntanyola-Saura, ``Heterophily in social groups formation: a social network analysis,'' \emph{Quality \& Quantity}, vol.~53, no.~2, pp. 599--619, 2019.

\bibitem{chanpuriya2022simplified}
S.~Chanpuriya and C.~Musco, ``Simplified graph convolution with heterophily,'' in \emph{NeurIPS}, 2022.

\bibitem{suresh2021breaking}
S.~Suresh, V.~Budde, J.~Neville, P.~Li, and J.~Ma, ``Breaking the limit of graph neural networks by improving the assortativity of graphs with local mixing patterns,'' in \emph{KDD}, 2021.

\bibitem{Page1999Pagerank}
L.~Page, S.~Brin, R.~Motwani, and T.~Winograd, ``The pagerank citation ranking: Bringing order to the web,'' Stanford InfoLab, Tech. Rep., 1999.

\bibitem{xing2004weighted}
W.~Xing and A.~Ghorbani, ``Weighted pagerank algorithm,'' in \emph{Proceedings. Second Annual Conference on Communication Networks and Services Research, 2004.}\hskip 1em plus 0.5em minus 0.4em\relax IEEE, 2004, pp. 305--314.

\bibitem{bahmani2010fast}
B.~Bahmani, A.~Chowdhury, and A.~Goel, ``Fast incremental and personalized pagerank,'' \emph{arXiv preprint arXiv:1006.2880}, 2010.

\bibitem{li2019hierarchical}
C.~Li, K.~Jia, D.~Shen, C.-J.~R. Shi, and H.~Yang, ``Hierarchical representation learning for bipartite graphs.'' in \emph{IJCAI}, vol.~19, 2019, pp. 2873--2879.

\bibitem{zhao2022learning}
Y.~Zhao, S.~Wei, H.~Du, X.~Chen, Q.~Li, F.~Zhuang, J.~Liu, and G.~Kou, ``Learning bi-typed multi-relational heterogeneous graph via dual hierarchical attention networks,'' \emph{TKDE}, no.~01, pp. 1--12, 2022.

\bibitem{zhao2022stock}
Y.~Zhao, H.~Du, Y.~Liu, S.~Wei, X.~Chen, H.~Feng, Q.~Shuai, Q.~Li, F.~Zhuang, and G.~Kou, ``Stock movement prediction based on bi-typed and hybrid-relational market knowledge graph via dual attention networks,'' \emph{TKDE}, 2022.

\bibitem{vaswani2017attention}
A.~Vaswani, N.~Shazeer, N.~Parmar, J.~Uszkoreit, L.~Jones, A.~N. Gomez, Łukasz Kaiser, and I.~Polosukhin, ``Attention is all you need,'' in \emph{NIPS}, 2017, pp. 5998--6008.

\bibitem{Wan2020Reinforcement}
G.~Wan, B.~Du, S.~Pan, and G.~Haffari, ``Reinforcement learning based meta-path discovery in large-scale heterogeneous information networks,'' in \emph{AAAI}, 2020, pp. 6094--6101.

\bibitem{Miller2013Parallel}
G.~L. Miller, R.~Peng, and S.~C. Xu, ``Parallel graph decompositions using random shifts,'' in \emph{ACM}, 2013, pp. 196--203.

\bibitem{Ellens2013Graph}
W.~Ellens and R.~EKooij, ``Graph measures and network robustness,'' \emph{arXiv}, 2013.

\bibitem{Baswana2006Approximate}
S.~Baswana and S.~Sen, ``Approximate distance oracles for unweighted graphs in expected o (n 2) time,'' \emph{TALG}, vol.~2, no.~4, pp. 557--577, 2006.

\bibitem{Wang2022Minority}
K.~Wang, J.~An, M.~Zhou, Z.~Shi, X.~Shi, and Q.~Kang, ``Minority-weighted graph neural network for imbalanced node classification in social networks of internet of people,'' \emph{IEEE Internet of Things Journal}, 2022.

\bibitem{Zhou2021Core}
W.~Zhou, H.~Huang, Q.-S. Hua, D.~Yu, H.~Jin, and X.~Fu, ``Core decomposition and maintenance in weighted graph,'' \emph{World Wide Web}, vol.~24, no.~2, pp. 541--561, 2021.

\bibitem{Toivonen2011Compression}
H.~Toivonen, F.~Zhou, A.~Hartikainen, and A.~Hinkka, ``Compression of weighted graphs,'' in \emph{KDD}, 2011, pp. 965--973.

\bibitem{Dhillon2007Weighted}
I.~S. Dhillon, Y.~Guan, and B.~Kulis, ``Weighted graph cuts without eigenvectors a multilevel approach,'' \emph{IEEE transactions on pattern analysis and machine intelligence}, vol.~29, no.~11, pp. 1944–--1957, 2007.

\bibitem{Umeyama1988Eigendecomposition}
S.~Umeyama, ``An eigendecomposition approach to weighted graph matching problems,'' \emph{TPAMI}, vol.~10, no.~5, pp. 695--703, 1988.

\bibitem{MarkNewman2004FindingAE}
M.~Newman and M.~Girvan, ``Finding and evaluating community structure in networks.'' \emph{Physical Review E}, 2004.

\bibitem{Peng2018Large}
H.~Peng, J.~Li, Y.~He, Y.~Liu, M.~Bao, L.~Wang, Y.~Song, and Q.~Yang, ``Large-scale hierarchical text classification with recursively regularized deep graph-cnn,'' in \emph{WWW}, 2018, pp. 1063--–1072.

\bibitem{Shi2019Skeleton}
L.~Shi, Y.~Zhang, J.~Cheng, and H.~Lu, ``Skeleton-based action recognition with directed graph neural networks,'' in \emph{CVPR}, 2019, pp. 7912--7921.

\bibitem{Feng2019Hypergraph}
Y.~Feng, H.~You, Z.~Zhang, R.~Ji, and Y.~Gao, ``Hypergraph neural networks,'' in \emph{AAAI}, 2019, pp. 3558--3565.

\bibitem{Berge1984Hypergraphs}
B.~Claude, \emph{Hypergraphs: combinatorics of finite sets}.\hskip 1em plus 0.5em minus 0.4em\relax Elsevier, 1984, vol.~45.

\bibitem{Sun2008Hypergraph}
L.~Sun, S.~Ji, and J.~Ye, ``Hypergraph spectral learning for multi-label classification,'' in \emph{KDD}, "2008", pp. 668--676.

\bibitem{Sun2021Heterogeneous}
X.~Sun, H.~Yin, B.~Liu, H.~Chen, J.~Cao, Y.~Shao, and N.~Q.~V. Hung, ``Heterogeneous hypergraph embedding for graph classification,'' in \emph{WSDM}, 2021, pp. 725--733.

\bibitem{St2022Influential}
G.~St-Onge, I.~Iacopini, V.~Latora, A.~Barrat, G.~Petri, A.~Allard, and L.~H{\'e}bert-Dufresne, ``Influential groups for seeding and sustaining nonlinear contagion in heterogeneous hypergraphs,'' \emph{Commun. Phys.}, vol.~5, no.~1, pp. 1--16, 2022.

\bibitem{Zhang2019Heterogeneous}
C.~Zhang, D.~Song, C.~Huang, A.~Swami, and N.~V. Chawla, ``Heterogeneous graph neural network,'' in \emph{KDD}, 2019, pp. 793--803.

\bibitem{Niepert2016Learning}
M.~Niepert, M.~Ahmed, and K.~Kutzkov, ``Learning convolutional neural networks for graphs,'' in \emph{ICML}, 2016, pp. 2014--2023.

\bibitem{ye2022learning}
J.~Ye, Z.~Liu, B.~Du, L.~Sun, W.~Li, Y.~Fu, and H.~Xiong, ``Learning the evolutionary and multi-scale graph structure for multivariate time series forecasting,'' in \emph{KDD}, 2022, pp. 2296--2306.

\bibitem{pareja2020evolvegcn}
A.~Pareja, G.~Domeniconi, J.~Chen, T.~Ma, T.~Suzumura, H.~Kanezashi, T.~Kaler, T.~Schardl, and C.~Leiserson, ``Evolvegcn: Evolving graph convolutional networks for dynamic graphs,'' in \emph{AAAI}, vol.~34, no.~04, 2020, pp. 5363--5370.

\bibitem{gehring2017convolutional}
J.~Gehring, M.~Auli, D.~Grangier, D.~Yarats, and Y.~N. Dauphin, ``Convolutional sequence to sequence learning,'' in \emph{ICML}.\hskip 1em plus 0.5em minus 0.4em\relax PMLR, 2017, pp. 1243--1252.

\bibitem{trivedi2019dyrep}
R.~Trivedi, M.~Farajtabar, P.~Biswal, and H.~Zha, ``Dyrep: Learning representations over dynamic graphs,'' in \emph{ICML}, 2019.

\bibitem{Ma2020Streaming}
Y.~Ma, Z.~Guo, Z.~Ren, J.~Tang, and D.~Yin, ``Streaming graph neural networks,'' in \emph{SIGIR}, 2020, pp. 719--728.

\bibitem{Zhang2018Dynamic}
Z.~Zhang, H.~Lin, Y.~Gao, and K.~BNRist, ``Dynamic hypergraph structure learning.'' in \emph{IJCAI}, 2018, pp. 3162--3169.

\bibitem{Jiang2019Dynamic}
J.~Jiang, Y.~Wei, Y.~Feng, J.~Cao, and Y.~Gao, ``Dynamic hypergraph neural networks.'' in \emph{IJCAI}, 2019, pp. 2635--2641.

\bibitem{zhu2020unsupervised}
X.~Zhu, S.~Zhang, Y.~Zhu, P.~Zhu, and Y.~Gao, ``Unsupervised spectral feature selection with dynamic hyper-graph learning,'' \emph{TKDE}, 2020.

\bibitem{Jeh2002Simrankff}
G.~Jeh and J.~Widom, ``Simrank: a measure of structural-context similarity,'' in \emph{KDD}, "2002", pp. 538--543.

\bibitem{Koutra2013Big}
D.~Koutra, H.~Tong, and D.~Lubensky, ``Big-align: Fast bipartite graph alignment,'' in \emph{ICDM}, 2013, pp. 389--398.

\bibitem{Zhang2017User}
D.~Zhang, J.~Yin, X.~Zhu, and C.~Zhang, ``User profile preserving social network embedding,'' in \emph{IJCAI}, 2017, pp. 3378--3384.

\bibitem{gui2021pine}
S.~Gui, X.~Zhang, P.~Zhong, S.~Qiu, M.~Wu, J.~Ye, Z.~Wang, and J.~Liu, ``Pine: Universal deep embedding for graph nodes via partial permutation invariant set functions,'' \emph{TPAMI}, vol.~44, no.~2, pp. 770--782, 2021.

\bibitem{Xu2017Scene}
D.~Xu, Y.~Zhu, C.~B. Choy, and L.~Fei-Fei, ``Scene graph generation by iterative message passing,'' in \emph{CVPR}, 2017.

\bibitem{kim2019edge}
J.~Kim, T.~Kim, S.~Kim, and C.~D. Yoo, ``Edge-labeling graph neural network for few-shot learning,'' in \emph{CVPR}, 2019, pp. 11--20.

\bibitem{jiang2020co}
X.~Jiang, R.~Zhu, S.~Li, and P.~Ji, ``Co-embedding of nodes and edges with graph neural networks,'' \emph{IEEE Transactions on Pattern Analysis and Machine Intelligence}, 2020.

\bibitem{Ying2018Hierarchical}
R.~Ying, J.~You, C.~Morris, X.~Ren, W.~L. Hamilton, and J.~Leskovec, ``Hierarchical graph representation learning with differentiable pooling,'' in \emph{NeurIPS}, 2018, pp. 4801–--4811.

\bibitem{Pan2018Adversarially}
S.~Pan, R.~Hu, G.~Long, J.~Jiang, L.~Yao, and C.~Zhang, ``Adversarially regularized graph autoencoder for graph embedding,'' in \emph{IJCAI}, 2018, pp. 2609--2615.

\bibitem{Park2020Unsupervised}
C.~Park, D.~Kim, J.~Han, and H.~Yu, ``Unsupervised attributed multiplex network embedding,'' in \emph{AAAI}, 2020, pp. 5371--5378.

\bibitem{Ribeiro2017Struc2vec}
L.~F. Ribeiro, P.~H. Saverese, and D.~R. Figueiredo, ``Struc2vec: Learning node representations from structural identity,'' in \emph{KDD}, 2017, pp. 385--–394.

\bibitem{Huang2017Label}
X.~Huang, J.~Li, and X.~Hu, ``Label informed attributed network embedding,'' in \emph{WSDM}, 2017, pp. 731--739.

\bibitem{dong2021individual}
Y.~Dong, J.~Kang, H.~Tong, and J.~Li, ``Individual fairness for graph neural networks: A ranking based approach,'' in \emph{KDD}, 2021, pp. 300--310.

\bibitem{Gilbert2004Compressing}
A.~C. Gilbert and K.~Levchenko, ``Compressing network graphs,'' in \emph{KDD}, 2004.

\bibitem{Clauset2008Hierarchical}
A.~Clauset, C.~Moore, and M.~E. Newman, ``Hierarchical structure and the prediction of missing links in networks,'' \emph{Nature}, vol. 453, no. 7191, pp. 98--101, 2008.

\bibitem{zhu2021neural}
Z.~Zhu, Z.~Zhang, L.-P. Xhonneux, and J.~Tang, ``Neural bellman-ford networks: A general graph neural network framework for link prediction,'' in \emph{NeurIPS}, vol.~34, 2021, pp. 29\,476--29\,490.

\bibitem{Xu2018Exploring}
L.~Xu, X.~Wei, J.~Cao, and P.~S. Yu, ``On exploring semantic meanings of links for embedding social networks,'' in \emph{WWW}, 2018, pp. 479--488.

\bibitem{ArisAnagnostopoulos2016CommunityDO}
A.~Anagnostopoulos, J.~Łącki, S.~Lattanzi, S.~Leonardi, and M.~Mahdian, ``Community detection on evolving graphs,'' \emph{NeurIPS}, 2016.

\bibitem{Chen2017Supervised}
Z.~Chen, J.~Bruna, and L.~Li, ``Supervised community detection with line graph neural networks,'' in \emph{ICLR}, 2019.

\bibitem{shchur2019overlapping}
O.~Shchur and S.~G{\"u}nnemann, ``Overlapping community detection with graph neural networks,'' \emph{arXiv}, 2019.

\bibitem{Pan2017Task}
S.~Pan, J.~Wu, X.~Zhu, G.~Long, and C.~Zhang, ``Task sensitive feature exploration and learning for multitask graph classification,'' \emph{TCYB}, vol.~47, no.~3, pp. 744--–758, 2017.

\bibitem{Ahmed2013Distributed}
A.~Ahmed, N.~Shervashidze, S.~Narayanamurthy, V.~Josifovski, and A.~J. Smola, ``Distributed large-scale natural graph factorization,'' in \emph{WWW}, 2013, pp. 37--48.

\bibitem{Ou2016Asymmetric}
M.~Ou, P.~C.~J. Pei, Z.~Zhang, and Wenwu, ``Asymmetric transitivity preserving graph embedding,'' in \emph{KDD}, 2016, pp. 1105--1114.

\bibitem{Mikolov2013Distributed}
T.~Mikolov, I.~Sutskever, K.~Chen, G.~S. Corrado, and J.~Dean, ``Distributed representations of words and phrases and their compositionality,'' in \emph{NeurIPS}, 2013, pp. 3111–--3119.

\bibitem{Mikolov2013Efficient}
T.~Mikolov, K.~Chen, G.~Corrado, and J.~Dean, ``Efficient estimation of word representations in vector space,'' \emph{arXiv}, 2013.

\bibitem{Perozzi2014Deepwalk}
B.~Perozzi, R.~Al-Rfou, and S.~Skiena, ``Deepwalk: Online learning of social representations,'' in \emph{KDD}, 2014, pp. 701--710.

\bibitem{Cao2016Deep}
S.~Cao, W.~Lu, and Q.~Xu, ``Deep neural networks for learning graph representations,'' in \emph{AAAI}, 2016, pp. 1145--1152.

\bibitem{Velickovic2019Deep}
P.~Velickovic, W.~Fedus, W.~L. Hamilton, P.~Li, Y.~Bengio, and R.~D. Hjelm, ``Deep graph infomax,'' in \emph{ICLR}, 2019.

\bibitem{Simonovsky2018Graphvae}
M.~Simonovsky and authorNikos Komodakis, ``Graphvae: Towards generation of small graphs using variational autoencoders,'' in \emph{ICANN}, 2018, pp. 412--–422.

\bibitem{zhao2022fisrebp}
Y.~Zhao, S.~Wei, Y.~Guo, Q.~Yang, and G.~Kou, ``Fisrebp: Enterprise bankruptcy prediction via fusing its intra-risk and spillover-risk,'' \emph{arXiv}, 2022.

\bibitem{zhao2022connecting}
Y.~Zhao, H.~Zhou, A.~Zhang, R.~Xie, Q.~Li, and F.~Zhuang, ``Connecting embeddings based on multiplex relational graph attention networks for knowledge graph entity typing,'' \emph{TKDE}, 2022.

\bibitem{peng2020motif}
H.~Peng, J.~Li, Q.~Gong, Y.~Ning, S.~Wang, and L.~He, ``Motif-matching based subgraph-level attentional convolutional network for graph classification,'' in \emph{AAAI}, vol.~34, no.~04, 2020, pp. 5387--5394.

\bibitem{Hu2019Heterogeneous}
L.~Hu, T.~Yang, C.~Shi, H.~Ji, and X.~Li, ``Heterogeneous graph attention networks for semi-supervised short text classification,'' in \emph{EMNLP}, 2019, pp. 4821--4830.

\bibitem{Wan2021Contrastive}
S.~Wan, S.~Pan, J.~Yang, and C.~Gong, ``Contrastive and generative graph convolutional networks for graph-based semi-supervised learning,'' in \emph{AAAI}, 2021.

\bibitem{jiang2019semi}
B.~Jiang, Z.~Zhang, D.~Lin, J.~Tang, and B.~Luo, ``Semi-supervised learning with graph learning-convolutional networks,'' in \emph{CVPR}, 2019, pp. 11\,313--11\,320.

\bibitem{Sun2020Multi}
K.~Sun, Z.~Lin, and Z.~Zhu, ``Multi-stage self-supervised learning for graph convolutional networks on graphs with few labeled nodes,'' in \emph{AAAI}, 2020, pp. 5892--5899.

\bibitem{tian2017deepcluster}
K.~Tian, S.~Zhou, and J.~Guan, ``Deepcluster: A general clustering framework based on deep learning,'' in \emph{Machine Learning and Knowledge Discovery in Databases: European Conference, ECML PKDD 2017, Skopje, Macedonia, September 18--22, 2017, Proceedings, Part II 17}.\hskip 1em plus 0.5em minus 0.4em\relax Springer, 2017, pp. 809--825.

\bibitem{Hu2020Gpt}
Z.~Hu, Y.~Dong, K.~Wang, K.-W. Chang, and Y.~Sun, ``Gpt-gnn: Generative pre-training of graph neural networks,'' in \emph{KDD}, 2020, pp. 1857--1867.

\bibitem{wang2021self}
X.~Wang, N.~Liu, H.~Han, and C.~Shi, ``Self-supervised heterogeneous graph neural network with co-contrastive learning,'' in \emph{KDD}, 2021.

\bibitem{Kipf2018Neural}
T.~Kipf, E.~Fetaya, K.-C. Wang, M.~Welling, and R.~Zemel, ``Neural relational inference for interacting systems,'' in \emph{ICML}, 2018, pp. 2688--2697.

\bibitem{Tang2012Unsupervised}
J.~Tang and H.~Liu, ``Unsupervised feature selection for linked social media data,'' in \emph{KDD}, 2012, pp. 904--912.

\bibitem{teru2020inductive}
K.~Teru, E.~Denis, and W.~Hamilton, ``Inductive relation prediction by subgraph reasoning,'' in \emph{ICML}.\hskip 1em plus 0.5em minus 0.4em\relax PMLR, 2020, pp. 9448--9457.

\bibitem{barabasi1999emergence}
A.-L. Barab{\'a}si and R.~Albert, ``Emergence of scaling in random networks,'' \emph{science}, vol. 286, no. 5439, pp. 509--512, 1999.

\bibitem{girvan2002community}
M.~Girvan and M.~E. Newman, ``Community structure in social and biological networks,'' \emph{in PNAS}, vol.~99, no.~12, pp. 7821--7826, 2002.

\bibitem{newman2004finding}
M.~E. Newman and M.~Girvan, ``Finding and evaluating community structure in networks,'' \emph{Physical review E}, vol.~69, no.~2, 2004.

\bibitem{fortunato2010community}
S.~Fortunato, ``Community detection in graphs,'' \emph{Physics Reports}, vol. 486, no. 3-5, pp. 75--174, 2010.

\bibitem{defferrard2016convolutional}
M.~Defferrard, X.~Bresson, and P.~Vandergheynst, ``Convolutional neural networks on graphs with fast localized spectral filtering,'' in \emph{NeurIPS}, 2016, pp. 3844–--3852.

\bibitem{Bordes2013Translating}
A.~Bordes, N.~Usunier, A.~Garcia-Duran, J.~Weston, and O.~Yakhnenko, ``Translating embeddings for modeling multi-relational data,'' in \emph{NeurIPS}, vol.~26, 2013.

\bibitem{Gilmer2017Neural}
J.~Gilmer, S.~S. Schoenholz, P.~F. Riley, O.~Vinyals, and G.~E. Dahl, ``Neural message passing for quantum chemistry,'' in \emph{ICML}, 2017, pp. 1263--1272.

\bibitem{Xu2019How}
K.~Xu, W.~Hu, J.~Leskovec, and S.~Jegelka, ``How powerful are graph neural networks?'' in \emph{ICLR}, 2019.

\bibitem{Liben2003Link}
D.~Liben-Nowell and J.~Kleinberg, ``The link prediction problem for social networks,'' in \emph{CIKM}, 2003, pp. 556--559.

\bibitem{Tang2015Line}
J.~Tang, M.~Qu, M.~Wang, M.~Zhang, J.~Yan, and Q.~Mei, ``Line: Large-scale information network embedding,'' in \emph{WWW}, 2015, pp. 1067--1077.

\bibitem{Bruna2013Spectral}
J.~Bruna, W.~Zaremba, A.~Szlam, and Y.~LeCun, ``Spectral networks and locally connected networks on graphs,'' in \emph{ICLR}, 2013.

\bibitem{newman2006finding}
M.~E. Newman, ``Finding community structure in networks using the eigenvectors of matrices,'' \emph{Physical review E}, vol.~74, no.~3, 2006.

\bibitem{Shuman2013Emerging}
D.~I. Shuman, S.~K. Narang, P.~Frossard, A.~Ortega, and P.~Vandergheynst, ``The emerging field of signal processing on graphs: Extending high-dimensional data analysis to networks and other irregular domains,'' \emph{IEEE Signal Process. Mag.}, vol.~30, no.~3, pp. 83--98, 2013.

\bibitem{wang2014knowledge}
Z.~Wang, J.~Zhang, J.~Feng, and Z.~Chen, ``Knowledge graph and text jointly embedding,'' in \emph{EMNLP}, 2014, pp. 1591--1601.

\bibitem{Duvenaud2015Convolutional}
D.~Duvenaud, D.~Maclaurin, J.~Aguilera-Iparraguirre, R.~Gómez-Bombarelli, T.~Hirzel, A.~Aspuru-Guzik, and R.~P. Adams, ``Convolutional networks on graphs for learning molecular fingerprints,'' in \emph{NeurIPS}, 2015, pp. 2224–--2232.

\bibitem{Adamic2003Friends}
L.~A. Adamic and E.~Adar, ``Friends and neighbors on the web,'' \emph{Social Networks}, vol.~25, no.~3, pp. 211--230, 2003.

\bibitem{Yan2006Graph}
S.~Yan, D.~Xu, B.~Zhang, H.-J. Zhang, Q.~Yang, and S.~Lin, ``Graph embedding and extensions: A general framework for dimensionality reduction,'' \emph{TPAMI}, vol.~29, no.~1, pp. 40--51, 2006.

\bibitem{Leskovec2007Graph}
J.~Leskovec, J.~Kleinberg, and C.~Faloutsos, ``Graph evolution: Densification and shrinking diameters,'' \emph{JACM}, vol.~1, no.~1, pp. 2--es, 2007.

\bibitem{Leskovec2005Graphs}
------, ``Graphs over time: densification laws, shrinking diameters and possible explanations,'' in \emph{KDD}, 2005, pp. 177--187.

\bibitem{Leskovec2007Dynamics}
J.~Leskovec, L.~A. Adamic, and B.~A. Huberman, ``The dynamics of viral marketing,'' \emph{TWEB}, vol.~1, no.~1, pp. 5--es, 2007.

\bibitem{White1976Social}
H.~C. White, S.~A. Boorman, and R.~L. Breiger, ``Social structure from multiple networks. i. blockmodels of roles and positions,'' \emph{Sociology}, vol.~81, no.~4, pp. 730--780, 1976.

\bibitem{Newman2005Measure}
M.~E. Newman, ``A measure of betweenness centrality based on random walks,'' \emph{Social networks}, vol.~27, no.~1, pp. 39--54, 2005.

\bibitem{Newman2004Detecting}
Newman, ``Detecting community structure in networks,'' \emph{The European physical journal B}, vol.~38, pp. 321--330, 2004.

\bibitem{yin2022survey}
D.~Yin, L.~Dong, H.~Cheng, X.~Liu, K.-W. Chang, F.~Wei, and J.~Gao, ``A survey of knowledge-intensive nlp with pre-trained language models,'' \emph{arXiv}, 2022.

\bibitem{zhang2022greaselm}
X.~Zhang, A.~Bosselut, M.~Yasunaga, H.~Ren, P.~Liang, C.~D. Manning, and J.~Leskovec, ``Greaselm: Graph reasoning enhanced language models for question answering,'' \emph{arXiv}, 2022.

\bibitem{shi2023chatgraph}
Y.~Shi, H.~Ma, W.~Zhong, G.~Mai, X.~Li, T.~Liu, and J.~Huang, ``Chatgraph: Interpretable text classification by converting chatgpt knowledge to graphs,'' \emph{arXiv}, 2023.

\bibitem{yasunaga2022linkbert}
M.~Yasunaga, J.~Leskovec, and P.~Liang, ``Linkbert: Pretraining language models with document links,'' \emph{arXiv preprint arXiv:2203.15827}, 2022.

\bibitem{yasunaga2022deep}
M.~Yasunaga, A.~Bosselut, H.~Ren, X.~Zhang, C.~D. Manning, P.~S. Liang, and J.~Leskovec, ``Deep bidirectional language-knowledge graph pretraining,'' \emph{Advances in Neural Information Processing Systems}, vol.~35, pp. 37\,309--37\,323, 2022.

\bibitem{wang2023dynamic}
Y.~Wang, H.~Zhang, J.~Liang, and R.~Li, ``Dynamic heterogeneous-graph reasoning with language models and knowledge representation learning for commonsense question answering,'' in \emph{Proceedings of the 61st Annual Meeting of the Association for Computational Linguistics (Volume 1: Long Papers)}, 2023, pp. 14\,048--14\,063.

\bibitem{dong2022incorporating}
Q.~Dong, Y.~Liu, S.~Cheng, S.~Wang, Z.~Cheng, S.~Niu, and D.~Yin, ``Incorporating explicit knowledge in pre-trained language models for passage re-ranking,'' \emph{arXiv}, 2022.

\bibitem{saha2022explanation}
S.~Saha, P.~Yadav, and M.~Bansal, ``Explanation graph generation via pre-trained language models: An empirical study with contrastive learning,'' \emph{arXiv}, 2022.

\bibitem{yu2022jaket}
D.~Yu, C.~Zhu, Y.~Yang, and M.~Zeng, ``Jaket: Joint pre-training of knowledge graph and language understanding,'' in \emph{AAAI}, vol.~36, no.~10, 2022, pp. 11\,630--11\,638.

\bibitem{li2022enhancing}
Y.~Li, J.~Cao, X.~Cong, Z.~Zhang, B.~Yu, H.~Zhu, and T.~Liu, ``Enhancing chinese pre-trained language model via heterogeneous linguistics graph,'' in \emph{Proceedings of the 60th Annual Meeting of the Association for Computational Linguistics (Volume 1: Long Papers)}, 2022, pp. 1986--1996.

\bibitem{hu2023survey}
L.~Hu, Z.~Liu, Z.~Zhao, L.~Hou, L.~Nie, and J.~Li, ``A survey of knowledge enhanced pre-trained language models,'' \emph{IEEE Transactions on Knowledge and Data Engineering}, 2023.

\bibitem{zhang2022opt}
S.~Zhang, S.~Roller, N.~Goyal, M.~Artetxe, M.~Chen, S.~Chen, C.~Dewan, M.~Diab, X.~Li, X.~V. Lin \emph{et~al.}, ``Opt: Open pre-trained transformer language models,'' \emph{arXiv}, 2022.

\bibitem{anil2023palm}
R.~Anil, A.~M. Dai, O.~Firat, M.~Johnson, D.~Lepikhin, A.~Passos, S.~Shakeri, E.~Taropa, P.~Bailey, Z.~Chen \emph{et~al.}, ``Palm 2 technical report,'' \emph{arXiv}, 2023.

\bibitem{openai2023gpt4}
OpenAI, ``Gpt-4 technical report,'' \emph{arXiv}, 2023.

\bibitem{zhang2023autoalign}
R.~Zhang, Y.~Su, B.~D. Trisedya, X.~Zhao, M.~Yang, H.~Cheng, and J.~Qi, ``Autoalign: Fully automatic and effective knowledge graph alignment enabled by large language models,'' \emph{IEEE Transactions on Knowledge and Data Engineering}, 2023.

\bibitem{he2023explanations}
X.~He, X.~Bresson, T.~Laurent, and B.~Hooi, ``Explanations as features: Llm-based features for text-attributed graphs,'' \emph{arXiv preprint arXiv:2305.19523}, 2023.

\bibitem{chen2023lmexplainer}
Z.~Chen, A.~K. Singh, and M.~Sra, ``Lmexplainer: a knowledge-enhanced explainer for language models,'' \emph{arXiv}, 2023.

\bibitem{xi2023towards}
Y.~Xi, W.~Liu, J.~Lin, J.~Zhu, B.~Chen, R.~Tang, W.~Zhang, R.~Zhang, and Y.~Yu, ``Towards open-world recommendation with knowledge augmentation from large language models,'' \emph{arXiv preprint arXiv:2306.10933}, 2023.

\bibitem{jiang2023graphcare}
P.~Jiang, C.~Xiao, A.~Cross, and J.~Sun, ``Graphcare: Enhancing healthcare predictions with open-world personalized knowledge graphs,'' \emph{arXiv}, 2023.

\bibitem{qin2023disentangled}
Y.~Qin, X.~Wang, Z.~Zhang, and W.~Zhu, ``Disentangled representation learning with large language models for text-attributed graphs,'' \emph{arXiv preprint arXiv:2310.18152}, 2023.

\bibitem{feng2023knowledge}
C.~Feng, X.~Zhang, and Z.~Fei, ``Knowledge solver: Teaching llms to search for domain knowledge from knowledge graphs,'' \emph{arXiv preprint arXiv:2309.03118}, 2023.

\bibitem{zhang2023making}
Y.~Zhang, Z.~Chen, W.~Zhang, and H.~Chen, ``Making large language models perform better in knowledge graph completion,'' \emph{arXiv preprint arXiv:2310.06671}, 2023.

\bibitem{jiang2023structgpt}
J.~Jiang, K.~Zhou, Z.~Dong, K.~Ye, W.~X. Zhao, and J.-R. Wen, ``Structgpt: A general framework for large language model to reason over structured data,'' \emph{arXiv preprint arXiv:2305.09645}, 2023.

\bibitem{choudhary2023complex}
N.~Choudhary and C.~K. Reddy, ``Complex logical reasoning over knowledge graphs using large language models,'' \emph{arXiv}, 2023.

\bibitem{luo2023reasoning}
L.~Luo, Y.-F. Li, G.~Haffari, and S.~Pan, ``Reasoning on graphs: Faithful and interpretable large language model reasoning,'' \emph{arXiv preprint arXiv:2310.01061}, 2023.

\bibitem{wen2023mindmap}
Y.~Wen, Z.~Wang, and J.~Sun, ``Mindmap: Knowledge graph prompting sparks graph of thoughts in large language models,'' \emph{arXiv preprint arXiv:2308.09729}, 2023.

\bibitem{hu2024grag}
Y.~Hu, Z.~Lei, Z.~Zhang, B.~Pan, C.~Ling, and L.~Zhao, ``Grag: Graph retrieval-augmented generation,'' \emph{arXiv preprint arXiv:2405.16506}, 2024.

\bibitem{xu2024retrieval}
Z.~Xu, M.~J. Cruz, M.~Guevara, T.~Wang, M.~Deshpande, X.~Wang, and Z.~Li, ``Retrieval-augmented generation with knowledge graphs for customer service question answering,'' in \emph{Proceedings of the 47th International ACM SIGIR Conference on Research and Development in Information Retrieval}, 2024, pp. 2905--2909.

\bibitem{he2024g}
X.~He, Y.~Tian, Y.~Sun, N.~V. Chawla, T.~Laurent, Y.~LeCun, X.~Bresson, and B.~Hooi, ``G-retriever: Retrieval-augmented generation for textual graph understanding and question answering,'' \emph{arXiv preprint arXiv:2402.07630}, 2024.

\bibitem{traag2019louvain}
V.~A. Traag, L.~Waltman, and N.~J. Van~Eck, ``From louvain to leiden: guaranteeing well-connected communities,'' \emph{Scientific reports}, vol.~9, no.~1, pp. 1--12, 2019.

\bibitem{wang2023can}
H.~Wang, S.~Feng, T.~He, Z.~Tan, X.~Han, and Y.~Tsvetkov, ``Can language models solve graph problems in natural language?'' \emph{arXiv}, 2023.

\bibitem{zhu2023llms}
Y.~Zhu, X.~Wang, J.~Chen, S.~Qiao, Y.~Ou, Y.~Yao, S.~Deng, H.~Chen, and N.~Zhang, ``Llms for knowledge graph construction and reasoning: Recent capabilities and future opportunities,'' \emph{arXiv}, 2023.

\bibitem{guo2023gpt4graph}
J.~Guo, L.~Du, and H.~Liu, ``Gpt4graph: Can large language models understand graph structured data? an empirical evaluation and benchmarking,'' \emph{arXiv preprint arXiv:2305.15066}, 2023.

\bibitem{Fan2019Graph}
W.~Fan, Y.~Ma, Q.~Li, Y.~He, E.~Zhao, J.~Tang, and D.~Yin, ``Graph neural networks for social recommendation,'' in \emph{WWW}, 2019, pp. 417--426.

\bibitem{Dong2012Link}
Y.~Dong, J.~Tang, S.~Wu, J.~Tian, N.~V. Chawla, J.~Rao, and H.~Cao, ``Link prediction and recommendation across heterogeneous social networks,'' in \emph{ICDM}, 2012, pp. 181--190.

\bibitem{Shi2015Semantic}
C.~Shi, Z.~Zhang, P.~Luo, P.~S. Yu, Y.~Yue, and B.~Wu, ``Semantic path based personalized recommendation on weighted heterogeneous information networks,'' in \emph{CIKM}, 2015, pp. 453--462.

\bibitem{Shi2019Heterogeneous}
C.~Shi, B.~Hu, W.~X. Zhao, and P.~S. Yu, ``Heterogeneous information network embedding for recommendation,'' \emph{TNNLS}, vol.~31, no.~2, pp. 357--370, 2019.

\bibitem{yu2021self}
J.~Yu, H.~Yin, J.~Li, Q.~Wang, N.~Q.~V. Hung, and X.~Zhang, ``Self-supervised multi-channel hypergraph convolutional network for social recommendation,'' in \emph{WWW}, 2021, pp. 413--424.

\bibitem{yang2022multi}
Y.~Yang, C.~Huang, L.~Xia, Y.~Liang, Y.~Yu, and C.~Li, ``Multi-behavior hypergraph-enhanced transformer for sequential recommendation,'' in \emph{KDD}, 2022, pp. 2263--2274.

\bibitem{xia2022self}
L.~Xia, C.~Huang, and C.~Zhang, ``Self-supervised hypergraph transformer for recommender systems,'' in \emph{KDD}, 2022, pp. 2100--2109.

\bibitem{Song2019session}
W.~Song, Z.~Xiao, Y.~Wang, L.~Charlin, M.~Zhang, and J.~Tang, ``Session-based social recommendation via dynamic graph attention networks,'' in \emph{ICDM}, 2019, pp. 555--563.

\bibitem{zhang2022dynamic}
M.~Zhang, S.~Wu, X.~Yu, Q.~Liu, and L.~Wang, ``Dynamic graph neural networks for sequential recommendation,'' \emph{TKDE}, 2022.

\bibitem{lin2021bertgcn}
Y.~Lin, Y.~Meng, X.~Sun, Q.~Han, K.~Kuang, J.~Li, and F.~Wu, ``Bertgcn: Transductive text classification by combining gcn and bert,'' \emph{arXiv}, 2021.

\bibitem{zhao2022multi}
Y.~Zhao, L.~Wang, C.~Wang, H.~Du, S.~Wei, H.~Feng, Z.~Yu, and Q.~Li, ``Multi-granularity heterogeneous graph attention networks for extractive document summarization,'' \emph{Neural Networks}, vol. 155, pp. 340--347, 2022.

\bibitem{Rocktaschel2015Injecting}
T.~Rockt{\"a}schel, S.~Singh, and S.~Riedel, ``Injecting logical background knowledge into embeddings for relation extraction,'' in \emph{NAACL}, 2015, pp. 1119--1129.

\bibitem{Song2018N-ary}
L.~Song, Y.~Zhang, Z.~Wang, and D.~Gildea, ``N-ary relation extraction using graph state lstm,'' in \emph{EMNLP}, 2018.

\bibitem{Zhang2018Graph2}
Y.~Zhang, P.~Qi, and C.~D. Manning, ``Graph convolution over pruned dependency trees improves relation extraction,'' in \emph{EMNLP}, 2018, pp. 2205--2215.

\bibitem{Teney2017Graph}
D.~Teney, L.~Liu, and A.~van~den Hengel, ``Graph-structured representations for visual question answering,'' in \emph{CVPR}, 2017, pp. 3233--–3241.

\bibitem{Narasimhan2018Out}
M.~Narasimhan, S.~Lazebnik, and A.~G. Schwing, ``Out of the box: Reasoning with graph convolution nets for factual visual question answering,'' in \emph{NeurIPS}, 2018, pp. 2655--2666.

\bibitem{kosasih2021machine}
E.~E. Kosasih and A.~Brintrup, ``A machine learning approach for predicting hidden links in supply chain with graph neural networks,'' \emph{IJPR}, pp. 1--14, 2021.

\bibitem{cheng2021modeling}
R.~Cheng and Q.~Li, ``Modeling the momentum spillover effect for stock prediction via attribute-driven graph attention networks,'' in \emph{AAAI}, vol.~35, no.~1, 2021, pp. 55--62.

\bibitem{cheng2020contagious}
D.~Cheng, Z.~Niu, and Y.~Zhang, ``Contagious chain risk rating for networked-guarantee loans,'' in \emph{KDD}, 2020, pp. 2715--2723.

\bibitem{li2021modeling}
W.~Li, R.~Bao, K.~Harimoto, D.~Chen, J.~Xu, and Q.~Su, ``Modeling the stock relation with graph network for overnight stock movement prediction,'' in \emph{IJCAI}, 2021, pp. 4541--4547.

\bibitem{huang2022asset}
J.~Huang, R.~Xing, and Q.~Li, ``Asset pricing via deep graph learning to incorporate heterogeneous predictors,'' \emph{IJIS}, vol.~37, no.~11, pp. 8462--8489, 2022.

\bibitem{Cheng2022Subsequence}
R.~Cheng and Q.~Li, ``Subsequence-based graph routing network for capturing multiple risk propagation processes,'' in \emph{IJCAI}, 2022, pp. 3810--3816.

\bibitem{zheng2021heterogeneous}
Y.~Zheng, V.~Lee, Z.~Wu, and S.~Pan, ``Heterogeneous graph attention network for small and medium-sized enterprises bankruptcy prediction,'' in \emph{PAKDD}.\hskip 1em plus 0.5em minus 0.4em\relax Springer, 2021, pp. 140--151.

\bibitem{cheng2020graph}
D.~Cheng, X.~Wang, Y.~Zhang, and L.~Zhang, ``Graph neural network for fraud detection via spatial-temporal attention,'' \emph{TKDE}, 2020.

\bibitem{liu2021pick}
Y.~Liu, X.~Ao, Z.~Qin, J.~Chi, J.~Feng, H.~Yang, and Q.~He, ``Pick and choose: a gnn-based imbalanced learning approach for fraud detection,'' in \emph{WWW}, 2021, pp. 3168--3177.

\bibitem{chen2022antibenford}
T.~Chen and C.~Tsourakakis, ``Antibenford subgraphs: Unsupervised anomaly detection in financial networks,'' in \emph{KDD}, 2022.

\bibitem{zhang2021weakly}
L.~Zhang, J.~Ding, Y.~Xu, Y.~Liu, and S.~Zhou, ``Weakly-supervised text classification based on keyword graph,'' \emph{arXiv}, 2021.

\bibitem{Jain2016Structural}
A.~Jain, A.~R. Zamir, S.~Savarese, and A.~Saxena, ``Structural-rnn: Deep learning on spatio-temporal graphs,'' in \emph{CVPR}, 2016, pp. 5308--5317.

\bibitem{hu2021naturalistic}
Y.-C.-T. Hu, B.-H. Kung, D.~S. Tan, J.-C. Chen, K.-L. Hua, and W.-H. Cheng, ``Naturalistic physical adversarial patch for object detectors,'' in \emph{ICCV}, 2021, pp. 7848--7857.

\bibitem{chen2018searching}
L.-C. Chen, M.~Collins, Y.~Zhu, G.~Papandreou, B.~Zoph, F.~Schroff, H.~Adam, and J.~Shlens, ``Searching for efficient multi-scale architectures for dense image prediction,'' in \emph{NeurIPS}, vol.~31, 2018.

\bibitem{huang2022dgraph}
X.~Huang, Y.~Yang, Y.~Wang, C.~Wang, Z.~Zhang, J.~Xu, L.~Chen, and M.~Vazirgiannis, ``Dgraph: A large-scale financial dataset for graph anomaly detection,'' in \emph{NeurIPS}, 2022.

\bibitem{rao2021xfraud}
S.~X. Rao, S.~Zhang, Z.~Han, Z.~Zhang, W.~Min, Z.~Chen, Y.~Shan, Y.~Zhao, and C.~Zhang, ``xfraud: explainable fraud transaction detection,'' \emph{VLDB Endowment}, no.~3, pp. 427--436, 2021.

\bibitem{bommasani2021opportunities}
R.~Bommasani, D.~A. Hudson, E.~Adeli, R.~Altman, S.~Arora, S.~von Arx, M.~S. Bernstein, J.~Bohg, A.~Bosselut, E.~Brunskill \emph{et~al.}, ``On the opportunities and risks of foundation models,'' \emph{arXiv preprint arXiv:2108.07258}, 2021.

\bibitem{kirillov2023segment}
A.~Kirillov, E.~Mintun, N.~Ravi, H.~Mao, C.~Rolland, L.~Gustafson, T.~Xiao, S.~Whitehead, A.~C. Berg, W.-Y. Lo \emph{et~al.}, ``Segment anything,'' in \emph{Proceedings of the IEEE/CVF International Conference on Computer Vision}, 2023, pp. 4015--4026.

\bibitem{hong2024spectralgpt}
D.~Hong, B.~Zhang, X.~Li, Y.~Li, C.~Li, J.~Yao, N.~Yokoya, H.~Li, P.~Ghamisi, X.~Jia \emph{et~al.}, ``Spectralgpt: Spectral remote sensing foundation model,'' \emph{IEEE Transactions on Pattern Analysis and Machine Intelligence}, 2024.

\bibitem{xia2024opengraph}
L.~Xia, B.~Kao, and C.~Huang, ``Opengraph: Towards open graph foundation models,'' \emph{arXiv preprint arXiv:2403.01121}, 2024.

\bibitem{ye2023natural}
R.~Ye, C.~Zhang, R.~Wang, S.~Xu, Y.~Zhang \emph{et~al.}, ``Natural language is all a graph needs,'' \emph{arXiv preprint arXiv:2308.07134}, vol.~4, no.~5, p.~7, 2023.

\bibitem{yeh2022embedding}
C.-C.~M. Yeh, M.~Gu, Y.~Zheng, H.~Chen, J.~Ebrahimi, Z.~Zhuang, J.~Wang, L.~Wang, and W.~Zhang, ``Embedding compression with hashing for efficient representation learning in large-scale graph,'' in \emph{KDD}, 2022, pp. 4391--4401.

\bibitem{liu2021exact}
Z.~Liu, K.~Zhou, F.~Yang, L.~Li, R.~Chen, and X.~Hu, ``Exact: Scalable graph neural networks training via extreme activation compression,'' in \emph{ICLR}, 2021.

\bibitem{cui2022allie}
L.~Cui, X.~Tang, S.~Katariya, N.~Rao, P.~Agrawal, K.~Subbian, and D.~Lee, ``Allie: Active learning on large-scale imbalanced graphs,'' in \emph{WWW}, 2022, pp. 690--698.

\bibitem{ying2019gnnexplainer}
Z.~Ying, D.~Bourgeois, J.~You, M.~Zitnik, and J.~Leskovec, ``Gnnexplainer: Generating explanations for graph neural networks,'' in \emph{NeurIPS}, vol.~32, 2019.

\bibitem{agarwal2021neural}
R.~Agarwal, L.~Melnick, N.~Frosst, X.~Zhang, B.~Lengerich, R.~Caruana, and G.~E. Hinton, ``Neural additive models: Interpretable machine learning with neural nets,'' in \emph{NeurIPS}, vol.~34, 2021, pp. 4699--4711.

\bibitem{amara2022graphframex}
K.~Amara, R.~Ying, Z.~Zhang, Z.~Han, Y.~Shan, U.~Brandes, S.~Schemm, and C.~Zhang, ``Graphframex: Towards systematic evaluation of explainability methods for graph neural networks,'' \emph{arXiv}, 2022.

\bibitem{dai2021towards}
E.~Dai and S.~Wang, ``Towards self-explainable graph neural network,'' in \emph{CIKM}, 2021, pp. 302--311.

\bibitem{wu2022discovering}
Y.-X. Wu, X.~Wang, A.~Zhang, X.~He, and T.-S. Chua, ``Discovering invariant rationales for graph neural networks,'' \emph{arXiv}, 2022.

\bibitem{luo2020parameterized}
D.~Luo, W.~Cheng, D.~Xu, W.~Yu, B.~Zong, H.~Chen, and X.~Zhang, ``Parameterized explainer for graph neural network,'' in \emph{NeurIPS}, 2020.

\bibitem{yuan2020xgnn}
H.~Yuan, J.~Tang, X.~Hu, and S.~Ji, ``Xgnn: Towards model-level explanations of graph neural networks,'' in \emph{KDD}, 2020, pp. 430--438.

\bibitem{Klicpera2018Predict}
J.~Klicpera, A.~Bojchevski, and S.~G{\"u}nnemann, ``Predict then propagate: Graph neural networks meet personalized pagerank,'' in \emph{ICLR}, 2019.

\bibitem{chen2020simple2}
M.~Chen, Z.~Wei, Z.~Huang, B.~Ding, and Y.~Li, ``Simple and deep graph convolutional networks,'' in \emph{ICML}, 2020, pp. 1725--1735.

\bibitem{do2021graph}
T.~H. Do, D.~M. Nguyen, G.~Bekoulis, A.~Munteanu, and N.~Deligiannis, ``Graph convolutional neural networks with node transition probability-based message passing and dropnode regularization,'' \emph{ESWA}, 2021.

\bibitem{zeng2021decoupling}
H.~Zeng, M.~Zhang, Y.~Xia, A.~Srivastava, A.~Malevich, R.~Kannan, V.~Prasanna, L.~Jin, and R.~Chen, ``Decoupling the depth and scope of graph neural networks,'' in \emph{NeurIPS}, vol.~34, 2021, pp. 19\,665--19\,679.

\bibitem{Dong2022Edits}
Y.~Dong, N.~Liu, B.~Jalaian, and J.~Li, ``Edits: Modeling and mitigating data bias for graph neural networks,'' in \emph{WWW}, 2022, pp. 1259--1269.

\bibitem{Song2022Guide}
W.~Song, Y.~Dong, N.~Liu, and J.~Li, ``Guide: Group equality informed individual fairness in graph neural networks,'' in \emph{KDD}, 2022.

\bibitem{Zhang2022Fairness}
T.~Zhang, T.~Zhu, M.~Han, F.~Chen, J.~Li, W.~Zhou, and P.~S. Yu, ``Fairness in graph-based semi-supervised learning,'' \emph{KAIS}, 2022.

\bibitem{Hsieh2021Netfense}
I.-C. Hsieh and C.-T. Li, ``Netfense: Adversarial defenses against privacy attacks on neural networks for graph data,'' \emph{TKDE}, 2021.

\bibitem{Wu2022Federated}
C.~Wu, F.~Wu, L.~Lyu, T.~Qi, Y.~Huang, and X.~Xie, ``A federated graph neural network framework for privacy-preserving personalization,'' \emph{Nat. Commun.}, vol.~13, no.~1, pp. 1--10, 2022.

\end{thebibliography}

\begin{IEEEbiography}[{\includegraphics[width=0.8in,height=1in,clip,keepaspectratio]{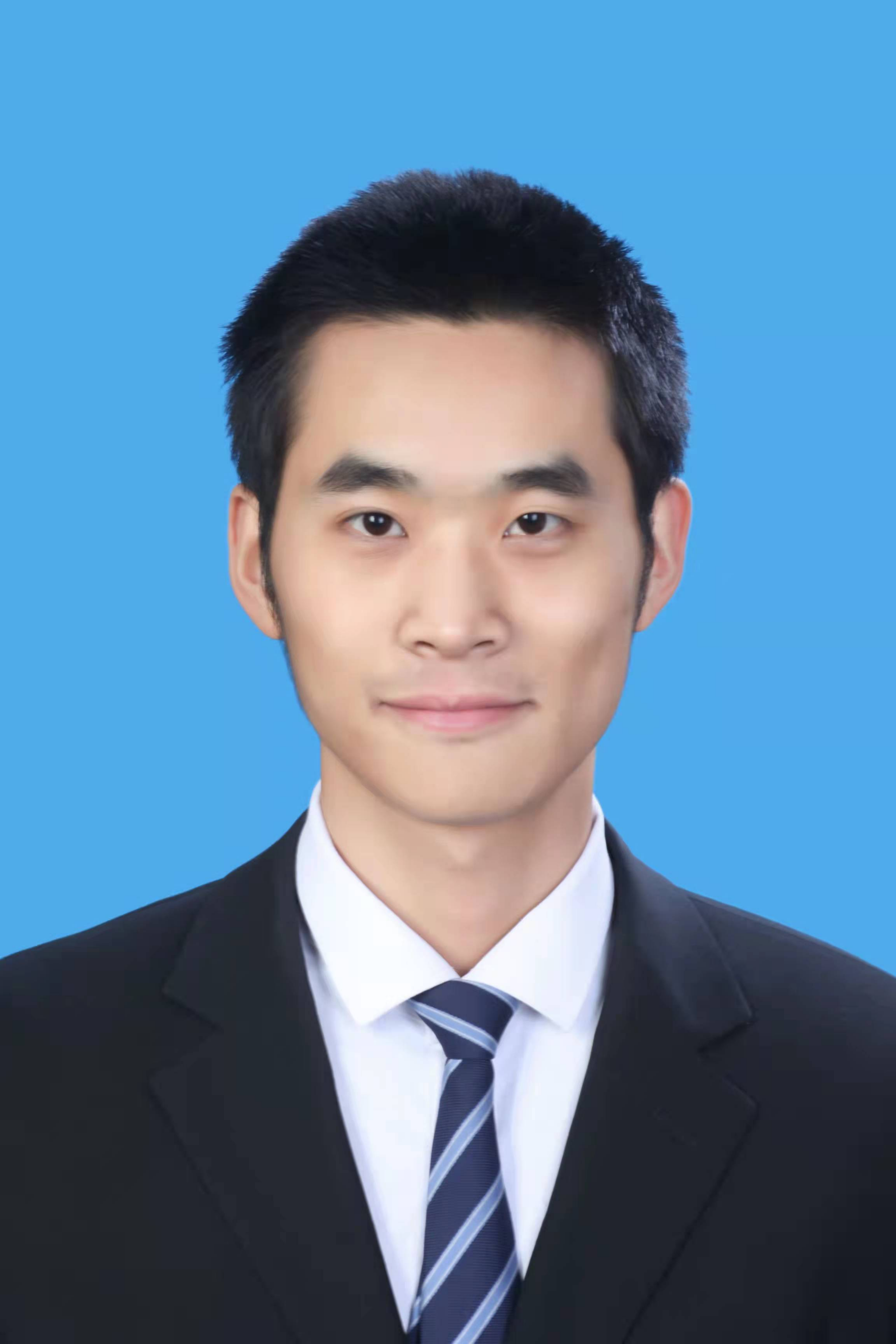}}]{Shaopeng Wei} received the B.S. degree from from Huazhong Agricultural University in 2019, and the Ph.D. degree from Southwestern University of Finance and Economics in 2024. He was a visiting scholar at ETH Zurich. He is currently an Assistant Professor with Guangxi University. His research interests include graph learning and its applications in Fintech. He has published papers in top journals such as IEEE TKDE.
\end{IEEEbiography}

\begin{IEEEbiography}[{\includegraphics[width=0.8in,height=1in,clip,keepaspectratio]{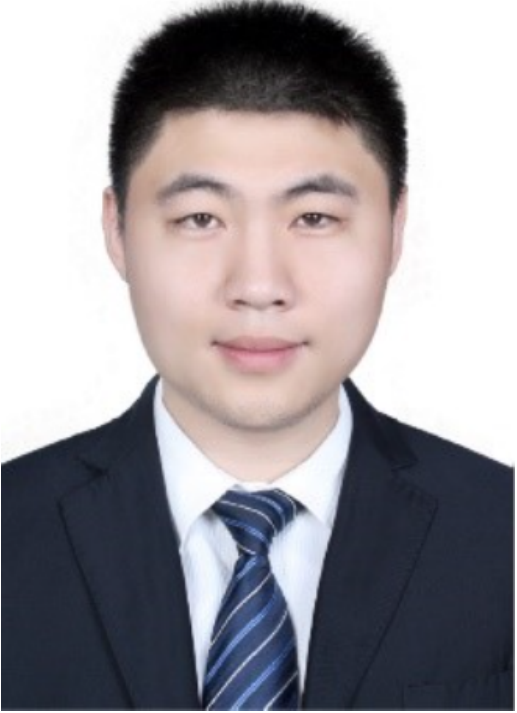}}]{Jun Wang} is an associate professor with the
Southwestern University of Finance and Economics, China. Prior to taking that post, he was a researcher with the Memorial University of Newfoundland at St. John’s, Canada. He was awarded
the National Scholarship in 2017. His research
interests include social media, social network analysis, financial analysis, and business intelligence.
\end{IEEEbiography}

\begin{IEEEbiography}[{\includegraphics[width=0.8in,height=1in,clip,keepaspectratio]{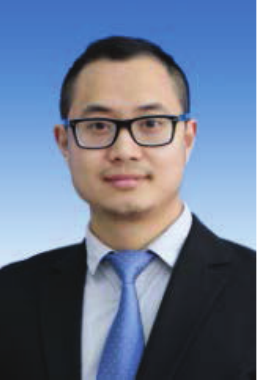}}]{Yu Zhao}
received the B.S. degree from Southwest Jiaotong University in 2006, and the M.S. and Ph.D. degrees from the Beijing University of Posts and Telecommunications in 2011 and 2017, respectively. He is a Professor at the Southwestern University of Finance and Economics. He has authored more than 30 papers including IEEE TKDE, IEEE TNNLS, IEEE TMC, IEEE TMM, KDD, ACL, ICME, etc.
% His current research interests include natural language processing, knowledge graph, machine learning, and recommendation system.
\end{IEEEbiography}

\begin{IEEEbiography}[{\includegraphics[width=0.8in,height=1in,clip,keepaspectratio]{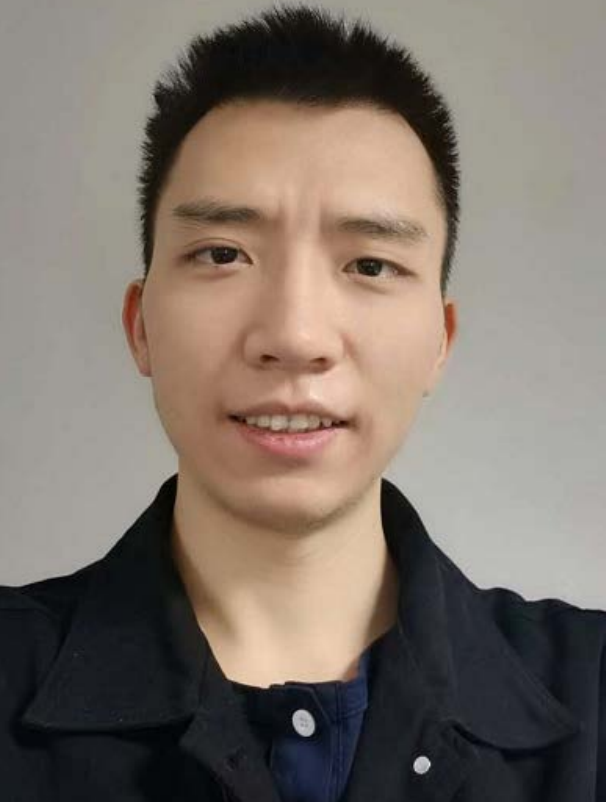}}]{Xingyan Chen}
	received the Ph. D degree in computer technology from Beijing University of Posts and Telecommunications (BUPT), in 2021. 
	% He is currently a lecturer with Southwestern University of Finance and Economics.
	He has published papers in the \textsc{IEEE TMC}, \textsc{IEEE TCSVT}, \textsc{IEEE TII}, and \textsc{IEEE INFOCOM} etc. 
	His research interests include Multimedia Communications, Multi-agent Reinforcement Learning and Stochastic Optimization.
\end{IEEEbiography}

\begin{IEEEbiography}[{\includegraphics[width=0.8in,height=1in,clip,keepaspectratio]{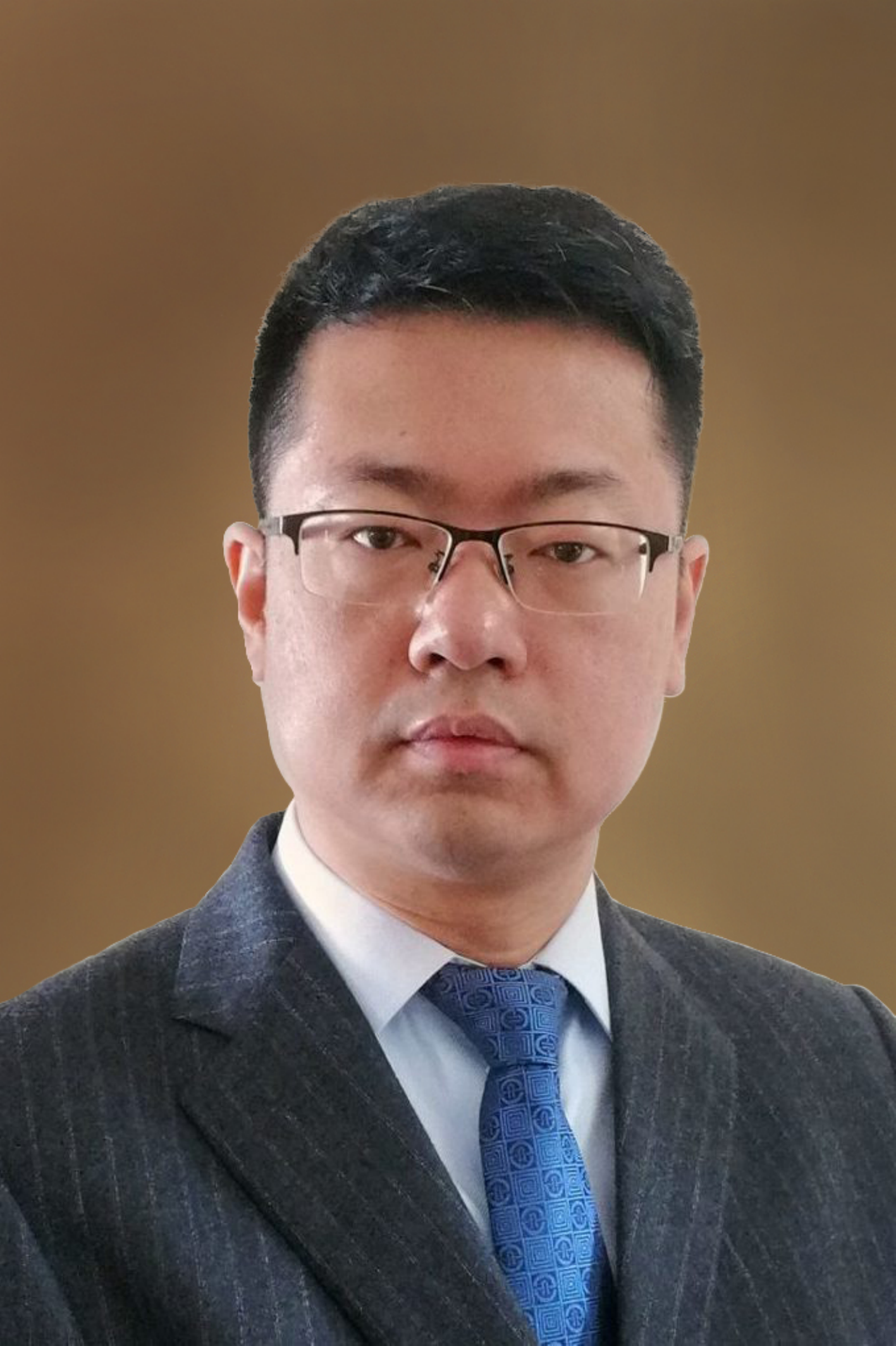}}]{Qing Li} received his Ph.D. degree from Kumoh National Institute of Technology in February of 2005, Korea, and his M.S. and B.S. degrees from Harbin Engineering University, China. 
He is a postdoctoral researcher at Arizona State University. 
He is a professor at Southwestern University of Finance and Economics, China. He has published more than 70 papers including IEEE TKDE, ACM TOIS, AAAI, WWW, etc.
% His research interests include natural language processing, FinTech. 
\end{IEEEbiography}

% \begin{IEEEbiography}[{\includegraphics[width=1in,height=1.25in,clip,keepaspectratio]{hanzhou.png}}]{Han Zhou}
% received the B.S. degree from Southwestern University of Finance and Economics in 2018, the M.S. degree in computer science from the University of Hong Kong. She is a data mining engineer in BAIDU. Her research interests include data mining and natural language processing.
% \end{IEEEbiography}

% \begin{IEEEbiography}[{\includegraphics[width=1in,height=1.25in,clip,keepaspectratio]{anxiangzhang.png}}]{Anxiang Zhang}
% received B.S. degree from the Southwestern University of Finance and Economics in 2019. 
% He is a M.S. student at Carnegie Mellon University. His research interests include machine learning, data mining.
% \end{IEEEbiography}

% \begin{IEEEbiography}[{\includegraphics[width=1in,height=1.25in,clip,keepaspectratio]{ruobingxie.jpg}}]{Ruobing Xie} received M.S. degree from Tsinghua University in 2017. He is an senior researcher in the WeChat Search Application Department, Tencent. He has published more than 30 papers in prestigious refereed journals and conferences. His research interests include knowledge graph and reprentation learning.
% \end{IEEEbiography}

\begin{IEEEbiography}[{\includegraphics[width=0.8in,height=1in,clip,keepaspectratio]{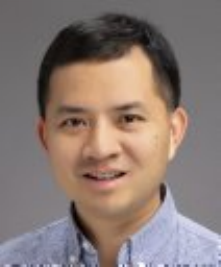}}]{Fuzhen Zhuang}
received the Ph.D. degree in computer science from the Institute of Computing Technology, Chinese Academy of Sciences. He is a professor at Beihang University. His research interests include transfer learning, machine learning, data mining. He has published more than 100 papers including Nature Communications, KDD, WWW, AAAI, IEEE TKDE, IEEE T-CYB, ACM TIST, etc.
\end{IEEEbiography}

\begin{IEEEbiography}[{\includegraphics[width=0.8in,height=1in,clip,keepaspectratio]{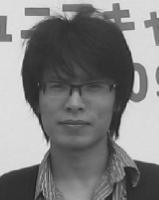}}]{Ji Liu}
received 
the B.S. degree from the University of Science and Technology of China, Hefei, China, in 2005, the master’s degree from Arizona State University, Tempe, AZ, USA, in 2010, and 
the Ph.D. degree from the University of Wisconsin–Madison, Madison, WI, USA, in 2014. 
% He was an Assistant Professor of computer science, electrical and computer engineering with the Goergen Institute for Data Science, University of Rochester (UR), Rochester, NY, USA, where he created the Machine Learning and Optimization Group.
He has authored more than 70 papers in top journals and conferences,
including JMLR, TPAMI, TNNLS, TKDD, NIPS, ICML, SIGKDD, ICCV, and CVPR.
Dr. Liu was a recipient of the Award of Best Paper Honorable Mention at SIGKDD 2010, the Award of Best Student Paper Award at UAI 2015, and the IBM Faculty Award. He is named one MIT technology review’s “35 innovators under 35 in China.”
\end{IEEEbiography}

\begin{IEEEbiography}[{\includegraphics[width=0.8in,height=1.in,clip,keepaspectratio]{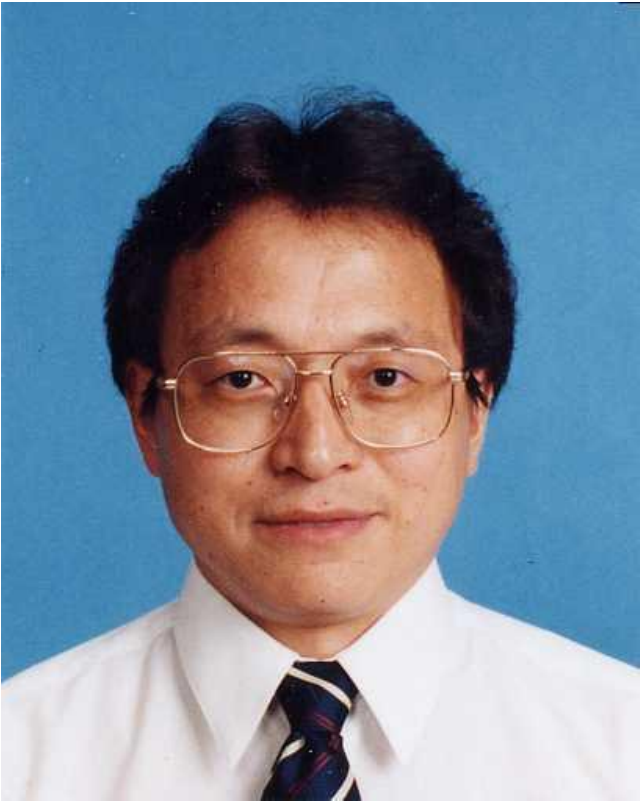}}]{Fuji Ren} received his Ph. D. degree in 1991 from the Faculty of Engineering, Hokkaido University, Japan. 
From 1991 to1994, he worked at CSK as a chief researcher. In 1994, he joined the Faculty of Information Sciences, Hiroshima City University, as an Associate Professor. Since 2001, he has been a Professor of the Faculty of Engineering, Tokushima University. 
He is a Chair Professor of University of Electronic Science and Technology of China from 2022. 
His current research interests include Natural Language Processing, Artificial Intelligence, Affective Computing, Emotional Robot. 
He is a member of the Academician of The Engineering Academy of Japan, the Academician of EU Academy of Sciences and a Foreign Full Member (Academician) of the Russian Academy of Engineering. 
He is a senior member of IEEE, Editor-in-Chief of International Journal of Advanced Intelligence, a vice president of CAAI, and a Fellow of The Japan Federation of Engineering Societies, a Fellow of IEICE, a Fellow of CAAI. He is the President of International Advanced Information Institute, Japan.
\end{IEEEbiography}

\begin{IEEEbiography}[{\includegraphics[width=0.8in,height=1.in,clip,keepaspectratio]{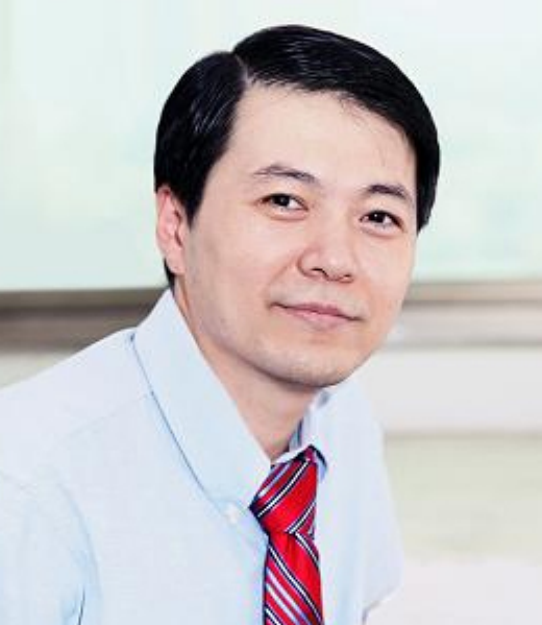}}]{Gang Kou} 
% is a Distinguished Professor of Chang Jiang Scholars Program in Southwestern University of Finance and Economics.
% He received his Ph.D. in 
% % Information Technology from the College of Information Science \& Technology, 
% Univ. of Nebraska at Omaha; 
% % Master degree in Dept of Computer Science, Univ. of Nebraska at Omaha; 
% and B.S. degree in 
% % Department of Physics, 
% Tsinghua University, China. 
% % He has published more than 100 papers in various peer-reviewed journals.
% His h-index is 70 and his papers have been cited for more than 17000 times.
% He is listed as the Highly Cited Researcher by Clarivate Analytics (Web of Science).
is a Distinguished Professor of Chang Jiang Scholars Program in Southwestern University of Finance and Economics, managing editor of International Journal of Information Technology \& Decision Making (SCI) and managing editor-in-chief of Financial Innovation (SSCI). He is also editors for other journals, such as: Decision Support Systems, and European Journal of Operational Research. Previously, he was a professor of School of Management and Economics, University of Electronic Science and Technology of China, and a research scientist in Thomson Co., R\&D. He received his Ph.D. in Information Technology from the College of Information Science \& Technology, Univ. of Nebraska at Omaha; Master degree in Dept of Computer Science, Univ. of Nebraska at Omaha; and B.S. degree in Department of Physics, Tsinghua University, China. He has published more than 100 papers in various peer-reviewed journals. Gang Kou’s h-index is 70 and his papers have been cited for more than 20000 times. He is listed as the Highly Cited Researcher by Clarivate Analytics (Web of Science).

\end{IEEEbiography}

\end{document}